\documentclass[11pt]{article}

\usepackage[final]{acl}

\usepackage{times}
\usepackage{latexsym}

\usepackage[T1]{fontenc}

\usepackage[utf8]{inputenc}

\usepackage{microtype}

\usepackage{inconsolata}

\usepackage{graphicx}
\usepackage{multirow}
\usepackage{booktabs}
\usepackage{array}
\usepackage{xcolor}
\usepackage{tikz}
\usepackage{tcolorbox} 
\usetikzlibrary{positioning, arrows.meta}
\usepackage{subcaption}

\title{How Language Models Conflate Logical Validity with Plausibility:\\ A Representational Analysis of Content Effects}

\author{
 \textbf{Leonardo Bertolazzi\textsuperscript{1}},
 \textbf{Sandro Pezzelle\textsuperscript{2}},
 \textbf{Raffaella Bernardi\textsuperscript{3}},
\\
\\
 \textsuperscript{1}University of Trento,
 \textsuperscript{2}University of Amsterdam,
 \textsuperscript{3}Free University of Bozen-Bolzano
\\
 {\small
   Correspondence: \texttt{leonardo.bertolazzi@unitn.it}
 }
}

\begin{document}
\maketitle
\begin{abstract}
Both humans and large language models (LLMs) exhibit \textit{content effects}: biases in which the plausibility of the semantic content of a reasoning problem influences judgments regarding its logical validity. While this phenomenon in humans is best explained by the dual-process theory of reasoning, the mechanisms behind content effects in LLMs remain unclear.  
In this work, we address this issue by investigating how LLMs encode the concepts of validity and plausibility within their internal representations. 
We show that both concepts are linearly represented and strongly aligned in representational geometry, leading models to conflate plausibility with validity. Using steering vectors, we demonstrate that plausibility vectors can causally bias validity judgements, and vice versa, and that the degree of alignment between these two concepts predicts the magnitude of behavioral content effects across models. Finally, we construct debiasing vectors that disentangle these concepts, reducing content effects and improving reasoning accuracy.
Our findings advance understanding of how abstract logical concepts are represented in LLMs and highlight representational interventions as a path toward more logical systems.
\end{abstract}

\section{Introduction}

\begin{figure}[t]
\centering
    \begin{tikzpicture}[
    arr/.style={->, thick, gray!60},
    steerarr/.style={->, line width=1.6pt, black},
    pill/.style={rounded corners=2pt, line width=0.4pt},
    llmbox/.style={rounded corners=3pt, fill=gray!12, draw=gray!40, line width=0.5pt},
    inputbox/.style={rounded corners=3pt, fill=gray!8, draw=gray!40, line width=0.5pt},
    vectorarr/.style={->, line width=2.0pt,
                      dash pattern=on 4pt off 2.5pt},
]

\node[font=\small\bfseries, text=gray!55] at (1.75, 0.40) {Original};
\node[font=\small\bfseries, text=gray!55] at (5.35, 0.40) {With steering};

\draw[inputbox] (0.10, 0.00) rectangle (3.40,-1.65);
\node[anchor=north west, font=\scriptsize\bfseries]
  at (0.18,-0.01) {Validity classification};
\draw[gray!30, line width=0.3pt] (0.10,-0.45) -- (3.40,-0.45);
\node[anchor=north west, font=\tiny, text=black]
  at (0.18,-0.50) {P1: All dogs are labradors};
\node[anchor=north west, font=\tiny, text=black]
  at (0.18,-0.78) {P2: Some labradors are not canines};
\node[anchor=north west, font=\tiny, text=black]
  at (0.18,-1.06) {C: No dogs are canines};

\draw[arr] (1.75,-1.65) -- (1.75,-1.95);

\draw[llmbox] (1.10,-1.95) rectangle (2.40,-2.30);
\node[font=\scriptsize] at (1.75,-2.13) {LLM};

\draw[arr] (1.75,-2.30) -- (1.75,-2.70);

\draw[pill, fill=green!15, draw=green!15]
  (1.10,-2.70) rectangle (2.40,-3.10);
\node[font=\tiny\bfseries, text=green!40!black]
  at (1.75,-2.90) {\textbf{Invalid}};

\draw[inputbox] (3.70, 0.00) rectangle (7.00,-1.65);
\node[anchor=north west, font=\scriptsize\bfseries]
  at (3.78,-0.01) {Validity classification};
\draw[gray!30, line width=0.3pt] (3.70,-0.45) -- (7.00,-0.45);
\node[anchor=north west, font=\tiny, text=black]
  at (3.78,-0.50) {P1: All dogs are labradors};
\node[anchor=north west, font=\tiny, text=black]
  at (3.78,-0.78) {P2: Some labradors are not canines};
\node[anchor=north west, font=\tiny, text=black]
  at (3.78,-1.06) {C: No dogs are canines};

\draw[arr] (5.35,-1.65) -- (5.35,-1.95);

\draw[llmbox] (4.70,-1.95) rectangle (6.00,-2.30);
\node[font=\scriptsize] at (5.35,-2.13) {LLM};

\draw[vectorarr, orange!70!black]
  (6.40,-2.13) -- (6.00,-2.13);
\node[font=\tiny\bfseries, text=orange!70!black, anchor=west]
  at (6.41,-2.13) {\parbox{1cm}{\tiny\bfseries $+$ Truth\newline\hspace*{6pt}dir}};

\draw[arr] (5.35,-2.30) -- (5.35,-2.70);

\draw[pill, fill=red!15, draw=red!15]
  (4.70,-2.70) rectangle (6.00,-3.10);
\node[font=\tiny\bfseries, text=red!55!black]
  at (5.35,-2.90) {\textbf{Valid}};

\draw[gray!25, line width=0.4pt, dashed] (0.0,-3.35) -- (7.60,-3.35);

\draw[inputbox] (0.10,-3.60) rectangle (3.40,-5.00);
\node[anchor=north west, font=\scriptsize\bfseries]
  at (0.18,-3.61) {Plausibility classification};
\draw[gray!30, line width=0.3pt] (0.10,-4.05) -- (3.40,-4.05);
\node[anchor=north west, font=\tiny, text=black]
  at (0.18,-4.10) {Statement:};
\node[anchor=north west, font=\tiny, text=black]
  at (0.18,-4.38) {No dogs are canines};

\draw[arr] (1.75,-5.00) -- (1.75,-5.30);

\draw[llmbox] (1.10,-5.30) rectangle (2.40,-5.65);
\node[font=\scriptsize] at (1.75,-5.48) {LLM};

\draw[arr] (1.75,-5.65) -- (1.75,-6.05);

\draw[pill, fill=blue!15, draw=blue!15]
  (1.10,-6.05) rectangle (2.40,-6.45);
\node[font=\tiny\bfseries, text=blue!55!black]
  at (1.75,-6.25) {\textbf{False}};

\draw[inputbox] (3.70,-3.60) rectangle (7.00,-5.00);
\node[anchor=north west, font=\scriptsize\bfseries]
  at (3.78,-3.61) {Plausibility classification};
\draw[gray!30, line width=0.3pt] (3.70,-4.05) -- (7.00,-4.05);
\node[anchor=north west, font=\tiny, text=black]
  at (3.78,-4.10) {Statement:};
\node[anchor=north west, font=\tiny, text=black]
  at (3.78,-4.38) {No dogs are canines};

\draw[arr] (5.35,-5.00) -- (5.35,-5.30);

\draw[llmbox] (4.70,-5.30) rectangle (6.00,-5.65);
\node[font=\scriptsize] at (5.35,-5.48) {LLM};

\draw[vectorarr, red!60!black] 
  (6.40,-5.48) -- (6.00,-5.48);
\node[font=\tiny\bfseries, text=red!60!black, anchor=west]
  at (6.41,-5.48) {\parbox{1cm}{\tiny\bfseries $+$ Validity\newline\hspace*{6pt}dir}};

\draw[arr] (5.35,-5.65) -- (5.35,-6.05);

\draw[pill, fill=orange!20, draw=orange!20]
  (4.70,-6.05) rectangle (6.00,-6.45);
\node[font=\tiny\bfseries, text=orange!70!black]
  at (5.35,-6.25) {\textbf{True}};

\end{tikzpicture}%
\caption{Cross-task steering changes classification behavior. \textbf{Top:} Adding a \textit{plausibility} (truth) direction cause the model to flip its validity judgment. \textbf{Bottom:} Adding a \textit{validity} direction cause the model to flip its plausibility judgment. This cross-task influence reflects the geometric entanglement between validity and plausibility directions in the model's representation space.}
\label{fig:intro}
\end{figure}

A pure abstract reasoner applies logical rules and manipulates symbols independently of the content or context in which they appear. Both humans and LLMs often deviate from this ideal, exhibiting systematic biases where content influences formal reasoning.

Content effects are well-documented in human reasoning tasks, such as judging the validity of syllogistic inferences. 
In this context, \textit{validity} is a logical property of an argument that depends solely on its structure, namely whether the conclusion necessarily follows from the premises independently of their instantiated truth values; \textit{plausibility} concerns whether a statement is true in the real world.\footnote{Although plausibility can be interpreted as a graded concept, throughout this paper we operationalize it as a binary notion; see Section \ref{sec:data}.}
For instance, human participants often judge syllogisms with plausible conclusions as valid, despite being logically incorrect \citep{evans1983conflict}.
Among the theories proposed to account for such biases, the dual-process theory of reasoning has been the most influential, positing two distinct modes of thought: a fast, intuitive, heuristic-driven system (System 1), and a slower, deliberative system responsible for analytical reasoning~\cite[System 2;][]{evans2008dual, kahneman2011thinking}. Neuroscientific studies have provided empirical support for this framework, highlighting different neural substrates associated with these reasoning processes \citep{goel2000dissociation, stollstorff2012neuralcorrelates}. Recent work has found that LLMs exhibit similar content effects in reasoning tasks \citep{lampinen2024content}; however, the underlying mechanisms driving these effects remain unknown.

In this work, we provide a representational account of why content effects may emerge in LLMs, investigating how the abstract concepts of validity and plausibility are encoded in their hidden representation space.
Specifically, we build upon the linear representation hypothesis \citep{park2023linear}, which proposes that many high-level concepts are encoded linearly within the latent space of LLMs. This hypothesis has been supported by empirical findings showing that concepts can often be captured by linear probes or manipulated with steering vectors \citep{liu2024steering, rimsky2024caa, marks2024geometry}.  
We hypothesize that content effects in LLMs may arise from the way validity and plausibility are entangled within the model's representational geometry. Specifically, we predict that LLMs conflate validity with plausibility, leading to systematic biases in reasoning.  
To test this hypothesis, we analyze ten different LLMs using zero-shot and chain-of-thought (CoT) prompting \citep{wei2022cot} and address the following research questions:

\begin{figure*}[t!]
    \centering
    \begin{tikzpicture}[node distance=1cm]
    \node[font=\Large\bfseries] at (3.5, 4.5) {Plausible};
    \node[font=\Large\bfseries] at (10.5, 4.5) {Implausible};
    
    \node[font=\Large\bfseries] at (-1, 3.0) {Valid};
    \node[font=\Large\bfseries] at (-1, 0.7) {Invalid};
    
    \node[fill=blue!20, rounded corners=10pt, minimum width=6.2cm, minimum height=2cm, 
          text width=5.5cm, align=left] at (3.5, 3.0) {
        \textbf{P1:} All labradors are canines.\\
        \textbf{P2:} All labradors are dogs.\\
        \textbf{C:}\hspace{0.35cm}\textit{Some dogs are canines.}
    };
    
    \node[fill=red!20, rounded corners=10pt, minimum width=6.6cm, minimum height=2cm,
          text width=5.5cm, align=left] at (10.5, 3.0) {
        \textbf{P1:} All canines are cats.\\
        \textbf{P2:} All canines are dogs.\\
        \textbf{C:}\hspace{0.35cm}\textit{Some dogs are cats.}
    };
    
    \node[fill=green!20, rounded corners=10pt, minimum width=6.0cm, minimum height=2cm,
          text width=5.5cm, align=left] at (3.5, 0.7) {
        \textbf{P1:} No dogs are cats.\\
        \textbf{P2:} No canines are cats.\\
        \textbf{C:}\hspace{0.35cm}\textit{Some dogs are canines.}
    };
    
    \node[fill=orange!20, rounded corners=10pt, minimum width=6.2cm, minimum height=2cm,
          text width=6.2cm, align=left] at (10.5, 0.7) {
        \hspace{0.2cm}\textbf{P1:} All dogs are labradors.\\
        \hspace{0.2cm}\textbf{P2:} Some labradors are not canines.\\
        \hspace{0.2cm}\textbf{C:}\hspace{0.35cm}\textit{No dogs are canines.}
    };
\end{tikzpicture}
    \caption{\textbf{Validity and plausibility configurations.} Illustrative examples of valid and invalid syllogisms with plausible and implausible conclusions. Here, \textit{plausible} indicates that the conclusion is true in the real world, whereas \textit{implausible} indicates that it is false.}
    \label{fig:data_sample}
\end{figure*}

\textbf{RQ1:} \textit{Do current LLMs exhibit content effects?} We find that models from the Qwen-2.5 \citep{qwen2024qwen2.5}, Qwen-3 \citep{qwen2025qwen3}, and Gemma-3 \citep{gemma2025gemma3} families display systematic biases where plausibility influences validity judgments.

\textbf{RQ2:} \textit{How do LLMs encode plausibility and validity?} We find single vectors that can control models' judgements for both validity and plausibility. Moreover, these vectors are highly similar.

\textbf{RQ3:} \textit{What do the representations reveal about behavioral content effects?} We demonstrate that greater geometric similarity between validity and plausibility vectors is correlated with stronger behavioral content effects across models. Moreover, we establish causal interaction across concepts: plausibility vectors steers validity judgments, and vice versa (see Figure~\ref{fig:intro}).

\textbf{RQ4:} \textit{Can we design an intervention to mitigate content effects?} We develop debiasing steering vectors that disentangle validity from plausibility, reducing content effects while improving reasoning accuracy.\footnote{The code and data used for this paper are publicly available at: \url{https://github.com/leobertolazzi/content-effect-interpretability.git}}

\section{Related Work}

\subsection{Content Effects in Humans and LLMs}

Content effects describe the well-documented tendency in humans to evaluate reasoning problems based on their semantic content and prior beliefs rather than logical structure \citep{markovits1989beliefbias}. This phenomenon has been extensively studied in syllogistic reasoning \citep{evans1983conflict, oakhill1985effect}, the Wason selection task \citep{wason1968task}, and Bayesian inference, where existing beliefs can override statistical evidence \citep{kahnemantversky1973prediction, barhillel1980baserate}.

Recent research has revealed that LLMs exhibit similar content effects. \citet{lampinen2024content} found that LLMs exhibit this bias in the Wason selection task and syllogistic inference. \citet{bertolazzi-etal-2024-systematic} showed that in a multiple-choice setting with plausible and implausible conclusions, LLMs favor the former as valid conclusions in syllogisms, regardless of their logical validity. \citet{balappanawar2025pigsflyllmslogically} demonstrated that content effects in LLMs extend beyond conditional reasoning and syllogistic inference to various inference rules in propositional logic, suggesting a broader pattern of belief-based reasoning. While this research provides robust behavioral evidence for content effects in LLMs, the underlying mechanisms governing this bias remain poorly understood.

\subsection{Linear Representations in LLMs}
Recent work suggests that many high-level concepts are encoded linearly in the latent space of LLMs and can be manipulated using steering vectors \citep{liu2024steering}.
For instance, \citet{zhao2025llmsencodeharmfulnessrefusal} identified a harmfulness direction, where steering along this dimension causes LLMs to interpret harmless instructions as harmful. Similarly, \citet{marks2024geometry} show that truth is linearly represented in LLMs, with interventions along the truthful direction leading models to treat false statements as true, and vice versa. Other concepts, including sentiment \citep{hollinsworth-etal-2024-language} and refusal \citep{arditi2025refusal}, have also been found to be linearly encoded within LLMs, among others. At the same time, recent work has also shown that not all concepts adhere to the linear representation hypothesis, with some high-level features requiring more complex, non-linear encoding structures \citep{engels2025nonlinear}.

While the concepts that interest us the most~---~truth \citep{marks2024geometry} and logical validity \citep{valentino2025mitigatingcontenteffectsreasoning}~---~have been shown to be linearly represented in the latent space of LLMs, no study has yet examined how content effects in LLMs might emerge from the interaction between these two dimensions.

\section{Method}
\label{sec:data}

\subsection{Datasets and Tasks}

We focus on syllogistic inferences, a type of logical problem in which LLMs exhibit content effects \citep{lampinen2024content, bertolazzi-etal-2024-systematic}.\footnote{For a detailed explanation of the structure of syllogisms, see Appendix \ref{app:syllogisms}.}

\paragraph{Data.} Starting from the data from \citet{bertolazzi-etal-2024-systematic}, we construct 1,280 syllogisms, each containing two premises and a conclusion. We utilize this dataset because it systematically covers all 64 types of syllogism across two distinct conditions: “plausible” and “implausible” ones. The plausible syllogisms have conclusions that are true in the actual world (e.g., “Some dogs are canines”), while implausible syllogisms' conclusions are false in the actual world (e.g., “Some dogs are cats”). The dataset is constructed using ten distinct triples of terms exhibiting a taxonomical relationship (e.g., “labradors” $\rightarrow$ “dogs” $\rightarrow$ “canines”). We provide further details on the data generation process in Appendix \ref{app:data}. Although the dataset is relatively small in scale, it is sufficiently large to cover all 64 syllogism types with meaningful semantic variation; additional instances would introduce redundancy rather than new logical configurations. Furthermore, the dataset is carefully constructed and balanced across syllogism types and plausible vs.~implausible conclusions.

\paragraph{Logical validity classification task.} Each syllogism can be classified as either valid or invalid and has a conclusion that can be either plausible or implausible. This creates four distinct categories of syllogisms that enable us to examine the interaction between logical structure and content plausibility. Figure \ref{fig:data_sample} provides illustrative examples of each combination of validity and plausibility labels. Across the experiments we perform, we partition this dataset using a 70-30 train-test split. Models are tasked with performing syllogistic reasoning as a binary classification problem, where they must determine the logical validity of presented syllogisms.

\paragraph{Plausibility classification task.} Plausibility can be interpreted as an inherently continuous and nuanced concept. However, to enable comparison with logical validity, we operationalize it as a binary notion. Throughout this paper, we define plausibility operationally as the truth value of a statement with respect to the actual world: a statement is considered plausible if it is factually true. We extract all unique conclusions present in the original data from \citet{bertolazzi-etal-2024-systematic} and construct a task requiring models to classify these statements as either true or false based on what models ``believe'' to be factually accurate.\footnote{The complete prompt formulations for the logical validity and the plausibility tasks are provided in Appendix \ref{app:data}.}

\paragraph{Auxiliary tasks.} We additionally incorporate control datasets for two auxiliary binary classification tasks. The first task requires, given a source and a target term in a taxonomical relationship, to determine whether the source is a hypernym or a hyponym of the target. Crucially, this dataset is built using the same ten triples that were used in the two binary classification tasks above. The second task involves harmful/harmless content classification using the data from \citet{arditi2025refusal}.\footnote{This dataset consists of a set of prompts or instructions, and an LLM is tasked with judging whether they are harmful or harmless. See Appendix \ref{app:data} for additional details on the data of these two control datasets.} These auxiliary binary classification tasks serve as experimental controls to verify that any observed interactions between plausibility and validity are not merely artifacts of both tasks being binary classifications or of sharing similar vocabulary and prompt structures.

\subsection{Representing Binary Concepts as Single Directions}
We represent each binary concept as a linear direction in the model's activation space using the difference-in-means approach \citep{belrose2023diff, rimsky2024caa, marks2024geometry}. 
For a binary concept with a positive class (e.g., valid, plausible, harmless) and a negative class (e.g., invalid, implausible, harmful), we compute the mean activation vectors for each class at layer $l$ and at the \emph{last token} position right before the label prediction. Importantly, we use the model's predicted labels rather than ground-truth labels to define class membership, as our focus is on understanding the model's own internal ``beliefs''. The concept direction is then defined as:

$$
v^l_\mathrm{concept} = \mu^l_{\mathrm{positive}} - \mu^l_{\mathrm{negative}}
$$

\noindent where $\mu^l_{\mathrm{positive}}$ and $\mu^l_{\mathrm{negative}}$ are the mean hidden activations of the positive and negative classes, respectively. This formulation defines a direction in activation space that points from the negative class toward the positive class. 

\subsection{Steering Approach}

Once we extract a vector representing a binary concept, we can then use it as a steering vector \citep{liu2024steering}. We manipulate the model's behavior by adding or subtracting the vector from the same layer $l$ from which it was extracted: adding $v^l_\mathrm{concept}$ to the activations steers the model toward the positive class (e.g., making outputs more valid, plausible, or harmless), while subtracting it steers toward the negative class (e.g., making outputs more invalid, implausible, or harmful). This bidirectional control allows us to test whether the concept is linearly represented at each layer by measuring how effectively these interventions change the model's classification behavior. Across our steering experiments, each vector is added or subtracted as is, without multiplying it by a scalar or normalizing, which means that we expect the vector to represent the concept faithfully in both direction and magnitude.

\subsection{Metrics}
\label{sec:metrics}

We use the following metrics in our experiments.

\paragraph{Content effect.} When evaluating models on the logical validity classification task, we need a way to precisely quantify the degree of content effect. To this end, we introduce a metric that enables fine-grained comparisons across models. Let $A(S)$ denote the accuracy of a model on a subset of the logical validity dataset $S \subseteq D$. We partition the dataset $D$ by validity ($v^+$ for valid, $v^-$ for invalid) and plausibility ($p^+$ for plausible, $p^-$ for implausible), yielding four disjoint subsets $D_{v^+,p^+}, D_{v^+,p^-}, D_{v^-,p^+}, D_{v^-,p^-}$. We then define:
$$
\begin{array}{l}
    \Delta_{v^+} = A(D_{v^+,p^+}) - A(D_{v^+,p^-}), \\
    \Delta_{v^-} = A(D_{v^-,p^-}) - A(D_{v^-,p^+})
\end{array}
$$
\noindent Here, $\Delta_{v^+}$ captures how much more accurately the model classifies valid arguments when conclusions are plausible, while $\Delta_{v^-}$ captures the corresponding effect for invalid arguments with implausible conclusions.

The overall \emph{content effect} (CE) is the mean of these two components:

$$
\mathrm{CE} = \frac{1}{2}(\Delta_{v^+} + \Delta_{v^-}).
$$

\noindent This metric is bounded between $-1$ and $1$: $\mathrm{CE}  = 1$ indicates judgments of validity are fully driven by plausibility, $\mathrm{CE}  = 0$ indicates independence of validity and plausibility.\footnote{A negative CE would instead indicate bias in the opposite direction.}

\paragraph{Steering power.} Since we work with binary concepts and apply steering vectors to control model behavior on binary classification tasks, we introduce \emph{steering power} as a measure of the effectiveness of such vectors at a given layer $l$. Following the approach described above, we always steer against the model's original prediction: \emph{adding} $v^l$ when the model predicts negative, and \emph{subtracting} it when the model predicts positive.

Formally, the \emph{steering power} (SP) of a vector $v^l$ is the proportion of examples whose predicted label flips when the signed vector is applied at layer $l$:
$$
\mathrm{SP}(v;\mathcal{D}) = \frac{1}{n}\sum_{i=1}^n 
\mathbf{1}\!\Big[\hat{y}_i^{\,\prime}\neq \hat{y}_i \Big]
$$
where $\hat{y}_i$ is the original prediction for input $x_i$, and $\hat{y}_i^{\,\prime}$ is the steered prediction.

\subsection{Models}

We ran experiments using instruction-tuned versions of three variants of LLMs: Qwen-2.5 \citep{qwen2024qwen2.5}, Qwen-3 \citep{qwen2025qwen3}, and Gemma-3 \citep{gemma2025gemma3}, with parameter counts ranging from 4B to 32B.
This choice combines both standard instruction-tuned models (Qwen-2.5 and Gemma-3) and newer ``thinking'' models (Qwen-3) that have been trained to produce long reasoning traces during inference.  

Across all experiments, we evaluate models using both zero-shot prompting and CoT prompting.
We primarily present results using \texttt{Qwen2.5-32B-Instruct} and \texttt{Qwen3-14B} as our main reference models. We observe consistent patterns across all evaluated models, with full comparative results in Appendix \ref{app:additional_results}.

\section{Experiments and Results}

\subsection{Do LLMs Exhibit Content Effects?}
\label{sec:behavioral}

We ask whether LLMs suffer from content effects on the logical validity classification task. We expect biased LLMs to classify more accurately syllogisms where plausibility supports logical validity: that is, when the conclusion is plausible and the argument is valid, or when the conclusion is implausible and the argument is invalid. Conversely, models should have lower accuracy judging the validity of other cases.

Table \ref{tab:behavioral} reports the accuracies of \texttt{Qwen2.5-32B-Instruct} and \texttt{Qwen3-14B} on the test split of the logical validity classification task. Both models exhibit content effects in the zero-shot setting, while the bias is greatly reduced with CoT prompting. Looking at the CE metric, we observe that for \texttt{Qwen2.5-32B-Instruct} it is relatively high in zero-shot ($0.348$), showing strong conflation of plausibility with validity. When prompted with CoT, CE drops substantially to $0.096$, indicating that explicit reasoning almost eliminates the bias. For \texttt{Qwen3-14B}, the zero-shot CE is lower ($0.213$), suggesting less susceptibility to content effects even without structured reasoning. When given CoT prompting, CE drops almost to zero ($0.014$), implying that validity judgments become nearly independent of plausibility.

Across all models, CoT prompting achieves a significantly lower CE compared to zero-shot (Mann-Whitney U test, $p < 0.01$; see the complete behavioral results in Table \ref{tab:behavioral-full} in the Appendix).

\begin{table}[t!]
    \centering
    \begin{tabular}{lcccc}
    \toprule
     & \multicolumn{2}{c}{\textbf{Qwen2.5-32B}} & \multicolumn{2}{c}{\textbf{Qwen3-14B}} \\
     \cmidrule(lr){2-3} \cmidrule(lr){4-5}
     & \textbf{0-shot} & \textbf{CoT} & \textbf{0-shot} & \textbf{CoT} \\
    \midrule
    $D_{v^+,p^+}$ & 100.00 & 98.67 & 97.33 & 95.31 \\
    $D_{v^-,p^+}$ & 67.50 & 86.64 & 90.83 & 99.10 \\
    \addlinespace
    $D_{v^+,p^-}$ & 60.92 & 93.10 & 60.92 & 92.50 \\
    $D_{v^-,p^-}$ & 98.04 & 100.00 & 97.06 & 99.90 \\
    \midrule
    Acc & 81.62 & 94.60 & 86.54 & 96.70 \\
    \midrule
    CE & 0.348 & 0.096 & 0.213 & 0.014 \\
    \bottomrule
    \end{tabular}
    \caption{\textbf{Behavioral performance.} Zero-shot vs CoT accuracy for \texttt{Qwen2.5-32B-Instruct} and \texttt{Qwen3-14B} on different subsets of syllogism dataset and overall content effect. Subsets are organized by validity label ($v^+ = \mathrm{valid}$, $v^- = \mathrm{invalid}$) and plausibility of the conclusion ($p^+ = \mathrm{plausible}$, $p^- = \mathrm{implausible}$).}
    \label{tab:behavioral}
\end{table}

\begin{figure*}[t!]
    \centering
    \begin{subfigure}[t]{\textwidth}
        \centering
        \includegraphics[width=\linewidth]{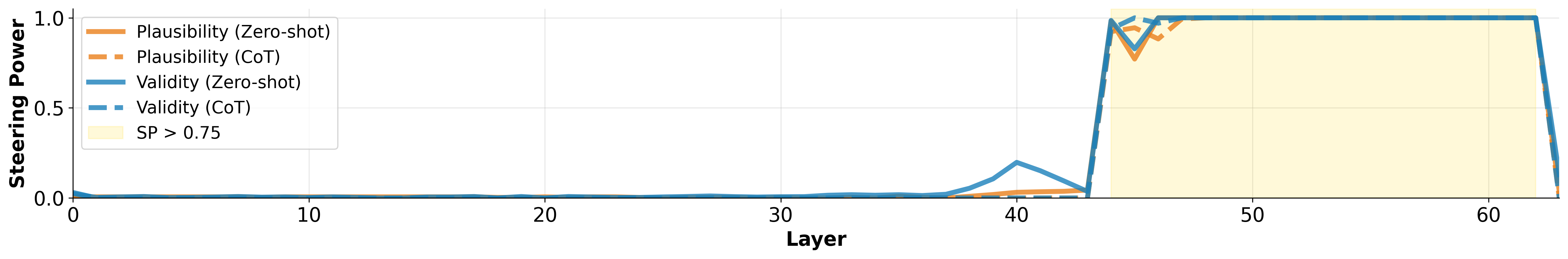}
        \caption{}
        \label{fig:steering_power}
    \end{subfigure}
        
    \begin{subfigure}[t]{0.35\textwidth}
        \centering
        \includegraphics[width=\linewidth]{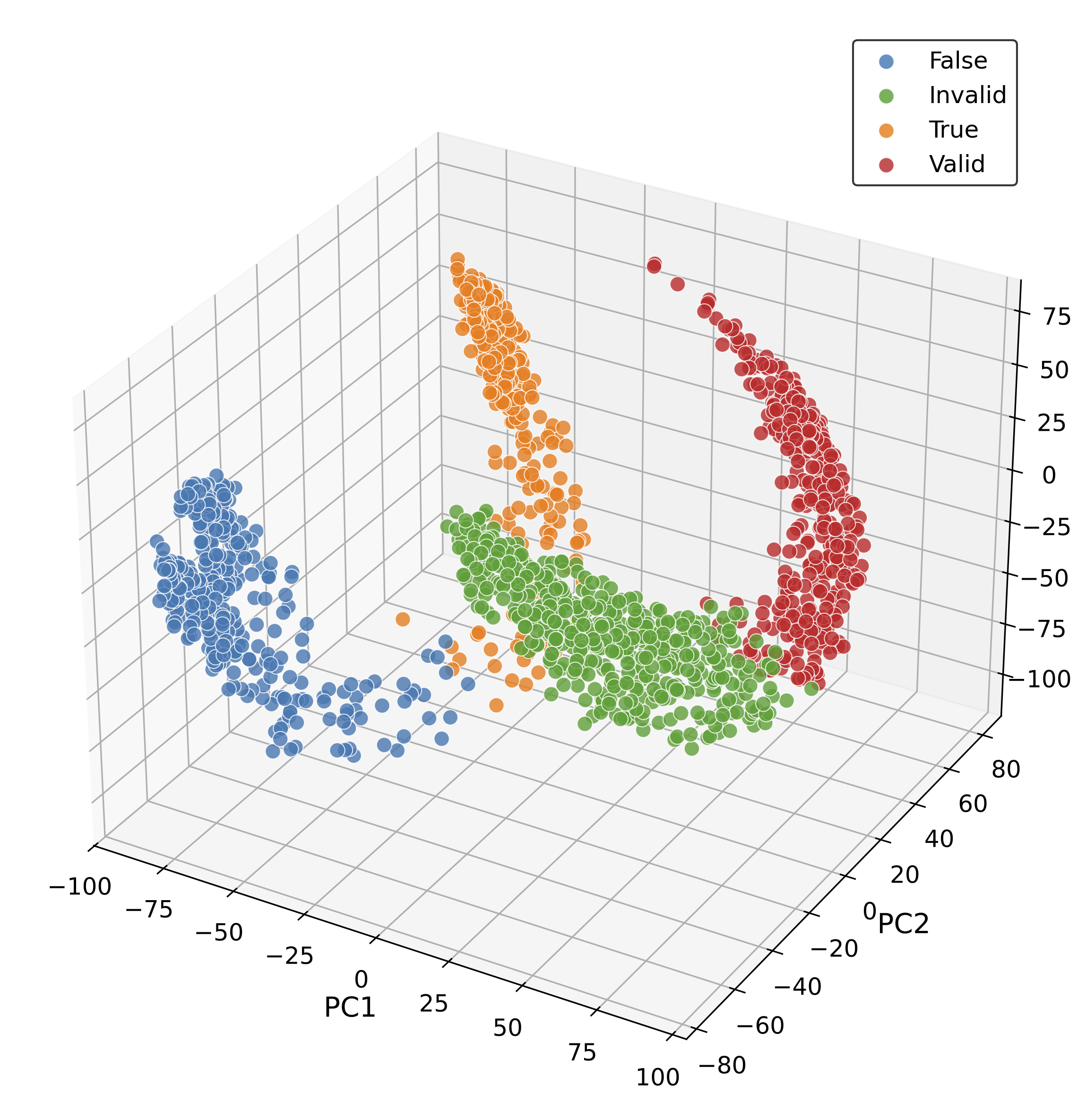}
        \caption{}
        \label{fig:pca}
    \end{subfigure}
    \hspace{2cm}
    \begin{subfigure}[t]{0.40\textwidth}
        \centering
        \includegraphics[width=\linewidth]{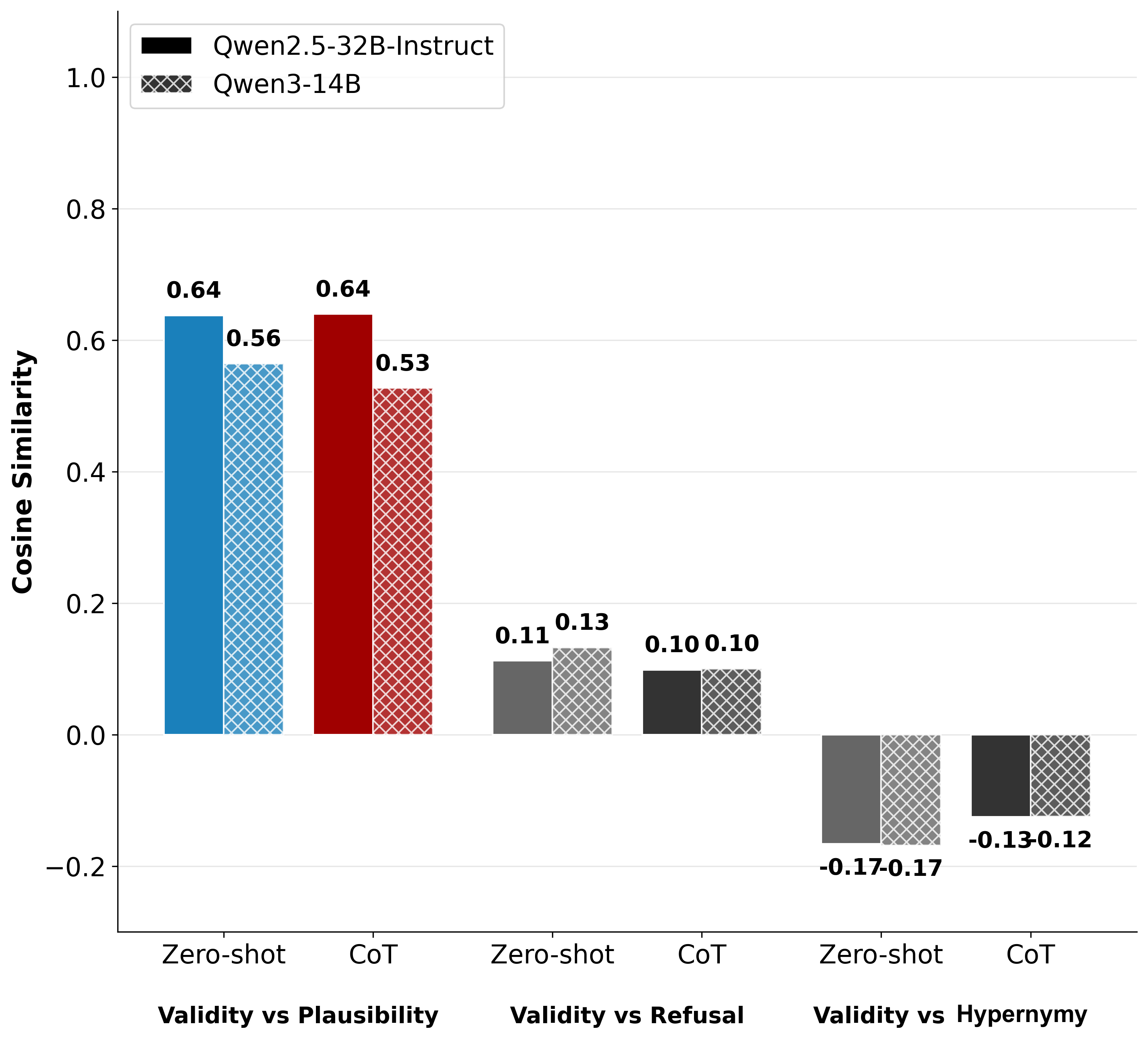}
        \caption{}
        \label{fig:similarity}
    \end{subfigure}
    
    \caption{\textbf{Representational analysis of validity and plausibility concepts}.
    (a) Steering power (SP) of validity and plausibility vectors applied at different hidden layers of \texttt{Qwen2.5-32B-Instruct}. The region in yellow highlights layers with $\mathrm{SP} > 0.75$. Validity and plausibility steering vectors show high SP at similar layers using both zero-shot and CoT prompting.
    (b) 3D PCA projection of hidden states from layer 50 of \texttt{Qwen2.5-32B-Instruct} in the zero-shot setting showing four distinct clusters corresponding to model predictions (valid/invalid, true/false). The parallel geometric structure between true/false and valid/invalid clusters suggests shared representational directions for plausibility and validity. 
    (c) Average cosine similarity between the validity vector and vectors for the concepts of plausibility, hypernymy, and harmlessness across all layers for both models under zero-shot and CoT prompting. High validity-plausibility alignment (0.53 to 0.64) contrasts with low alignment for other concepts (0.10 to 0.13 and -0.12 to -0.17), confirming specific representational entanglement.}
\end{figure*}

\subsection{How are Plausibility and Validity Encoded in LLMs?}
\label{sec:representations}

We now turn to how validity and plausibility are represented internally by LLMs. 

\paragraph{Single directions control validity and plausibility judgements.}  
To test the linear representation hypothesis on validity and plausibility, we conduct steering experiments to identify layers where these concepts are encoded linearly. For each hidden layer $l$, we compute the difference-in-means vectors
$$
v^l_\mathrm{validity} = \mu^l_{v^+} - \mu^l_{v^-}, \quad
v^l_\mathrm{plausibility} = \mu^l_{p^+} - \mu^l_{p^-}
$$
representing validity and plausibility directions at layer $l$. Analogously, using the auxiliary datasets (harmful/harmless, and hypernym/hyponym), we obtain $v^l_\mathrm{harmlessness} = \mu^l_{harm^+} - \mu^l_{harm^-}$ and $v^l_\mathrm{hypernymy} = \mu^l_{hyp^+} - \mu^l_{hyp^-}$, respectively.

Figure \ref{fig:steering_power} shows SP values across all 63 layers of \texttt{Qwen2.5-32B-Instruct} in both zero-shot and CoT prompting. Both validity and plausibility vectors achieve $\mathrm{SP} \approx 1$ at late layers, but near zero at early layers. Steering is equally effective in zero-shot and CoT, and both tasks peak at similar layers. This demonstrates that logical validity and plausibility classification can be effectively controlled by single directions in the model's latent space.

\paragraph{Validity and plausibility vectors are similar.}
Having identified the late layers where steering is effective, we now examine whether validity and plausibility are represented similarly. Figure \ref{fig:pca} provides a qualitative visualization through a 3D PCA projection of hidden states from layer 50 (of 64) of \texttt{Qwen2.5-32B-Instruct} using data from both validity and plausibility classification in the zero-shot setting. In the logical validity task, the model classifies arguments as valid or invalid, while in the plausibility task, it classifies statements as true or false.\footnote{Labels in the plot are model predictions rather than ground truth, since we are interested in the model's internal representation of what it "believes" to be valid/invalid or true/false.} At this late layer, four well-separated clusters emerge. Notably, the plausibility clusters (true/false) are displaced along a similar direction as validity clusters (valid/invalid), indicating parallel representational structure.\footnote{We include visualizations across layers for both the zero-shot and CoT settings in the Appendix \ref{app:visualization}.}

\begin{figure*}[t!]
    \centering
    \includegraphics[width=\linewidth]{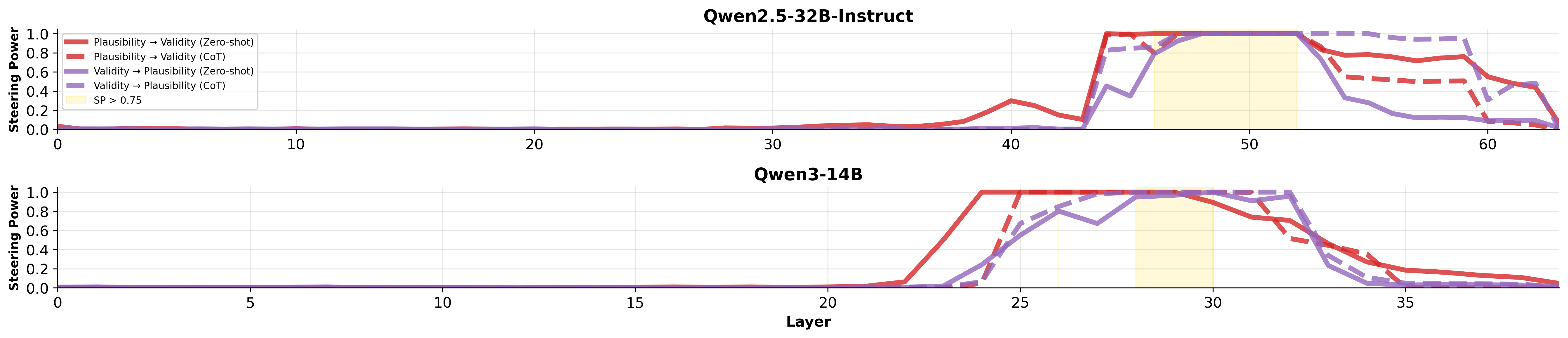}
    \caption{\textbf{Cross-task steering.} Average steering power (SP) of plausibility steering vectors when applied during the logical validity classification task (``plausibility $\rightarrow$ validity''), and vice versa (``validity $\rightarrow$ plausibility''), for \texttt{Qwen2.5-32B-Instruct} and \texttt{Qwen3-14B}, under both zero-shot and CoT prompting.}
    \label{fig:steering}
\end{figure*}

To quantitatively verify the similarity observed in the visualization, we compute the average cosine similarity between validity and plausibility steering vectors. Specifically, we restrict to layers with $\mathrm{SP} > 0.75$ (highlighted in yellow in Figure \ref{fig:steering_power}), since steering vectors are meaningful only when they can effectively control predictions.
Because both validity and plausibility concepts are extracted from prompts that instantiate binary classification tasks and have lexical overlap, we also compute similarities between validity and harmlessness, and between validity and hypernymy, as controls. Figure \ref{fig:similarity} reports the average cosine similarities across layers for \texttt{Qwen2.5-32B-Instruct} and \texttt{Qwen3-14B} under zero-shot and CoT prompting. For both models, validity and plausibility vectors are substantially aligned (cosine similarity $0.48$ to $0.64$), while the two other concepts have a significantly lower similarity ($0.10$ to $0.13$ and $-0.12$ to $-0.17$). This indicates that the observed similarity is specific to validity and plausibility. Across all evaluated models, unlike CE, the degree of alignment between validity and plausibility vectors does not differ significantly between prompting conditions (Mann-Whitney U test, $p = 0.625$).

\subsection{What do the Representations of Validity and Plausibility Reveal about Behavioral Content Effects?}
\label{sec:regression}

We have demonstrated that individual vectors can influence models' validity and plausibility judgments, and that these vectors are highly similar. The next question is whether this similarity is meaningful for better understanding how behavioral content effects emerge: specifically, is the degree of similarity between validity and plausibility vectors predictive of the magnitude of content effects observed in a model? Additionally, if an association exists, is this relationship merely correlational, or do these concepts exhibit a form of causal interaction, where a vector extracted from one concept (e.g., plausibility) can influence the model's predictions for the other concept (e.g., validity), and vice versa?

\begin{figure}[t]
    \centering
    \includegraphics[width=\linewidth]{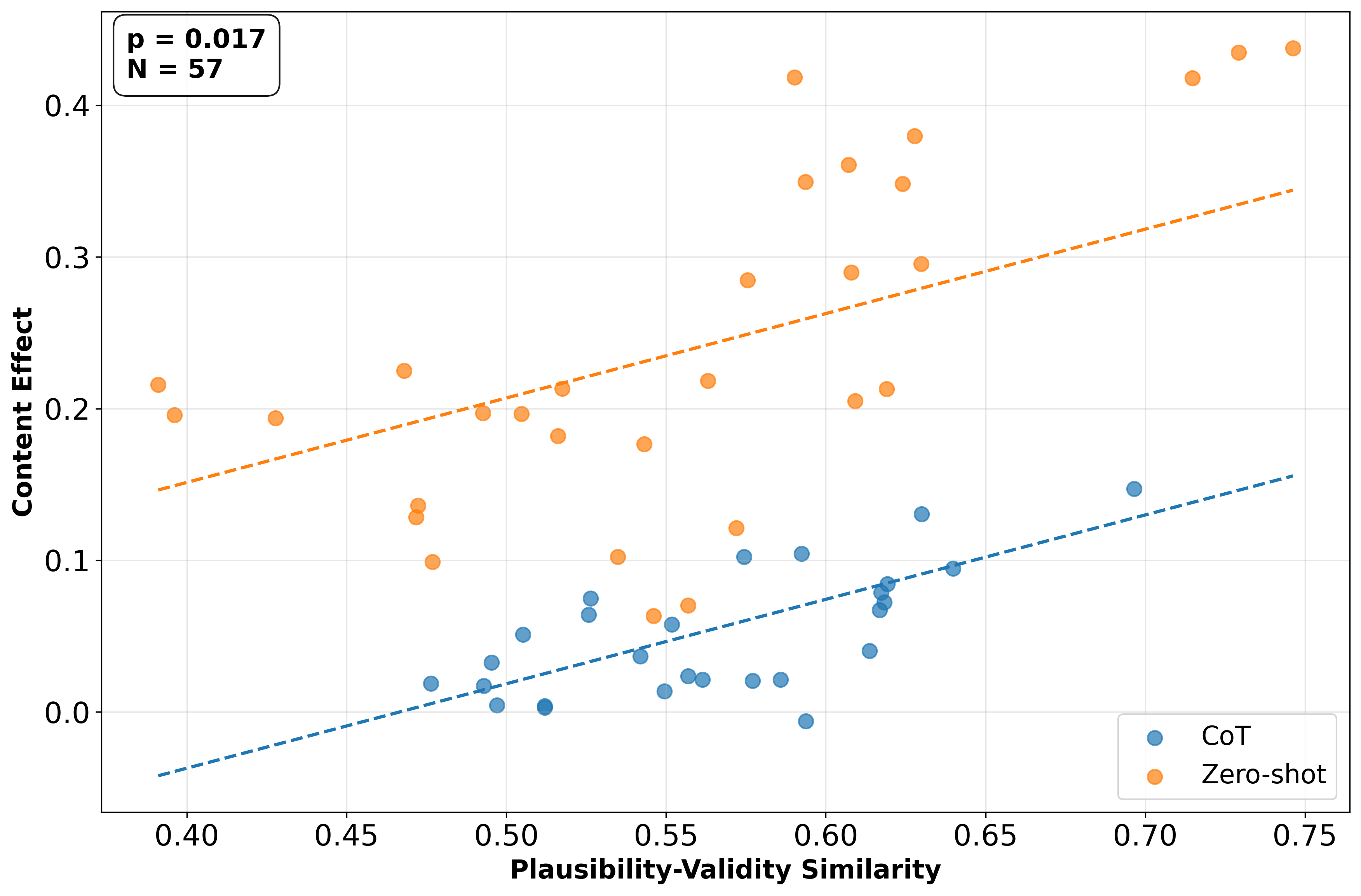}
    \caption{\textbf{Mixed-effects regression.} Relationship between average plausibility–validity similarity and content effect across model–prompt pairs. Points are colored by prompting style (zero-shot vs. CoT). As similarity increases, content effects generally increase, and zero-shot prompts tend to produce higher content effects than CoT prompts at comparable similarity levels.}  
    \label{fig:regression}
\end{figure}

\paragraph{Predicting content effect.} Figure \ref{fig:similarity} shows that within each prompting condition, models with higher CE exhibit higher similarity values. To rigorously test whether higher representational similarity between plausibility and validity vectors correlates with stronger behavioral CE we fit a linear mixed-effects. For this analysis, we aggregated data across all models, each evaluated under both zero-shot and CoT prompting. Additionally, for each model we considered two paraphrases of the original task prompts, resulting in up to three prompt variations per model–prompt pair (yielding a total of 57 data points after excluding certain model–prompt pairs\footnote{See Appendix \ref{app:full_mixed-effect} for details on which model–prompt pairs were excluded and complete regression results.}).

For each model-prompt pair, we computed the average plausibility–validity cosine similarity over highly steerable layers ($\mathrm{SP} > 0.75$). We then fit the following mixed-effects model, with content effect (CE) as the dependent variable and two independent variables: average similarity and prompting style. LLMs were included as a random intercept to account for repeated measurements across models:
$$
CE \sim \mathrm{Prompt} + \mathrm{AvgSim} + (1|\mathrm{LLM})
$$  

The regression results indicate that average similarity is a significant positive predictor of content effect ($\beta = 0.557$, $p = 0.017$), confirming that model–prompt pairs with higher plausibility–validity alignment tend to exhibit stronger content effects. Prompting style also had a significant effect, with zero-shot prompts being associated with increased content effects compared with CoT prompts ($\beta = 0.188$, $p < 0.001$). The random intercept for LLMs was small (variance = 0.001), indicating that these relationships were largely consistent across models.

Figure \ref{fig:regression} shows each point representing a model–prompt pair, with the x-axis corresponding to the average plausibility–validity similarity, and the y-axis corresponding to the content effect. CoT and zero-shot prompts are differentiated by color. As illustrated, content effects generally increase with average similarity, and zero-shot prompts are associated with higher content effects than CoT prompts at comparable levels of similarity.

\paragraph{Causal cross-influence between plausibility and validity.}
\label{sec:cross-influence}
Having observed that both plausibility and validity are linearly steerable, and that higher similarity between these vectors is associated with stronger content effects, we now turn to a causal experiment to investigate whether plausibility vectors can steer validity predictions, and vice versa. As before, steering is applied with a sign determined by the model’s predicted label: the vector is added if the model predicts the negative class and subtracted if it predicts the positive class. This ensures that steering always attempts to flip the model’s decision.

Figure \ref{fig:steering} shows the steering power of plausibility vectors when applied to hidden states during the validity classification task (“plausibility $\rightarrow$ validity”) and the steering power of validity vectors when applied during the plausibility classification task (“validity $\rightarrow$ plausibility”), for both \texttt{Qwen2.5-32B-Instruct} and \texttt{Qwen3-14B}, under zero-shot and CoT prompting. For both models and prompting styles, we observe high SP in a subset of late layers in both directions, indicating that plausibility vectors can causally influence validity judgments while validity vectors can causally influence plausibility judgments. Moreover, \texttt{Qwen3-14B}, which exhibits a lower CE than \texttt{Qwen2.5-32B-Instruct}, has a smaller set of layers where steering transfers effectively compared with the latter model.

\begin{figure}[t!]
    \centering
    \includegraphics[width=\linewidth]{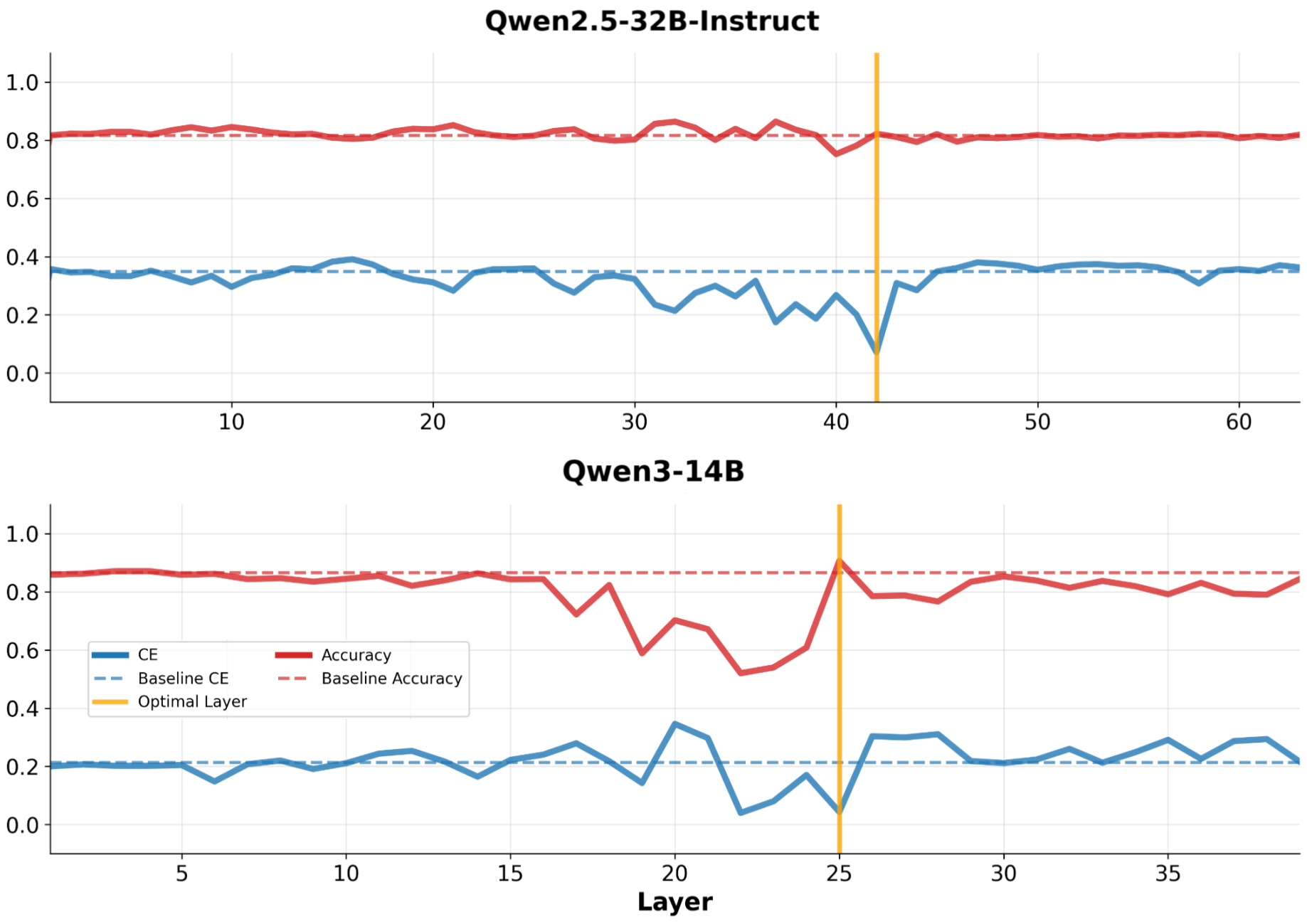}
    \caption{\textbf{Bias mitigation.} Per-layer accuracy (red) and content effect (blue) of zero-shot \texttt{Qwen2.5-32B-Instruct} and \texttt{Qwen3-14B} on the logical validity classification task after adding the task difference steering vector $\mu_{V-P}^l$ multiplied by a scalar value $\alpha = 1.5$ at different layers. For comparison, original accuracy and content effect are shown as dashed lines. The orange line indicates the layer that best retains (or improves) the original accuracy while lowering the content effect.}
    \label{fig:mitigation}
\end{figure}

\subsection{Can We Design an Intervention to Mitigate Content Effects?}
\label{sec:mitigation}

As a final experiment, we test whether the content effect can be mitigated by explicitly disentangling the representations of validity and plausibility. Motivated by our earlier analyses, we construct a steering vector that captures the difference between task-level activations for logical validity classification and plausibility classification. Let $P$ denote the plausibility classification dataset with labels $\{\mathrm{true}, \mathrm{false}\}$ and $V$ the logical validity classification dataset with labels $\{\mathrm{valid}, \mathrm{invalid}\}$. For each layer $l$, we compute the mean hidden activations $\mu_P^l$ and $\mu_V^l$ across all examples in $P$ and $V$, respectively. We then define the \emph{task-difference vector}:

$$
\mu_{V-P}^l = \mu_V^l - \mu_P^l
$$

\noindent This vector should isolate the representational dimensions specific to arguments’ validity, excluding those tied to conclusions’ plausibility. When added to hidden states during validity classification, $\mu_{V-P}^l$ should push the representation away from plausibility-sensitive directions, thereby reducing the influence of content effects.

Figure~\ref{fig:mitigation} reports results of this intervention applied to \texttt{Qwen2.5-32B-Instruct} and \texttt{Qwen3-14B} in the zero-shot setting, where content effects are strongest. For both models, we identify a $\mu_{V-P}^l$ that improves overall accuracy while reducing CE. Specifically, we use layer 43 for \texttt{Qwen2.5-32B-Instruct} and layer 26 for \texttt{Qwen3-14B}. Unlike in the previous experiment, we found that simply adding this vector was insufficient to mitigate the bias. Better results were obtained by first scaling $\mu_{V-P}^l$ with a factor $\alpha = 1.5$. Further details on the choice of $\alpha$ are provided in Appendix \ref{app:full_mitigation}. In \texttt{Qwen2.5-32B-Instruct}, accuracy increases from $81.62$ to $82.21$ and CE drops from $0.348$ to $0.072$. For \texttt{Qwen3-14B}, accuracy rises from $86.54$ to $96.70$, with CE reduced from $0.213$ to $0.043$. These results show that this simple intervention can render models nearly unbiased ($CE \approx 0$) in their validity judgments.

Our debiasing intervention differs from prior work in both motivation and design. \citet{bertolazzi-etal-2024-systematic} address content effects through fine-tuning on pseudo-word vocabularies, requiring retraining on carefully constructed synthetic data. Most closely related, \citet{valentino2025mitigatingcontenteffectsreasoning} apply activation steering at inference time using contrastive vectors derived from behaviorally-defined pairs: activations leading to correct versus content-biased predictions. While effective, this method steers toward logically correct behavior without providing an account of \textit{why} the bias exists. Our intervention, by contrast, follows directly from our interpretability analysis: having established that validity and plausibility are geometrically entangled and that this entanglement predicts content effects, the task-difference vector $\mu_{V-P}^l$ explicitly disentangles these two concepts rather than steering toward an externally defined notion of correctness. The success of this intervention thus serves as additional evidence for our theoretical account, demonstrating how interpretability can provide actionable insights for improving reasoning.

\section{Conclusion}
In this work, we provide new evidence regarding the emergence of content effects in LLMs performing logical reasoning tasks. By analyzing internal representations, we show that plausibility and validity are not only linearly encoded but also strongly aligned in the representational geometry of LLMs. This alignment predicts the extent to which models conflate plausibility with validity, producing systematic reasoning biases analogous to content effects in humans. Moreover, we demonstrate that plausibility vectors can causally influence validity judgments, and vice versa, and we propose a simple intervention that disentangles these concepts, reducing bias and improving accuracy without requiring parameter updates.

Taken together, these findings advance our understanding of how abstract logical concepts are encoded in LLMs. Beyond explaining a specific bias, our results highlight the usefulness of representational analyses for both diagnosing and mitigating systematic errors in LLM reasoning. Future work can extend this framework to other cognitive biases, investigate whether similar mechanisms underlie them, and explore how disentangled representations might support more reliable and trustworthy reasoning in models.

\section*{Limitations}
To the best of our knowledge, our work is the first to adopt interpretability techniques to explore how content effects emerge in LLMs. Nevertheless, our analysis has some limitations. First, we focus only on dense representations extracted from models' residual streams at the last token position before a model generates a validity or plausibility judgement. Future work could extend this analysis by investigating how representations of validity and plausibility develop throughout the entire sequence of tokens generated by models, and by employing methods to extract sparser features activated by models. This research direction appears particularly promising for explaining why CoT models show lower content effects than zero-shot models, despite exhibiting similar patterns in the aspects we investigated: plausibility and validity vectors show high similarity, this similarity predicts behavioral content effects, and we observe causal cross-influence between the two concepts.

Second, while we attribute behavioral content effects to the entanglement of validity and plausibility representations, we do not provide evidence explaining why models conflate these concepts. We conjecture that this conflation stems from properties of the training data. In most texts containing logical arguments, valid arguments are also sound, meaning their premises are true. Consequently, models may lack sufficient exposure during training to valid arguments with false premises, leading them to associate validity with plausibility. This hypothesis suggests a promising direction for future work: investigating the relationship between how these concepts are represented within LLMs and the statistical properties of their training data.

\section*{Acknowledgments}

We thank the members of the Dialogue Modelling Group (DMG) and the Multimodality, Language, and Interpretability (Mulini) Lab from the University of Amsterdam for their valuable feedback and stimulating discussions during LB’s visit to UvA. In particular, we thank Vera Neplenbroek and Michael Hanna for their helpful comments on an early version of this work.

\bibliography{custom}

\appendix

\section{Syllogisms}
\label{app:syllogisms}

Syllogisms are deductive arguments that consist of two premises and a conclusion. Each statement in a syllogism relates two terms (or predicates) using a quantifier and follows the structure \texttt{``Quantifier X are Y''} or \texttt{``Quantifier X are not Y''}. The quantifiers used in classical syllogistic logic are ``All,'' ``Some,'' ``No,'' and ``Some... not.''
In any syllogism, exactly three distinct terms appear: which we denote as $A$, $B$, and $C$. The term $B$ serves as the ``middle term'' that appears in both premises but not in the conclusion. Specifically, the first premise relates terms $A$ and $B$, the second premise relates terms $B$ and $C$, and the conclusion relates terms $A$ and $C$. For example, in the classic syllogism ``All humans are mortal; Socrates is human; therefore, Socrates is mortal,'' the terms would be ``Socrates'' ($A$), ``human'' ($B$), and ``mortal'' ($C$).

The logical form of a syllogism depends on several structural features: (1) which quantifier is used in each premise and the conclusion, (2) whether negation is present in each statement, and (3) the order in which the terms $A$, $B$, and $C$ are related across the premises (known as the ``figure'' of the syllogism). When we enumerate all possible combinations of these features, we obtain the 64 distinct combinations of premises of classical syllogistic logic. Of these 64 possible premise pairs, only 27 yield a logically valid conclusion, while the remaining 37 have no valid conclusion.

These syllogistic arguments can be formally represented using first-order logic, which provides a rigorous framework for defining which syllogisms are valid or invalid. It is important to note that the specific count of valid and invalid syllogisms used in our analysis depends on two key assumptions: (a) we consider all terms to denote non-empty sets (i.e., we assume that the categories referenced by our terms contain at least one member), and (b) we allow for conclusions that relate the terms $A$ and $C$ in either ordering: both $A-C$ and $C-A$.

\section{Dataset and Prompt Details}
\label{app:data}

Our experiments build on the syllogistic inference dataset introduced by \citet{bertolazzi-etal-2024-systematic}, which systematically examines content effects in a multiple-choice setting. There are 64 possible combinations of premises in classical syllogistic logic. Since this dataset uses ten distinct triples of terms, such as ``labradors $\rightarrow$ dogs $\rightarrow$ canines'' (see Figure \ref{fig:additional_prompts}), there are a total of 640 plausible syllogisms and 640 implausible syllogisms. To convert the original multiple-choice format into an NLI-style classification setup, we randomly sampled one valid conclusion for each valid syllogism and one invalid conclusion for each invalid syllogism. In this dataset, plausible syllogisms have conclusions that align with real-world knowledge, whereas implausible syllogisms contradict it. This procedure yielded a dataset of 1,280 syllogisms, approximately 42\% valid and 58\% invalid, reflecting the distribution across the 64 types. Train-test splits were created using a stratified 70-30 partition, ensuring that all 64 syllogism types were represented in both splits.

For the plausibility classification task, we extracted all unique conclusions that appeared as multiple-choice options in the original 1,280 syllogisms \citep{bertolazzi-etal-2024-systematic}, resulting in 1,056 distinct statements.

Logical validity and plausibility classification employed both zero-shot and CoT prompting formats. Figure~\ref{fig:prompt_comparison} shows the prompts used with these datasets in both formats. Additionally, for validity classification, we used two extra prompts in the mixed-effects linear regression experiment described in Section \ref{sec:regression}. Figures~\ref{fig:logical_variants} show these additional prompts.

For the ``thinking'' models from the Qwen-3 family, we appended the instruction ``Keep your thinking concise, avoid over-explaining, and reach a solution efficiently'' to reduce reasoning effort and limit the computational requirements during inference.

Figure \ref{fig:additional_prompts} shows the prompts used for the control datasets. The first is a hypernym-hyponym classification dataset using the same ten term triples. All 60 possible source-target pairs were considered, and three prompt augmentations per pair resulted in 180 examples. Models were tasked with determining whether the source term is a hypernym or a hyponym of the target. The second control dataset is a harmful vs. harmless content classification dataset comprising 916 examples drawn from \citet{arditi2025refusal}, balanced evenly between harmful and harmless prompts, which models had to classify accordingly.

\section{Implementation Details}

All experiments were implemented in PyTorch \citep{pytorch}, leveraging the Transformers \citep{transformers} library to load and interact with the models. Models were loaded in \texttt{bfloat16} precision. During evaluation, we used greedy decoding for all models except those in the Qwen-3 family, for which we followed the sampling parameters recommended in the Hugging Face model card. To ensure full reproducibility, all random number generators were seeded with the fixed value of 128.

During the project’s source code development, GitHub Copilot was used as an assistant tool, and ChatGPT was employed to correct minor grammatical errors within this document.

\section{Additional Results}
\label{app:additional_results}

\subsection{Behavioral Evaluation}
\label{app:full_behavioral}

We report extended results for all models on the logical validity classification task. Tables~\ref{tab:behavioral-full} and~\ref{tab:behavioral-variants} provide accuracy by validity--plausibility subsets as well as the content effect (CE) metric (see Section~\ref{sec:metrics} for the definition of CE).

Table~\ref{tab:behavioral-full} presents results obtained on all ten models, showing zero-shot and CoT settings for each model. In the zero-shot condition, all models exhibit non-zero CE. The strongest effects are observed in the Qwen-2.5 family: \texttt{Qwen2.5-7B-Instruct} ($CE=0.418$), \texttt{Qwen2.5-14B-Instruct} ($\mathrm{CE}=0.361$), and \texttt{Qwen2.5-32B-Instruct} ($\mathrm{CE}=0.348$). By contrast, Qwen-3 and Gemma-3 models achieve both higher accuracy and reduced CE, with zero-shot accuracies ranging from $80.61\%$ (\texttt{Qwen3-4B}) to $90.91\%$ (\texttt{Qwen3-32B}) and CE values between $0.063$ and $0.218$. The Gemma-3 series shows a similar trend, with accuracies between $81.02\%$ and $87.29\%$ and CE ranging from $0.129$ to $0.213$.  

In the CoT setting, we observe systematically lower CE across all models, ranging from $-0.006$ (\texttt{Gemma3-12B-it}) to $0.147$ (\texttt{Qwen2.5-7B-Instruct}), compared to zero-shot values between $0.063$ and $0.418$. At the same time, accuracies increase substantially, with all models exceeding $89\%$ under CoT prompting and several reaching above $97\%$ (\texttt{Qwen3-14B}, \texttt{Qwen3-4B}, \texttt{Gemma3-27B-it}).  

Table~\ref{tab:behavioral-variants} reports means and standard deviations across three prompt variants for each model. These are the same prompts used in the mixed-effects regression analysis in Section~\ref{sec:regression}. The overall pattern remains consistent: zero-shot prompts produce higher CE (up to $0.430 \pm 0.009$ for \texttt{Qwen2.5-7B-Instruct}), while CoT prompts reduce content effects and raise accuracy. Some smaller models (\texttt{Qwen2.5-7B-Instruct}, \texttt{Qwen3-4B}) show larger variance across prompts, but the direction of improvement under CoT holds across the model set.

\subsection{Low-dimensional Visualization}
\label{app:visualization}
We provide here a more extensive visualization of the hidden states from \texttt{Qwen2.5-32B-Instruct} during validity and plausibility classification tasks using 3D PCA. Data points in the plots are labeled according to model predictions, reflecting our focus on understanding how the model internally represents what it believes to be valid/invalid or true/false.

\begin{figure*}[htbp]
    \centering

    \begin{subfigure}[b]{0.25\textwidth}
        \includegraphics[width=\textwidth]{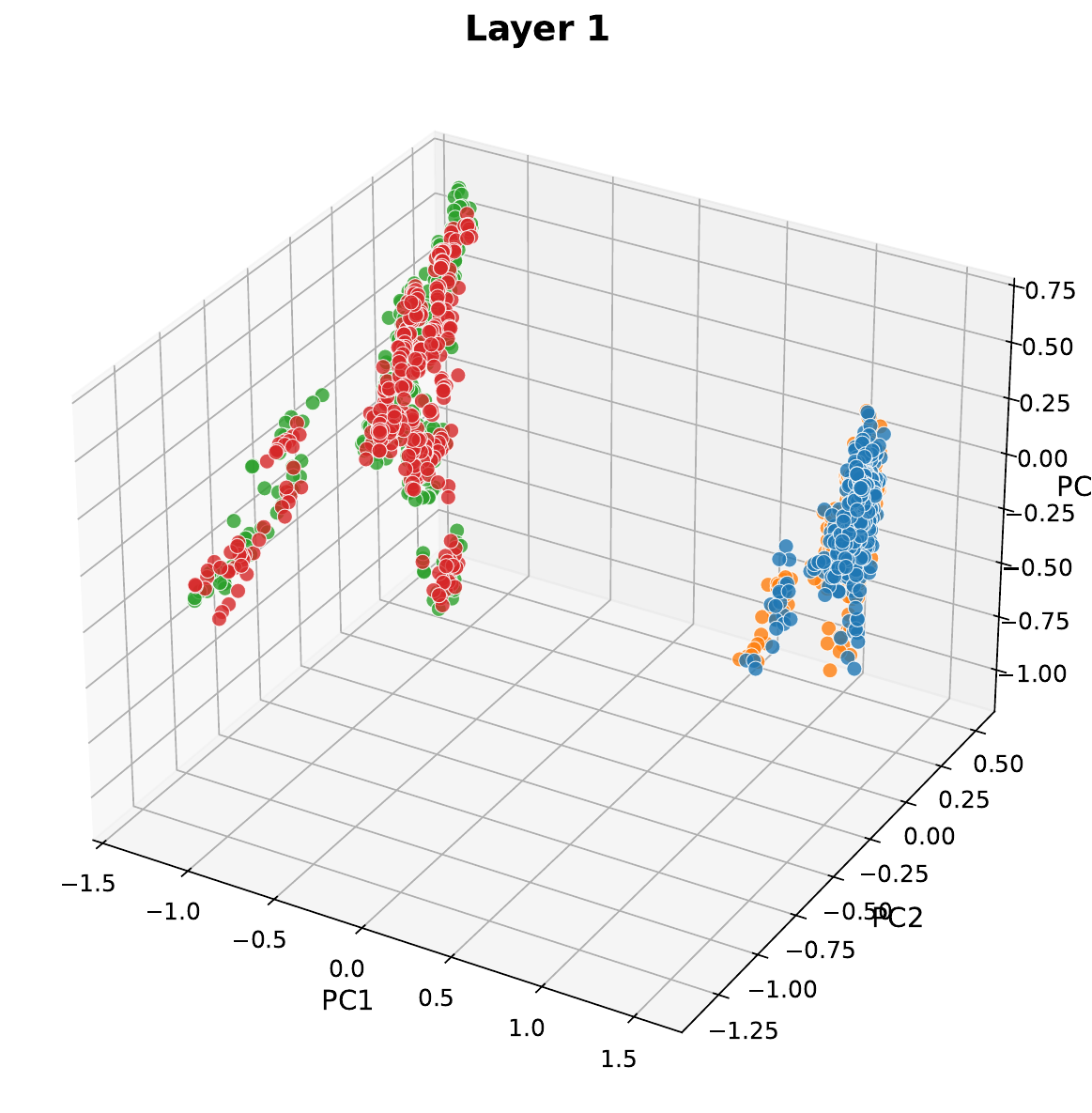}
    \end{subfigure}
    \begin{subfigure}[b]{0.25\textwidth}
        \includegraphics[width=\textwidth]{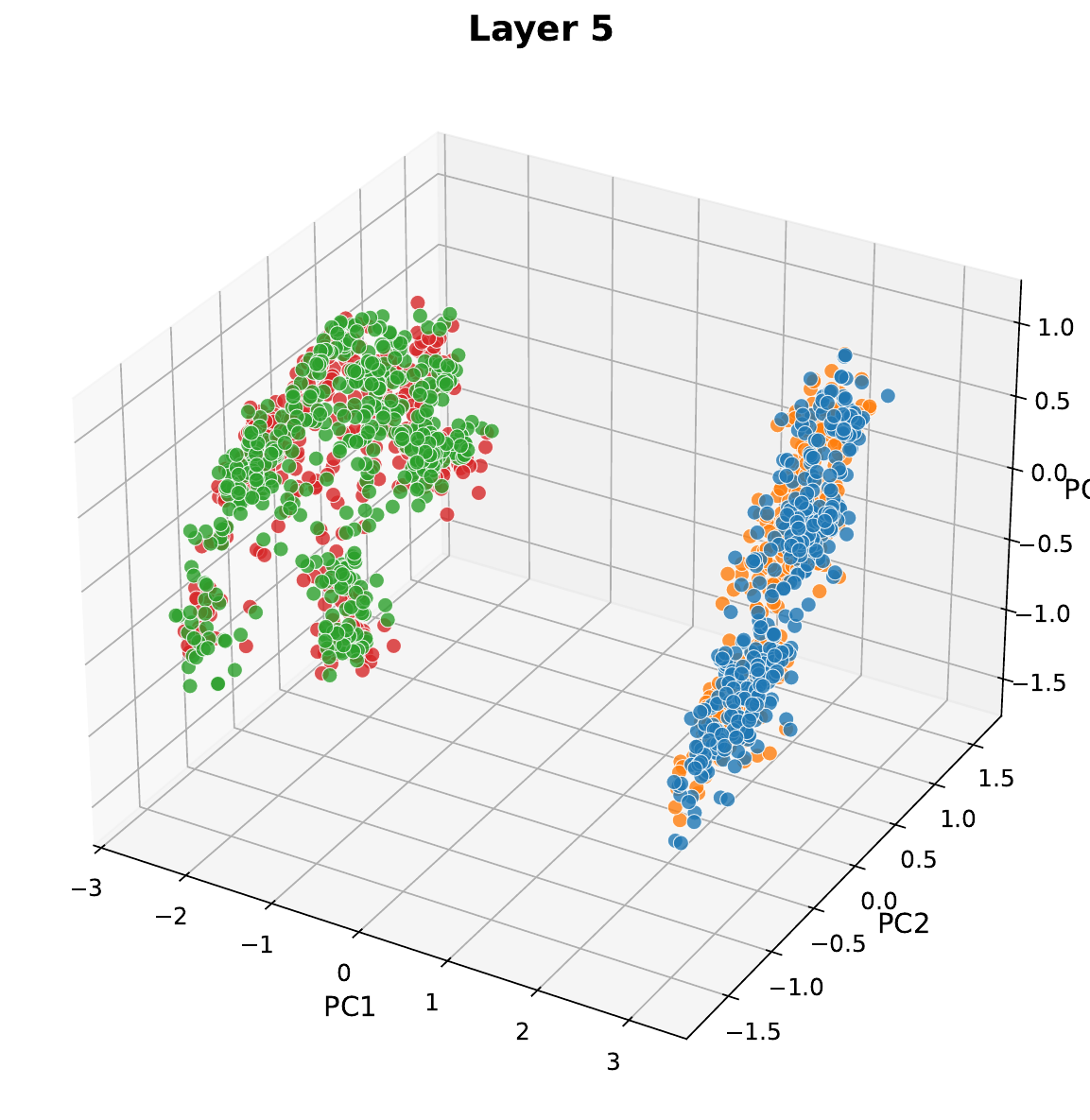}
    \end{subfigure}
    \begin{subfigure}[b]{0.25\textwidth}
        \includegraphics[width=\textwidth]{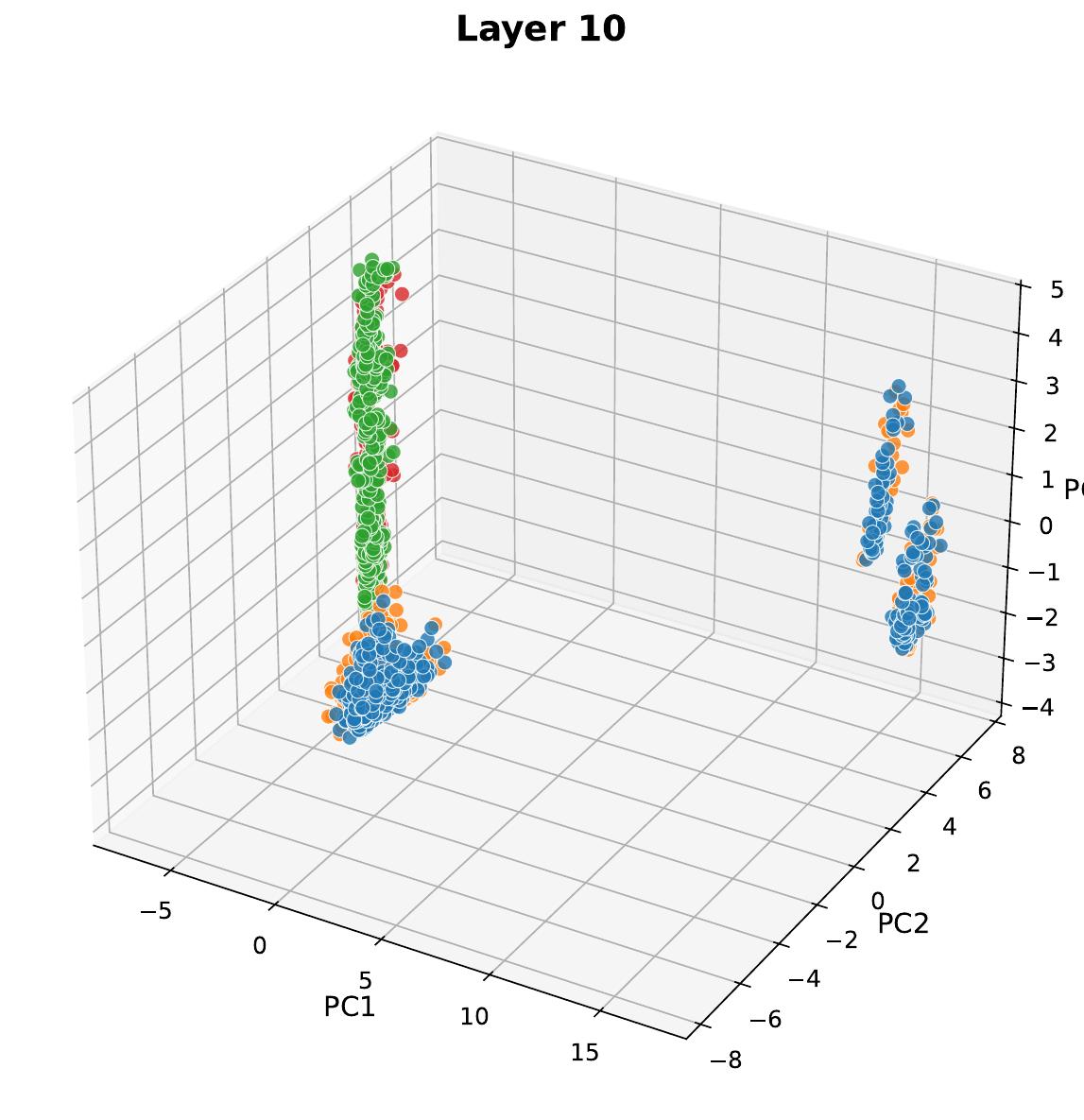}
    \end{subfigure}

    \begin{subfigure}[b]{0.25\textwidth}
        \includegraphics[width=\textwidth]{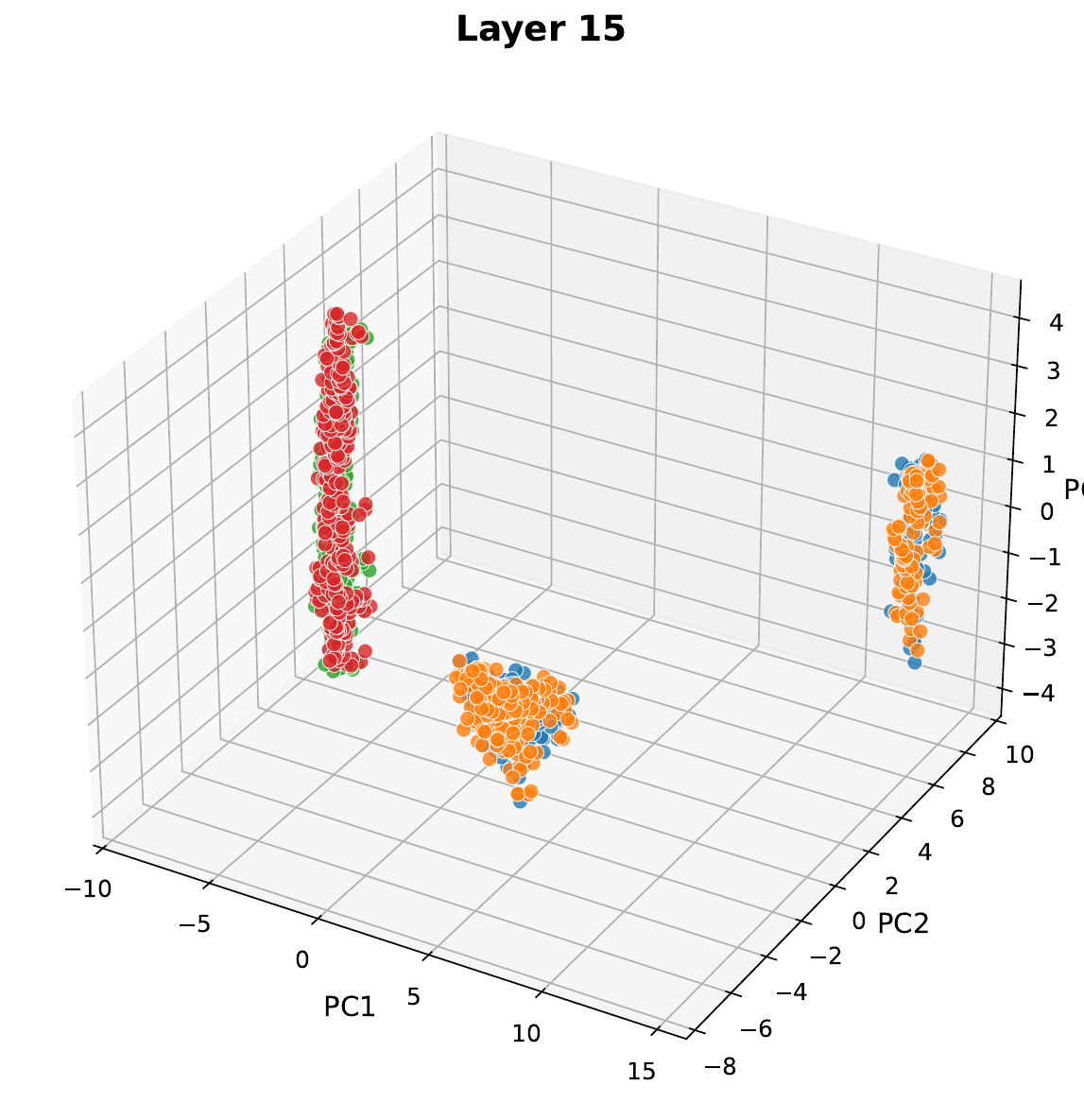}
    \end{subfigure}
    \begin{subfigure}[b]{0.25\textwidth}
        \includegraphics[width=\textwidth]{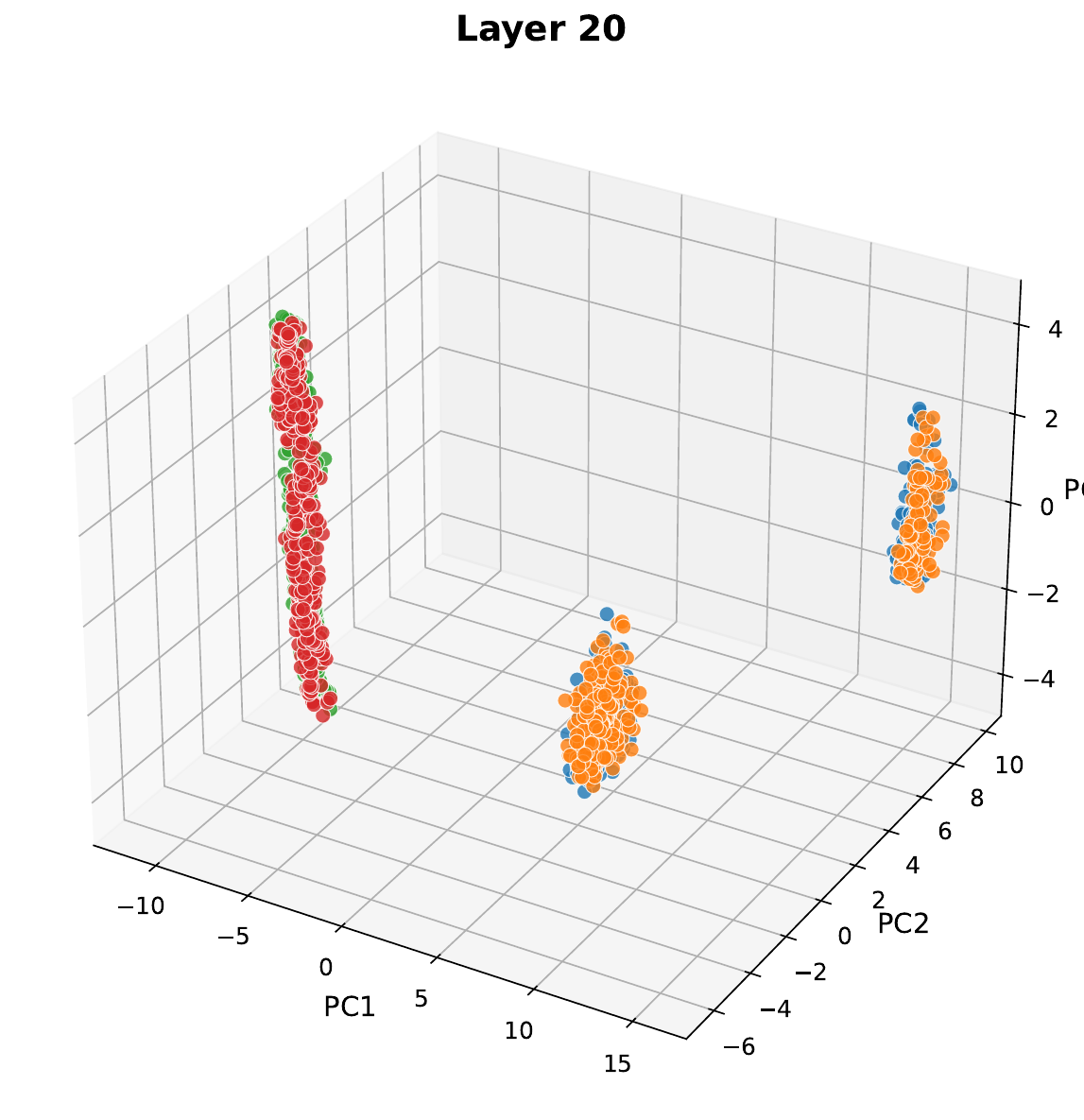}
    \end{subfigure}
    \begin{subfigure}[b]{0.25\textwidth}
        \includegraphics[width=\textwidth]{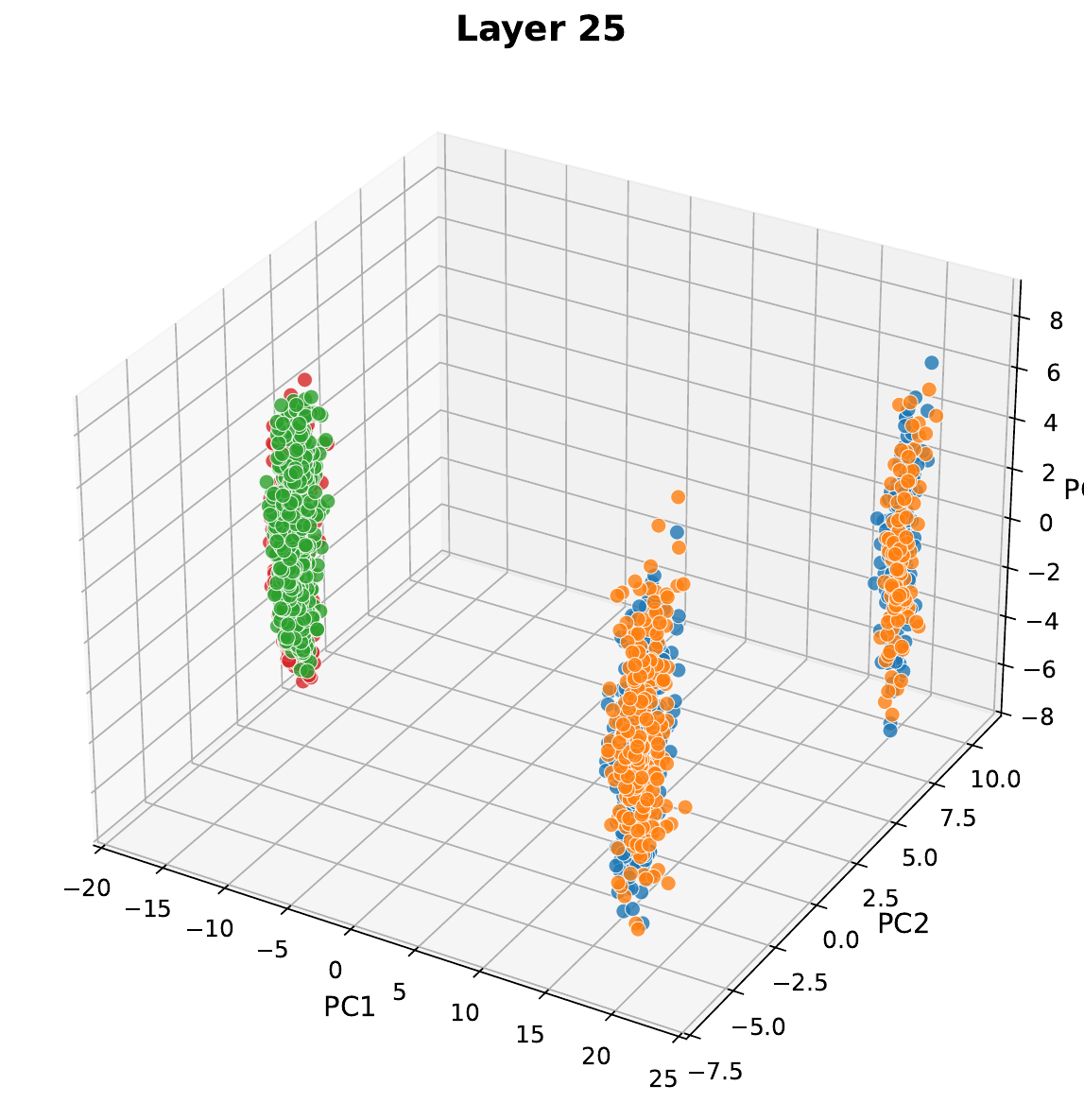}
    \end{subfigure}

    \begin{subfigure}[b]{0.25\textwidth}
        \includegraphics[width=\textwidth]{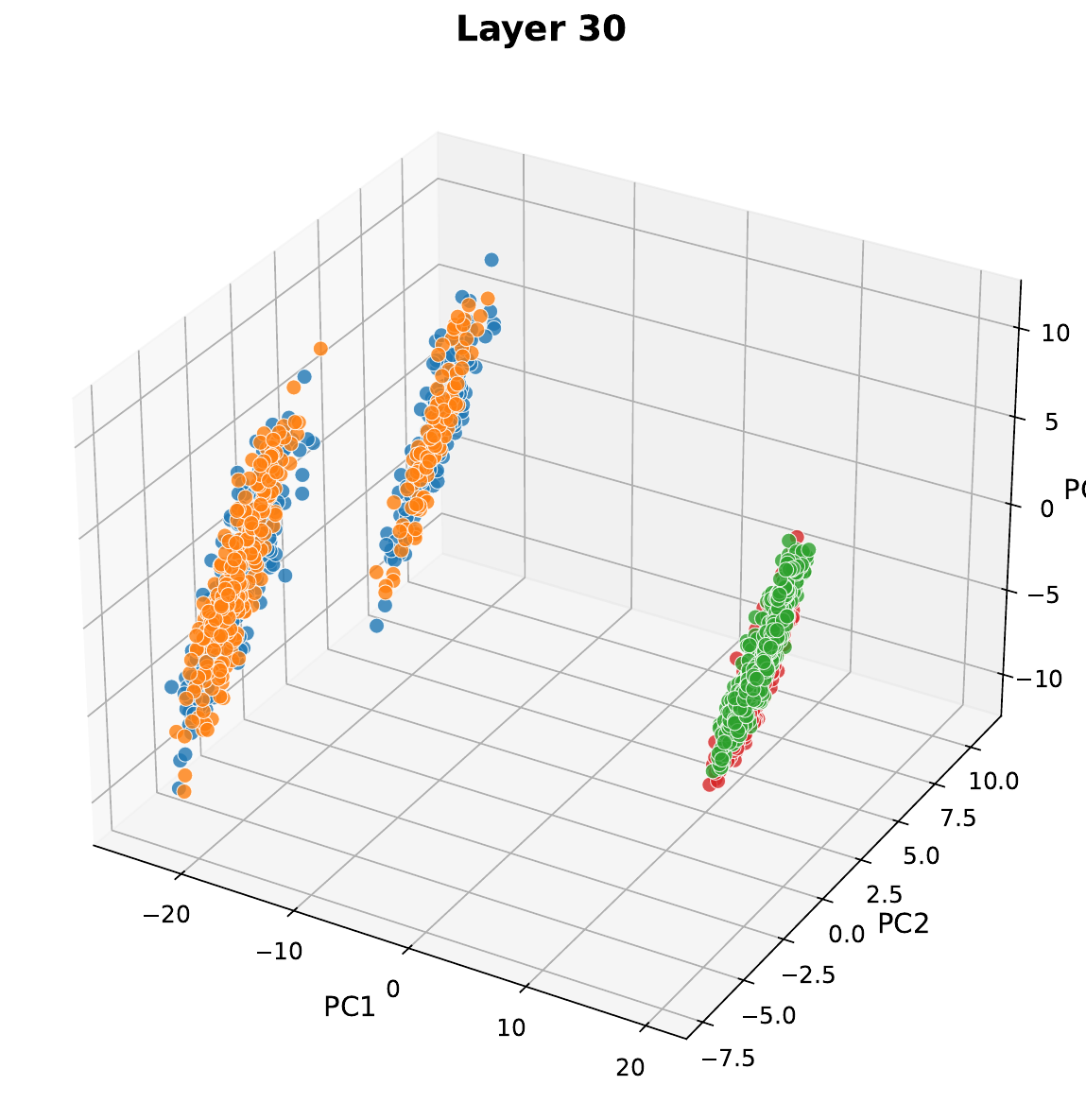}
    \end{subfigure}
    \begin{subfigure}[b]{0.25\textwidth}
        \includegraphics[width=\textwidth]{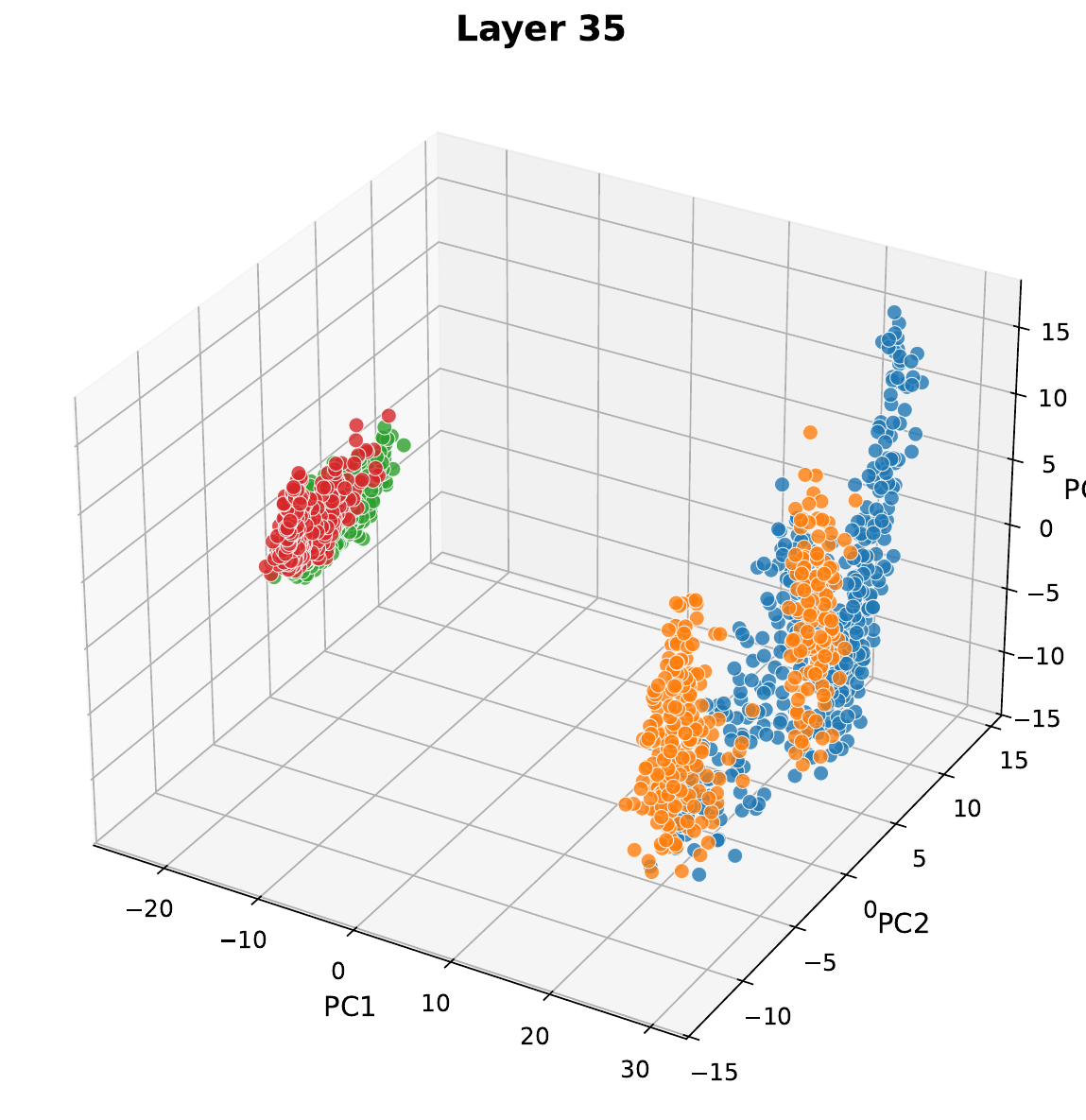}
    \end{subfigure}
    \begin{subfigure}[b]{0.25\textwidth}
        \includegraphics[width=\textwidth]{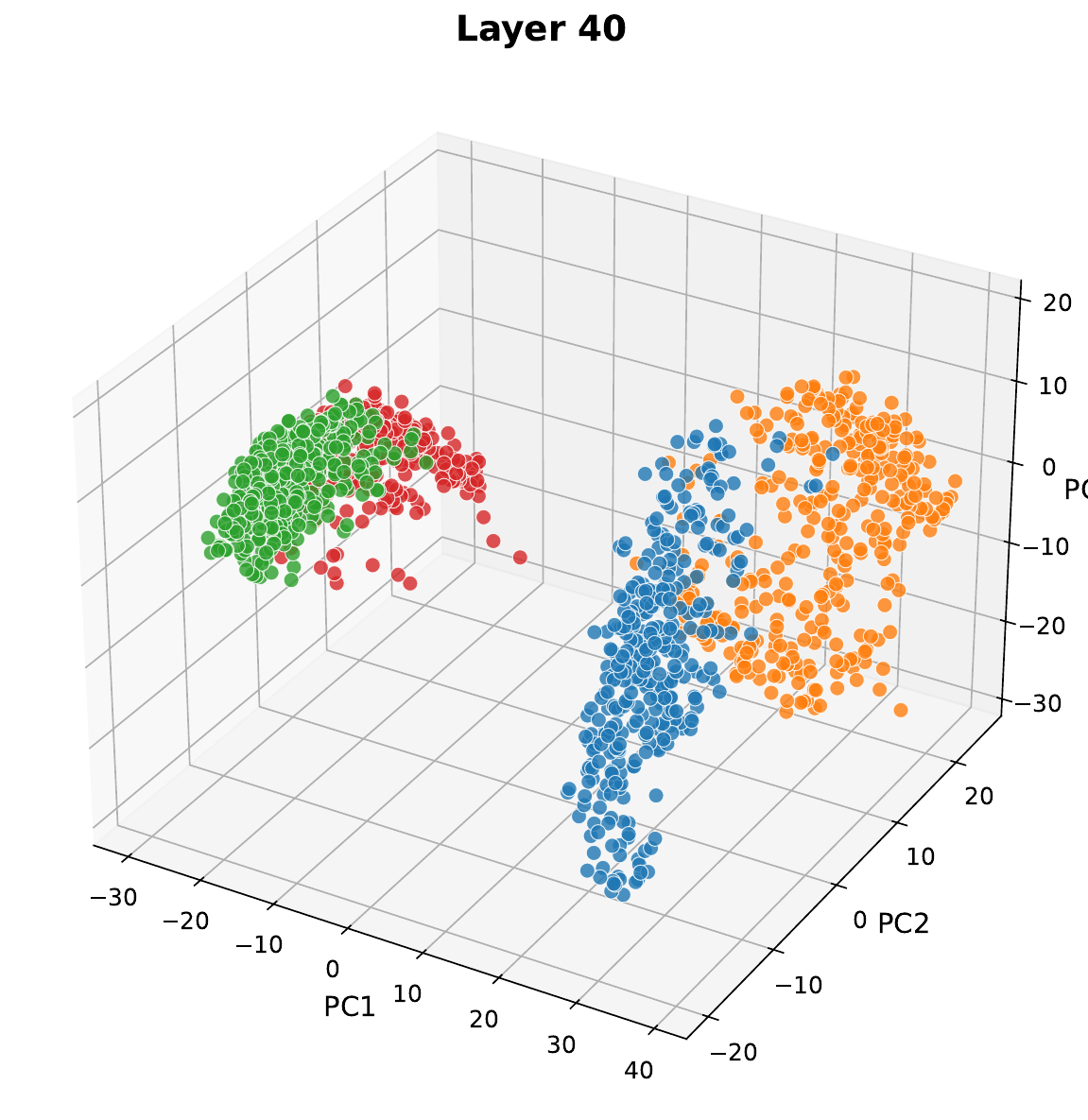}
    \end{subfigure}

    \begin{subfigure}[b]{0.25\textwidth}
        \includegraphics[width=\textwidth]{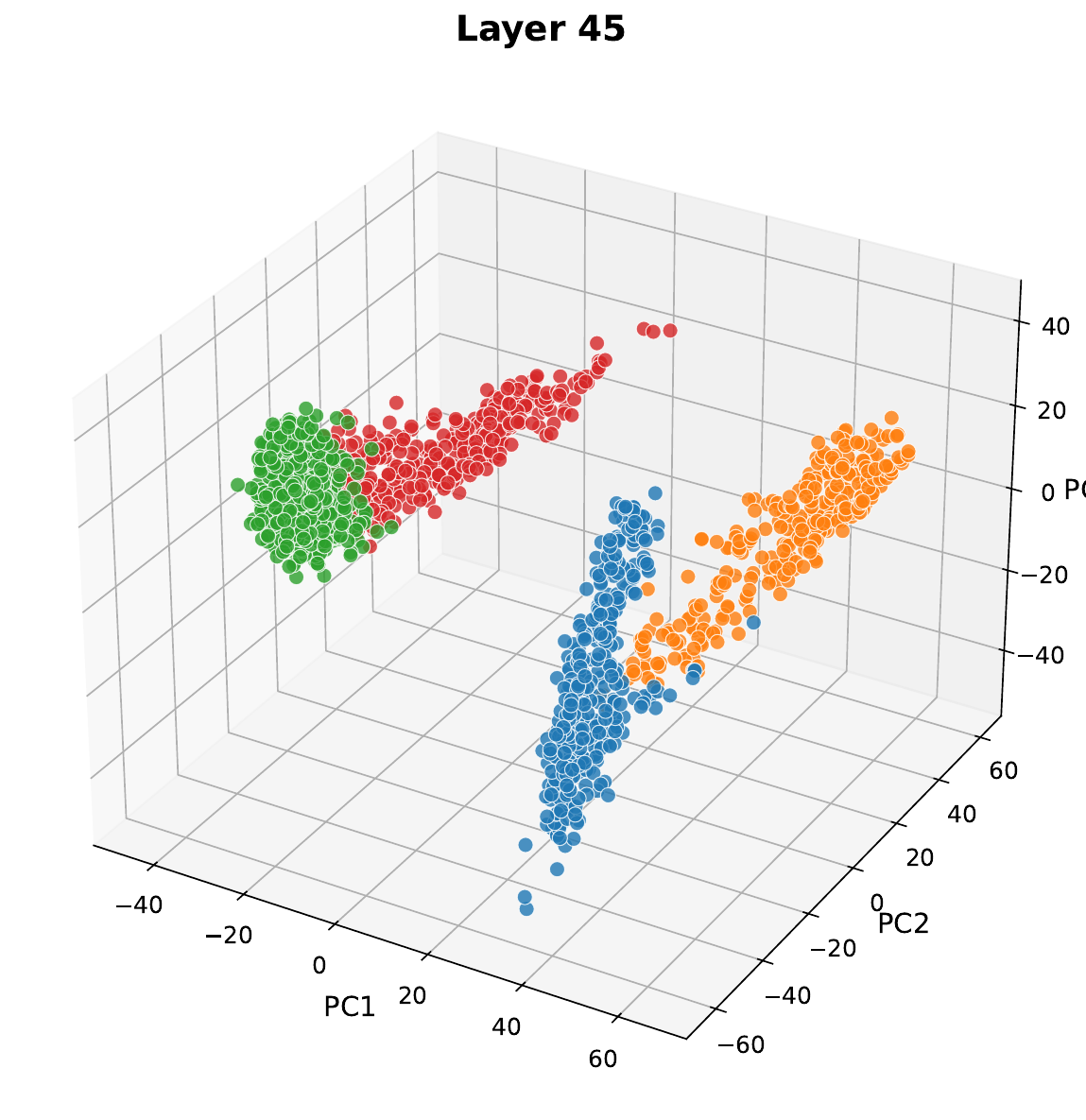}
    \end{subfigure}
    \begin{subfigure}[b]{0.25\textwidth}
        \includegraphics[width=\textwidth]{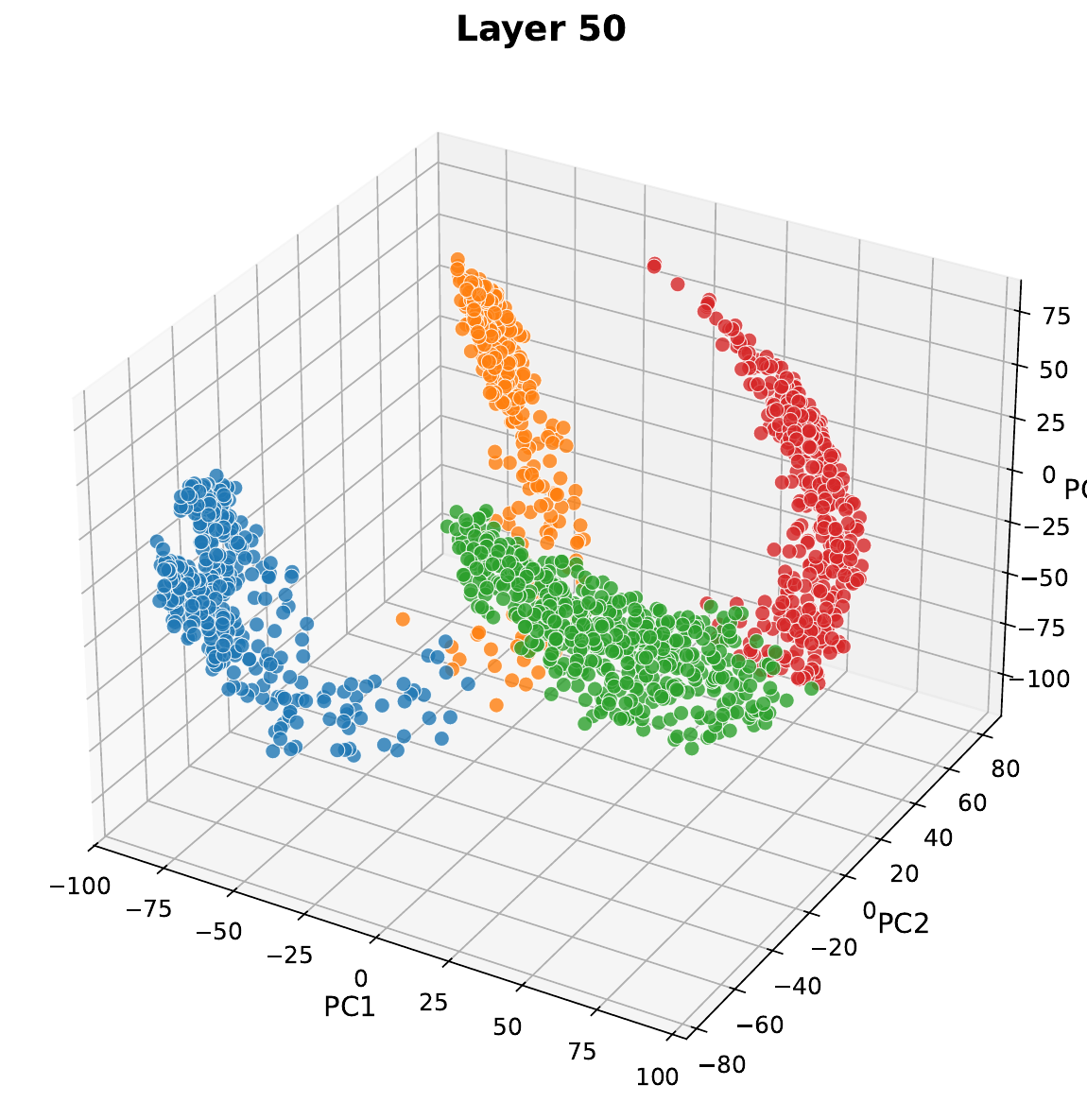}
    \end{subfigure}
    \begin{subfigure}[b]{0.25\textwidth}
        \includegraphics[width=\textwidth]{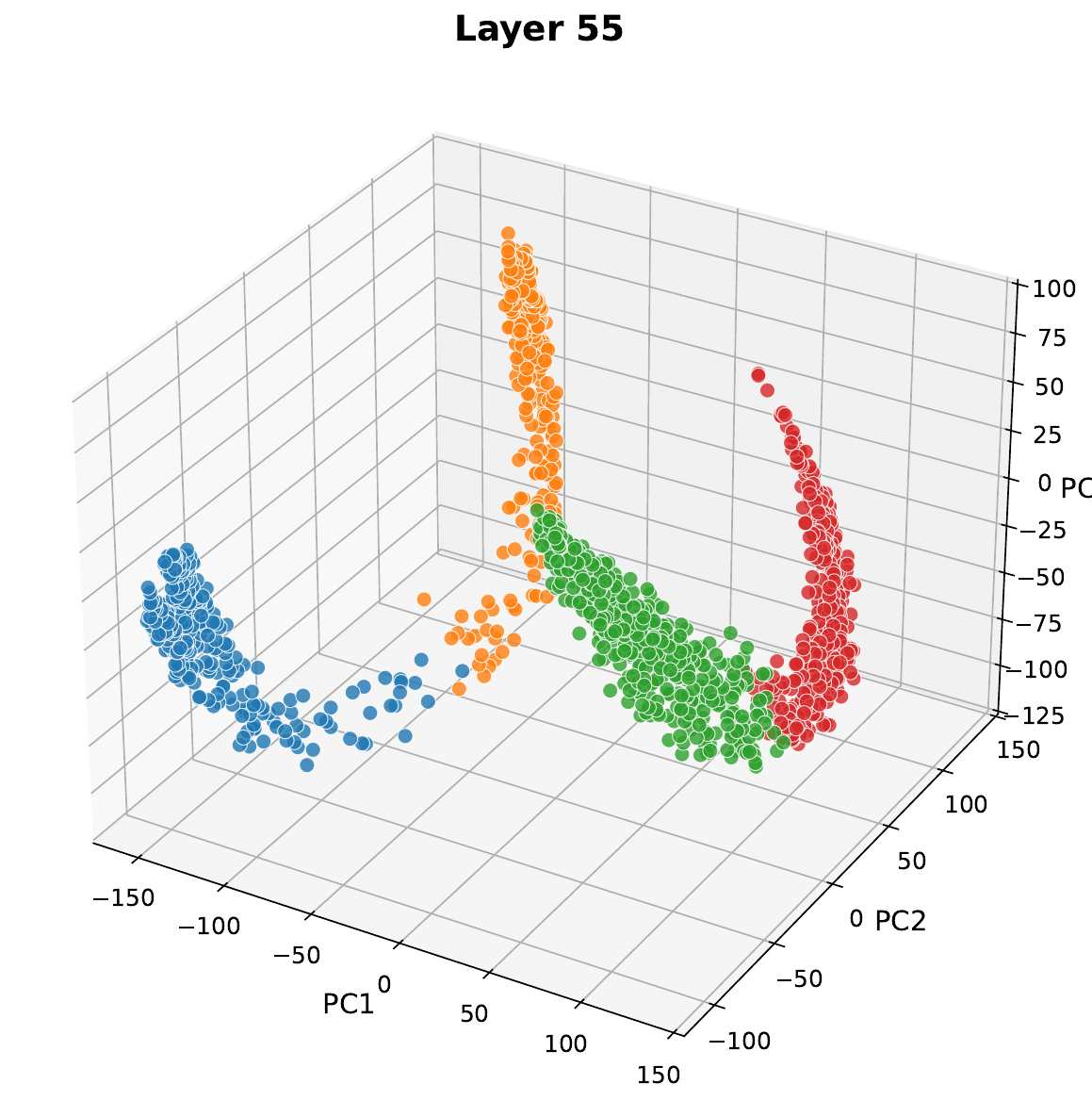}
    \end{subfigure}

    \begin{subfigure}[b]{0.25\textwidth}
        \includegraphics[width=\textwidth]{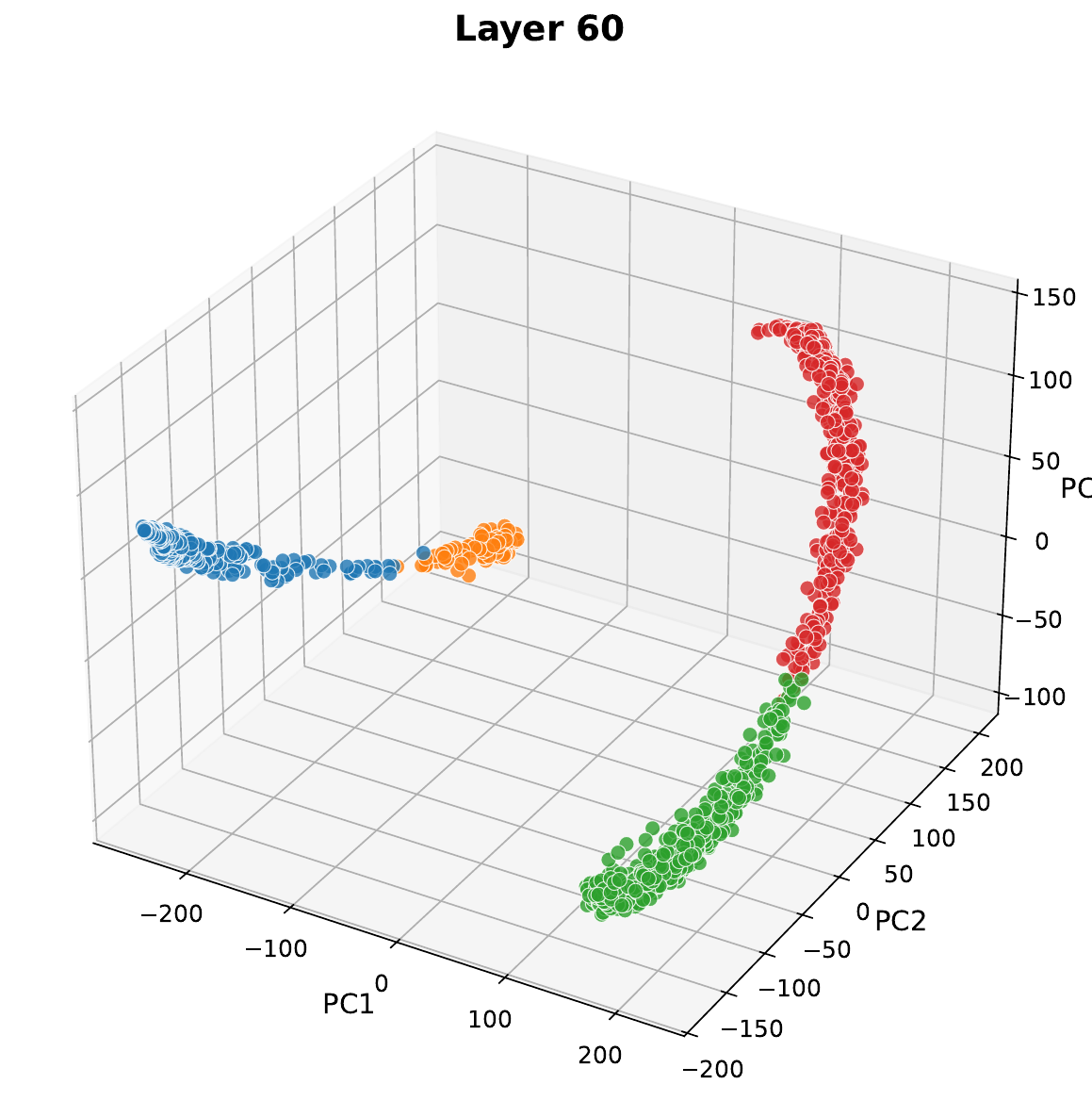}
    \end{subfigure}
    \begin{subfigure}[b]{0.25\textwidth}
        \includegraphics[width=\textwidth]{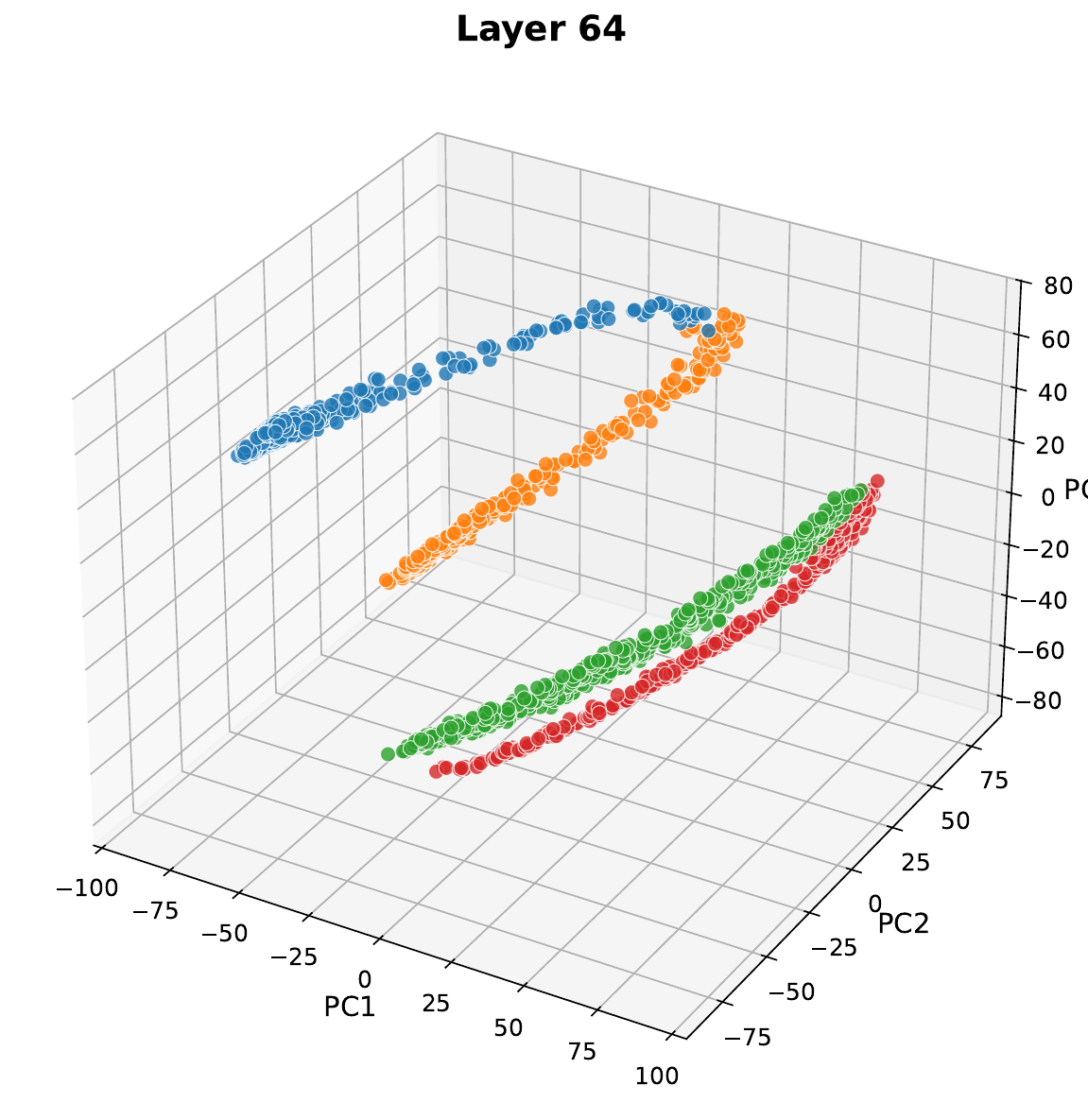}
    \end{subfigure}
    \begin{subfigure}[b]{0.25\textwidth}
        \centering
        \includegraphics[width=0.6\textwidth]{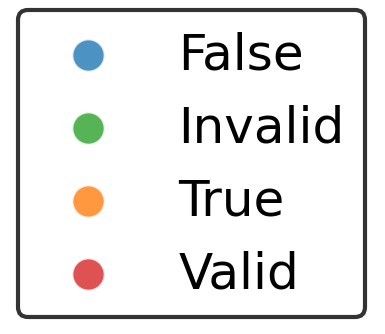}
        \vspace{1cm}
    \end{subfigure}

\caption{3D PCA projections of logical validity and plausibility classification prompts for \texttt{Qwen2.5-32B-Instruct} in the zero-shot setting. The projections illustrate how the model forms distinct clusters for validity and plausibility classes across layers.}
\label{fig:pca-zero}
\end{figure*}

\begin{figure*}[htbp]
    \centering

    \begin{subfigure}[b]{0.25\textwidth}
        \includegraphics[width=\textwidth]{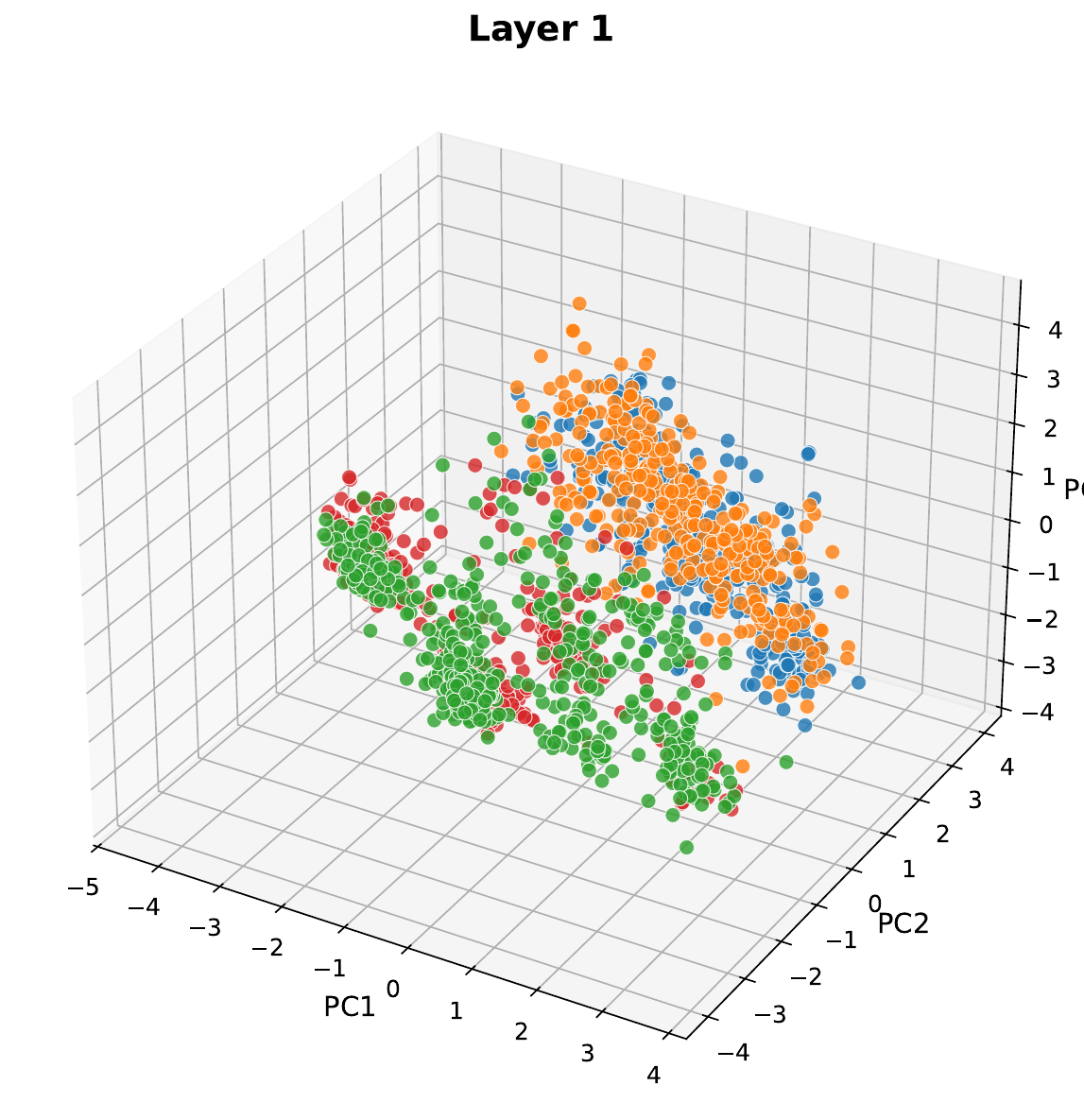}
    \end{subfigure}
    \begin{subfigure}[b]{0.25\textwidth}
        \includegraphics[width=\textwidth]{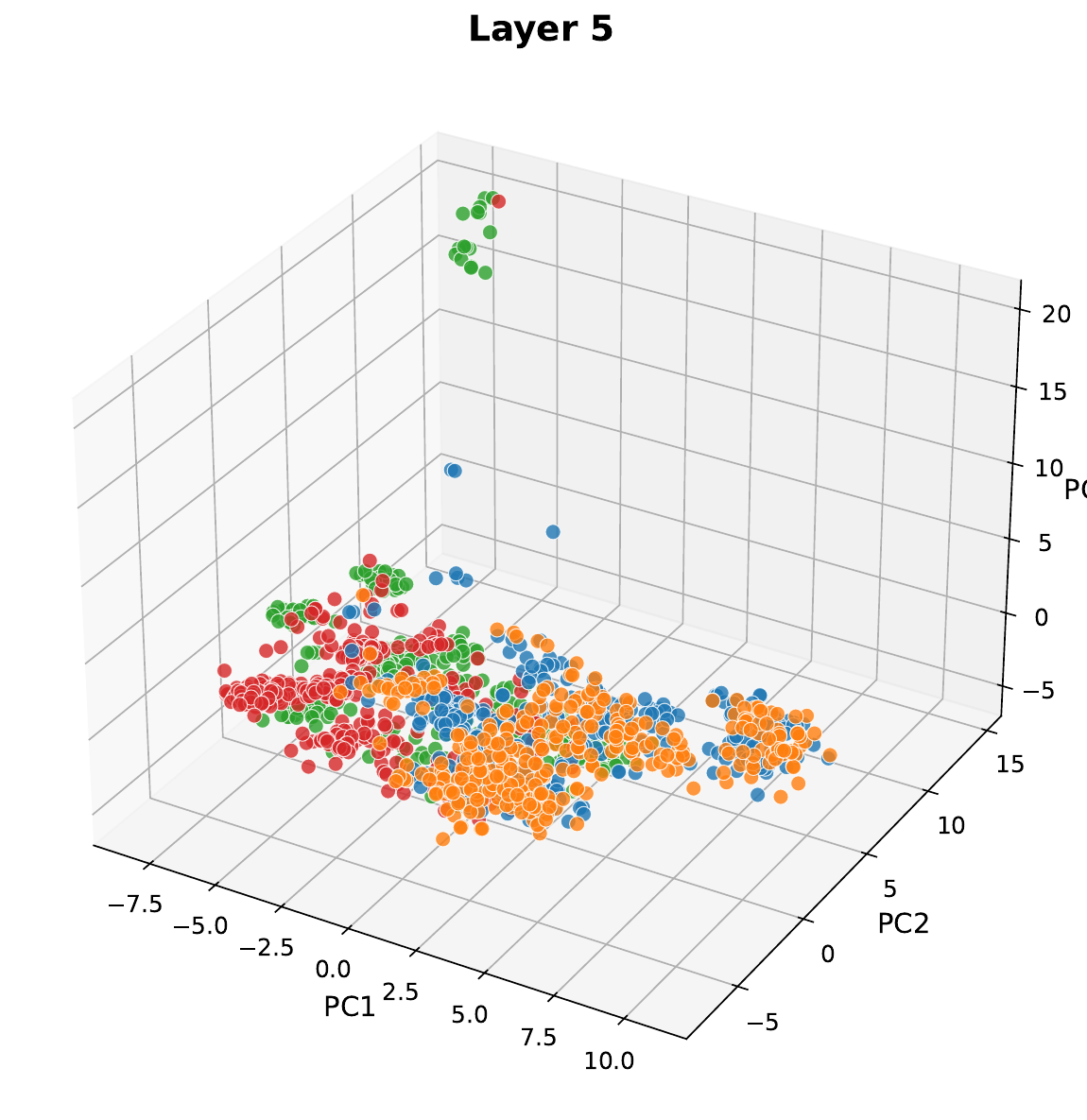}
    \end{subfigure}
    \begin{subfigure}[b]{0.25\textwidth}
        \includegraphics[width=\textwidth]{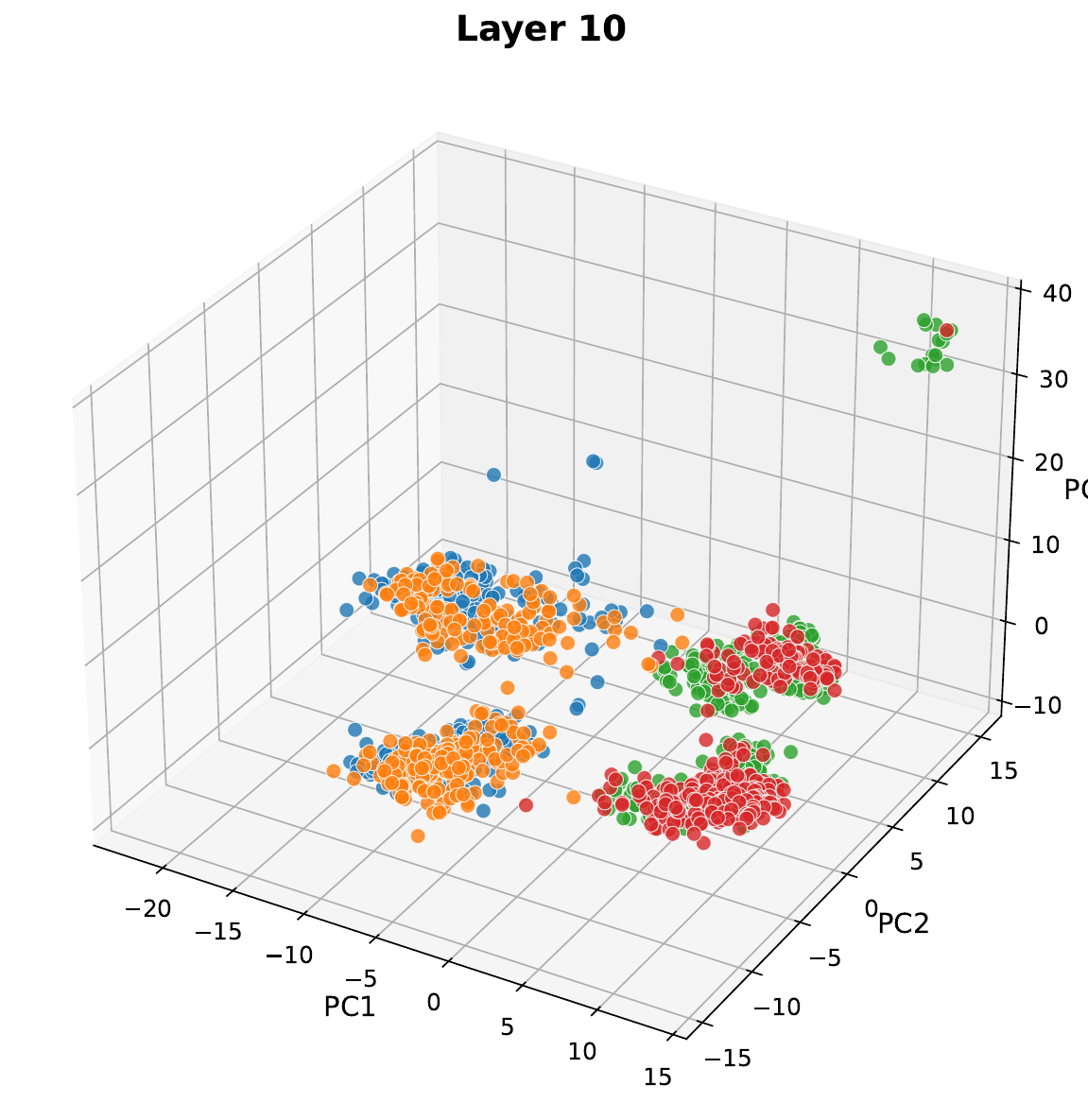}
    \end{subfigure}

    \begin{subfigure}[b]{0.25\textwidth}
        \includegraphics[width=\textwidth]{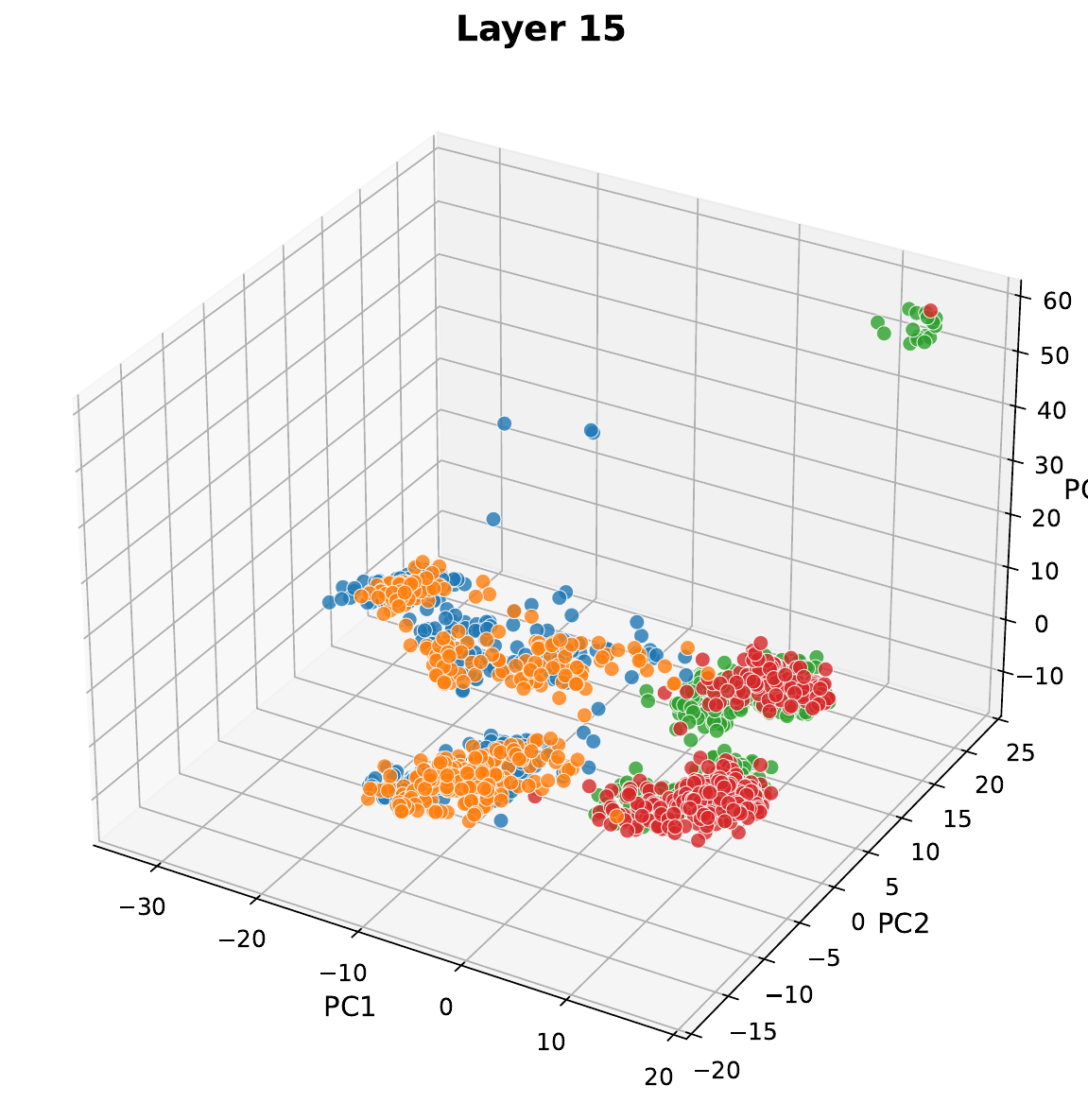}
    \end{subfigure}
    \begin{subfigure}[b]{0.25\textwidth}
        \includegraphics[width=\textwidth]{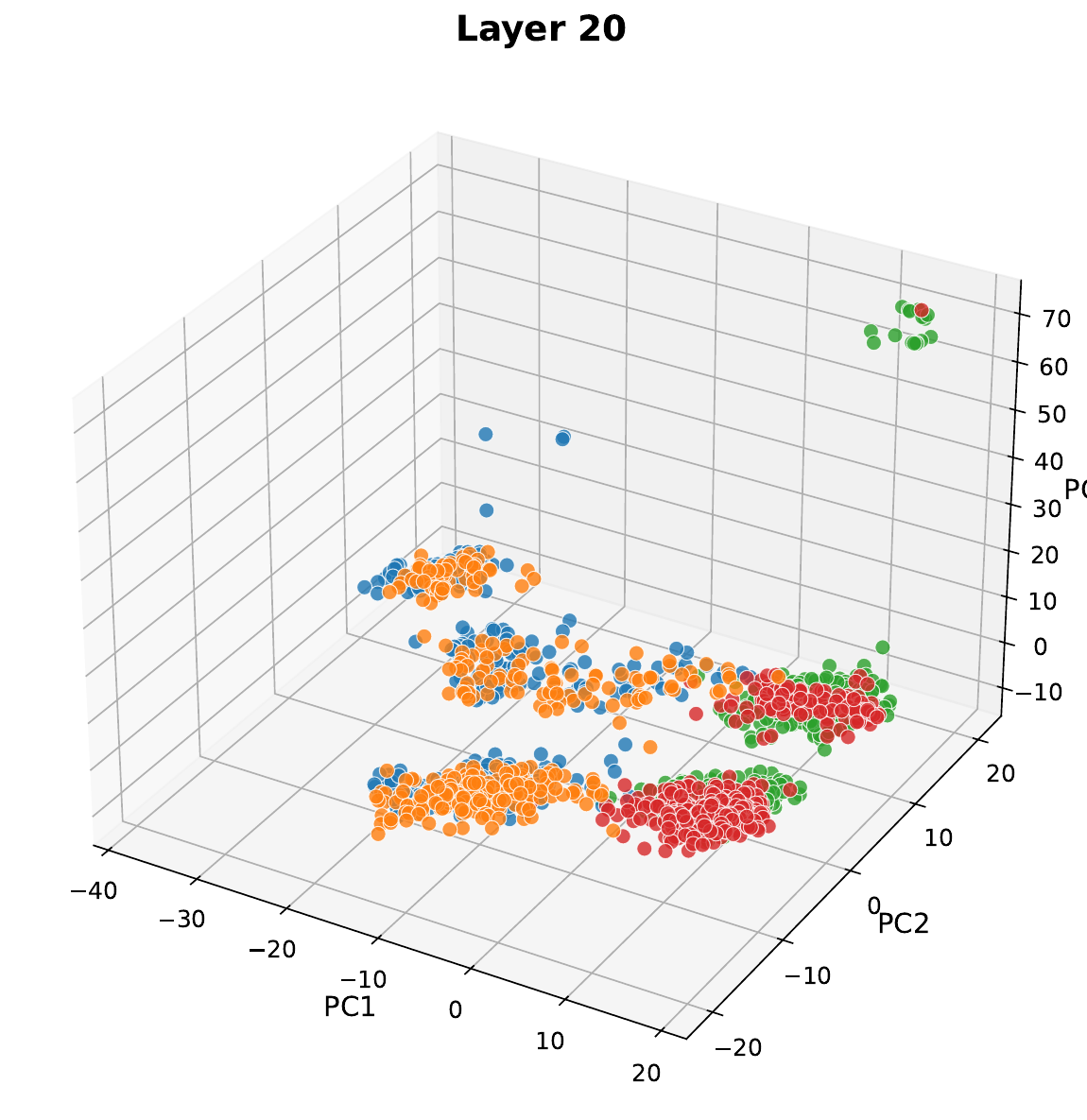}
    \end{subfigure}
    \begin{subfigure}[b]{0.25\textwidth}
        \includegraphics[width=\textwidth]{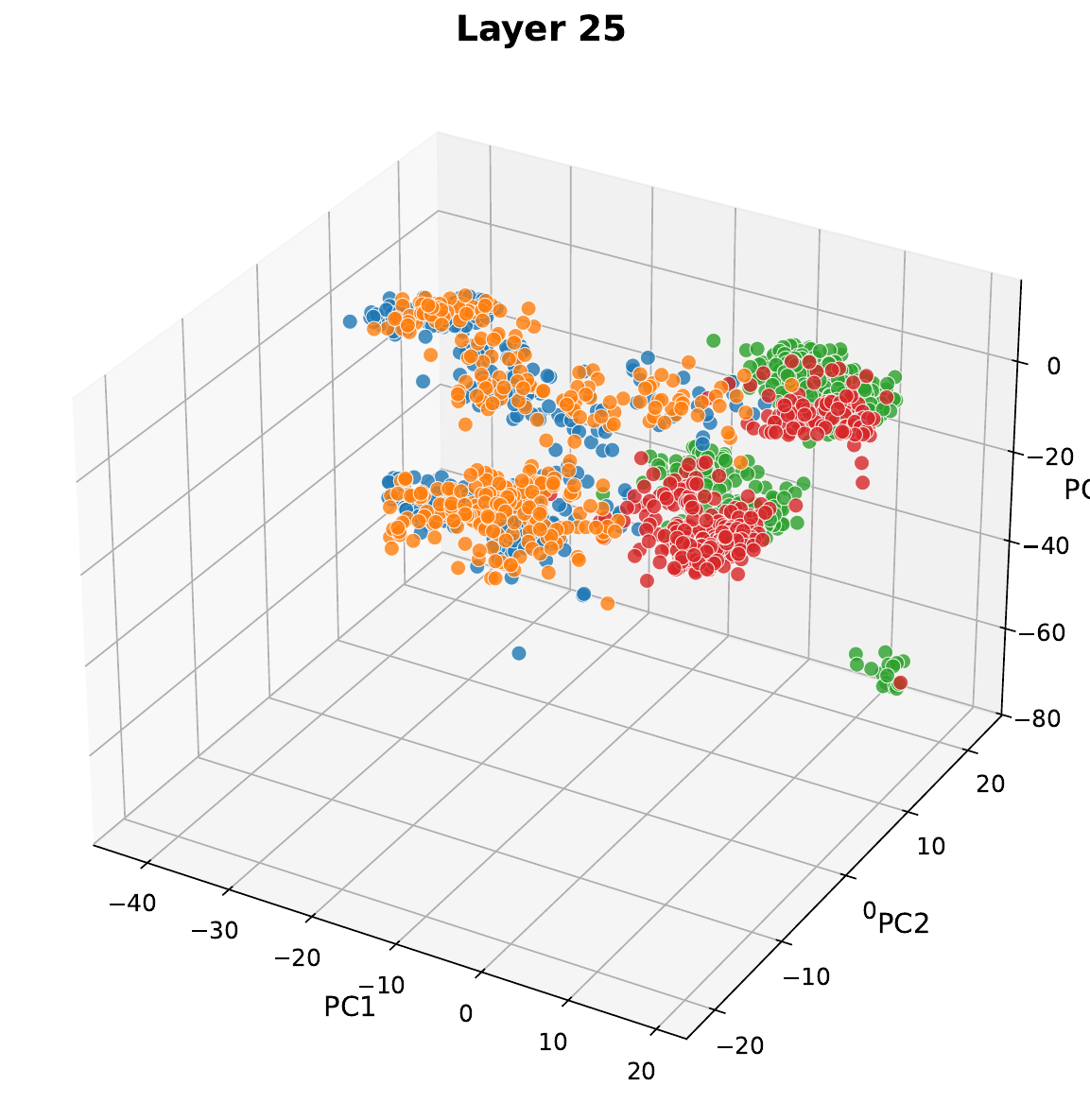}
    \end{subfigure}

    \begin{subfigure}[b]{0.25\textwidth}
        \includegraphics[width=\textwidth]{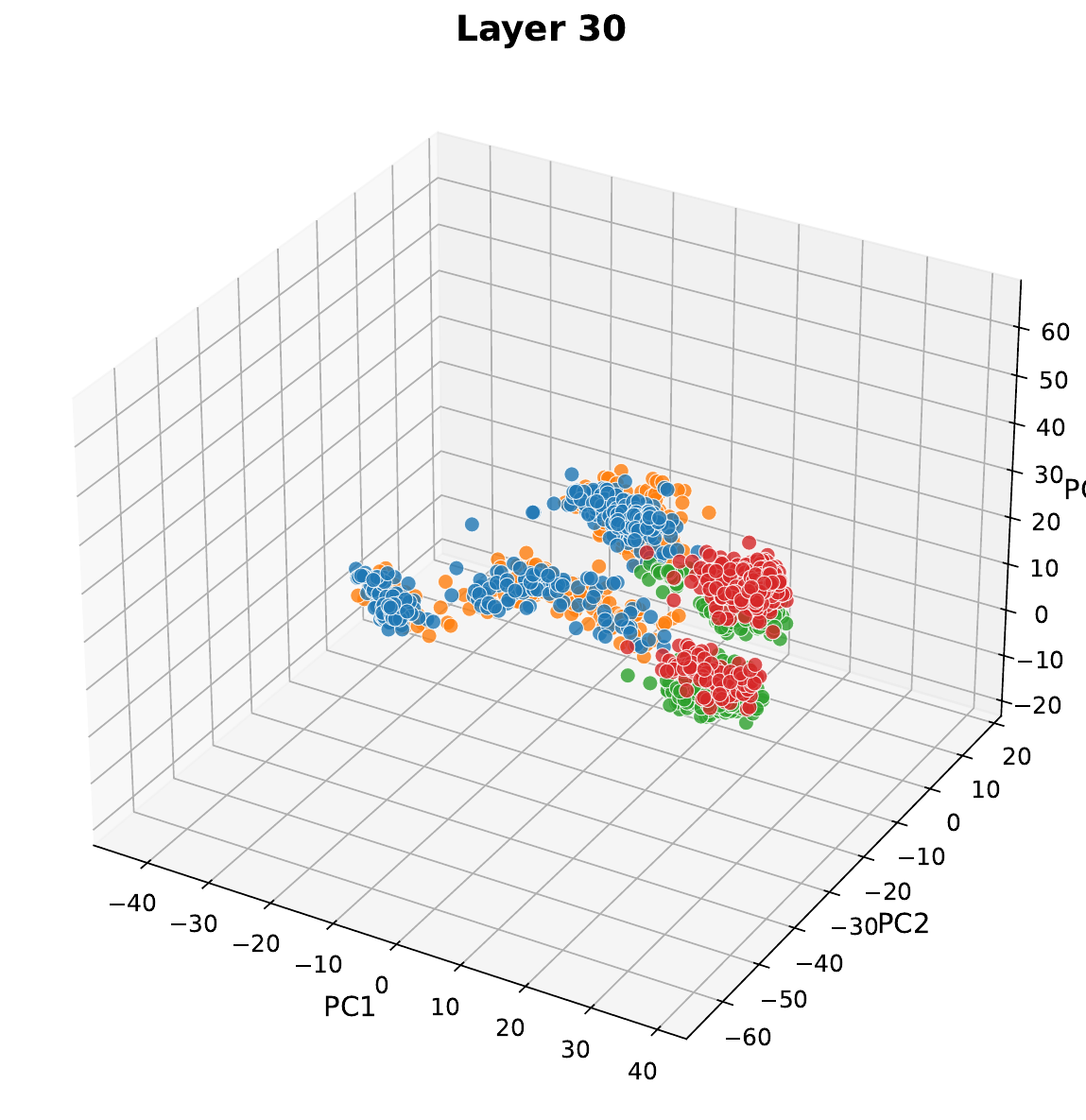}
    \end{subfigure}
    \begin{subfigure}[b]{0.25\textwidth}
        \includegraphics[width=\textwidth]{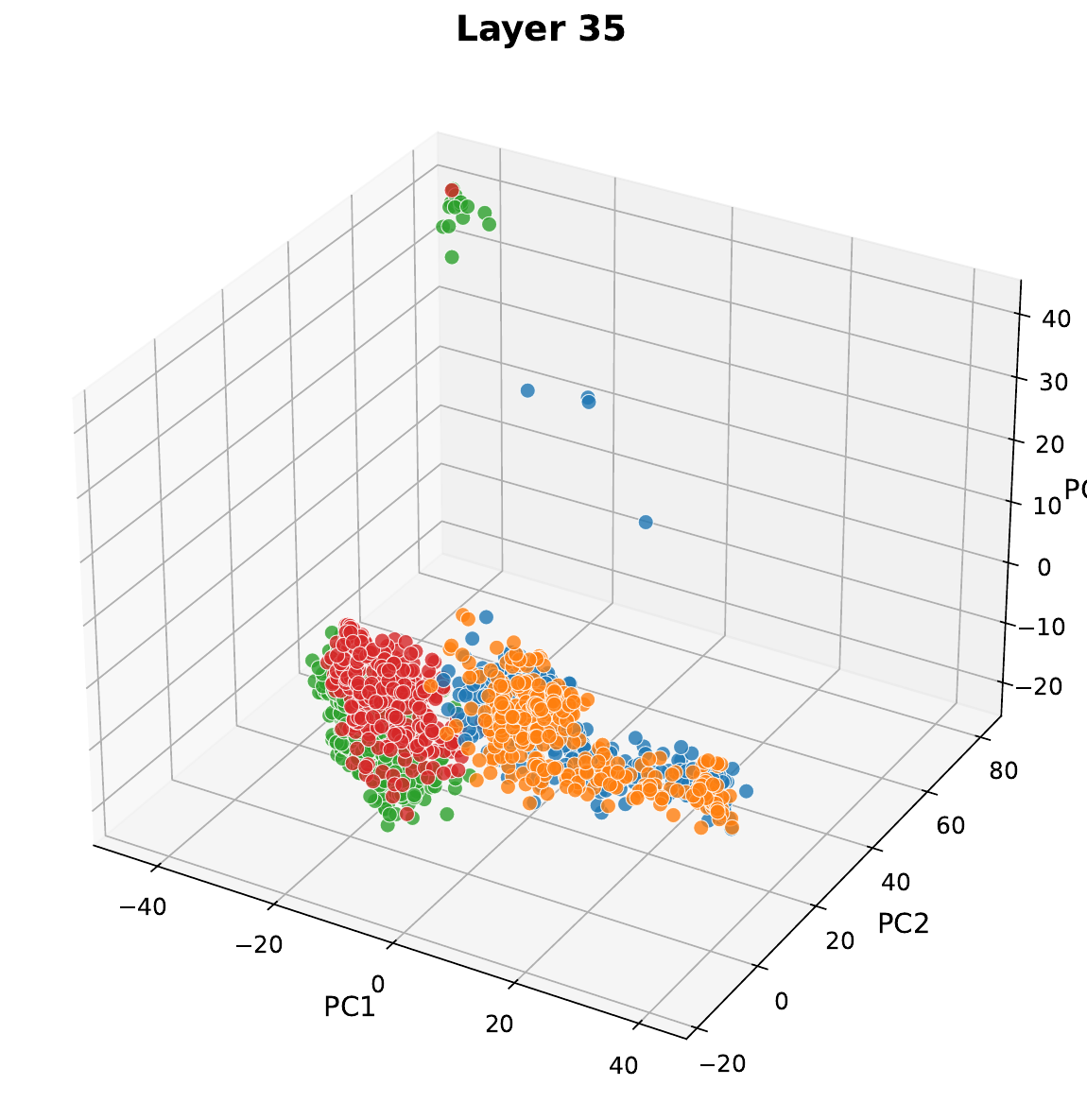}
    \end{subfigure}
    \begin{subfigure}[b]{0.25\textwidth}
        \includegraphics[width=\textwidth]{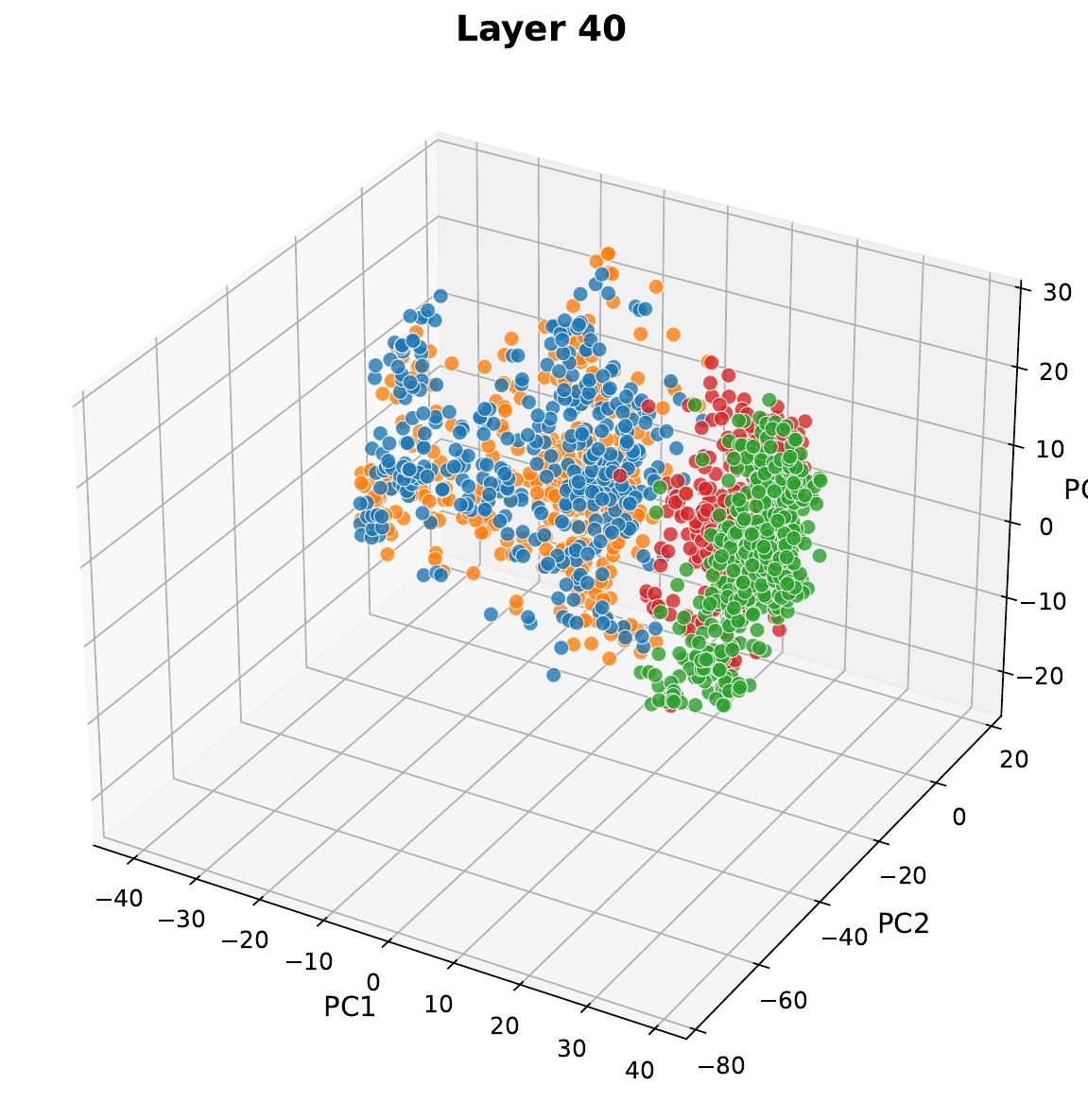}
    \end{subfigure}

    \begin{subfigure}[b]{0.25\textwidth}
        \includegraphics[width=\textwidth]{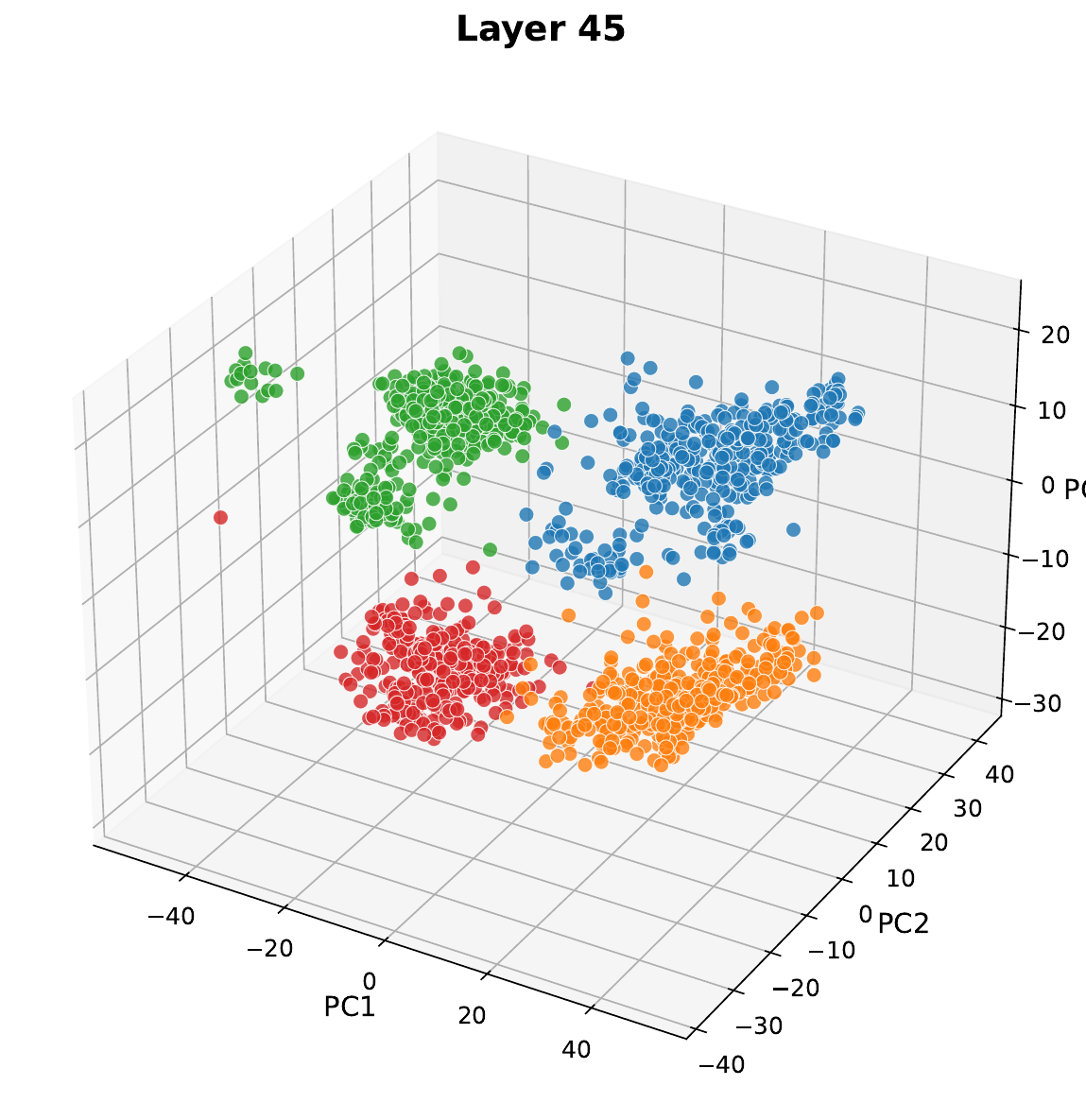}
    \end{subfigure}
    \begin{subfigure}[b]{0.25\textwidth}
        \includegraphics[width=\textwidth]{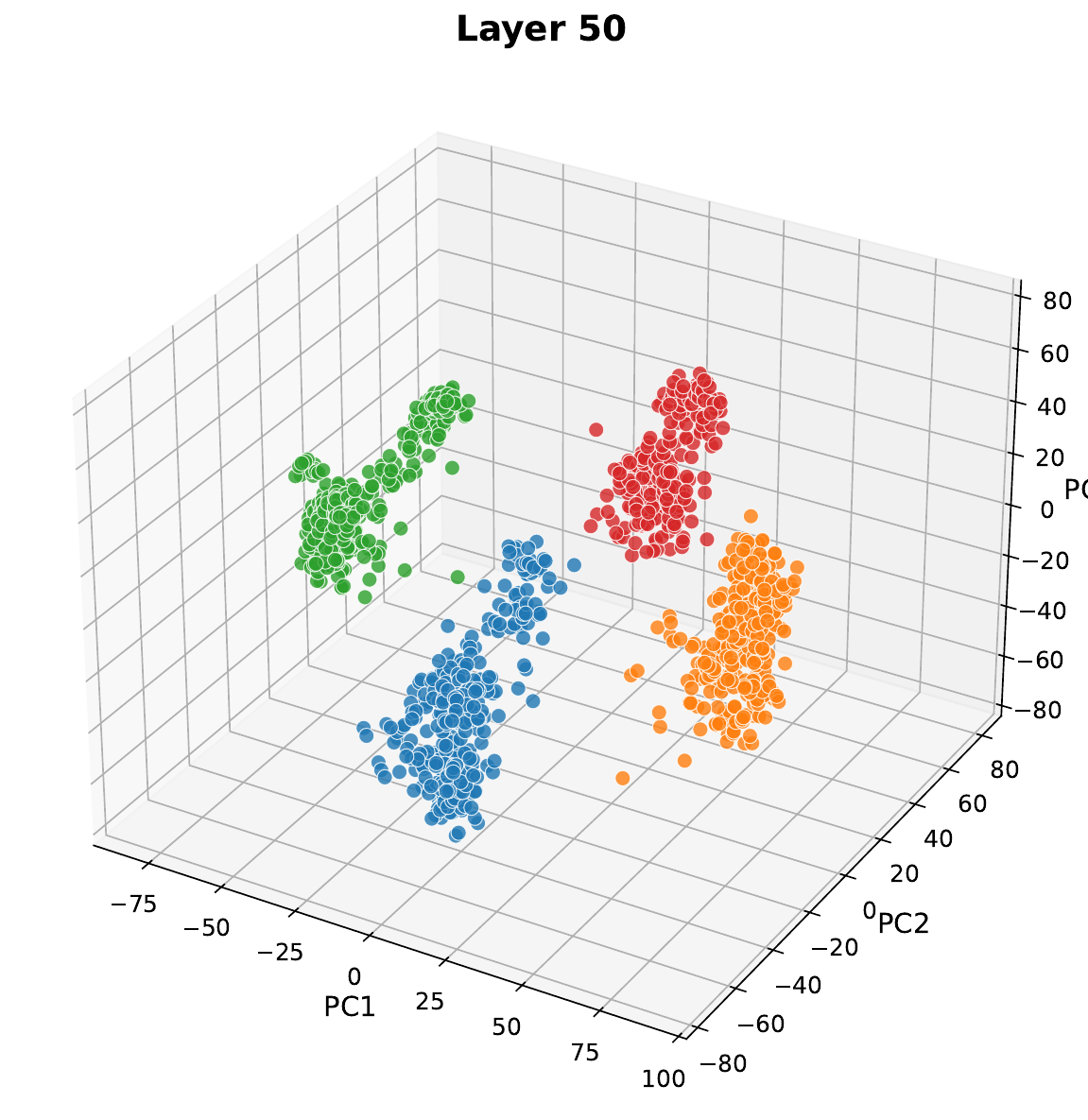}
    \end{subfigure}
    \begin{subfigure}[b]{0.25\textwidth}
        \includegraphics[width=\textwidth]{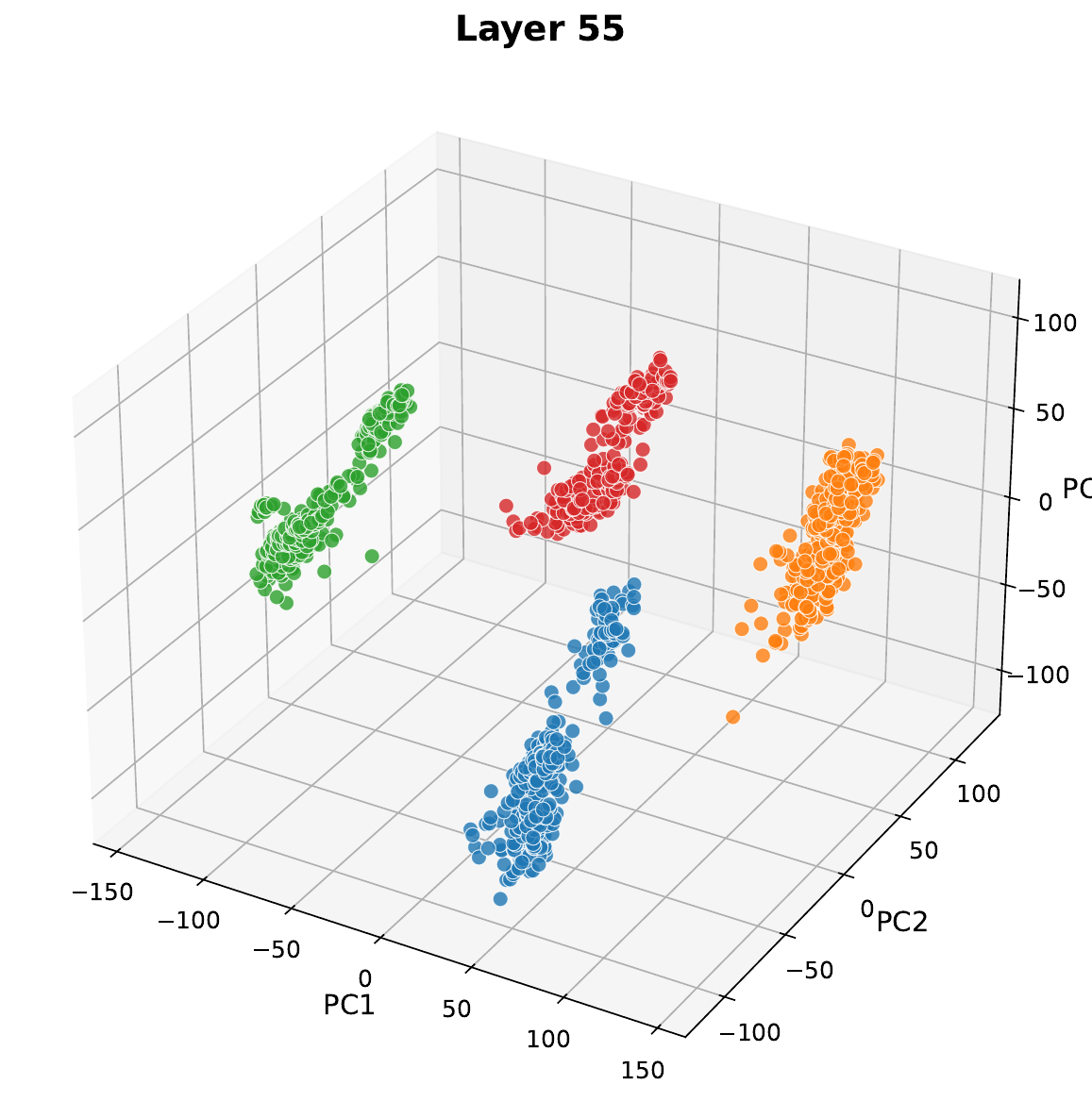}
    \end{subfigure}

    \begin{subfigure}[b]{0.25\textwidth}
        \includegraphics[width=\textwidth]{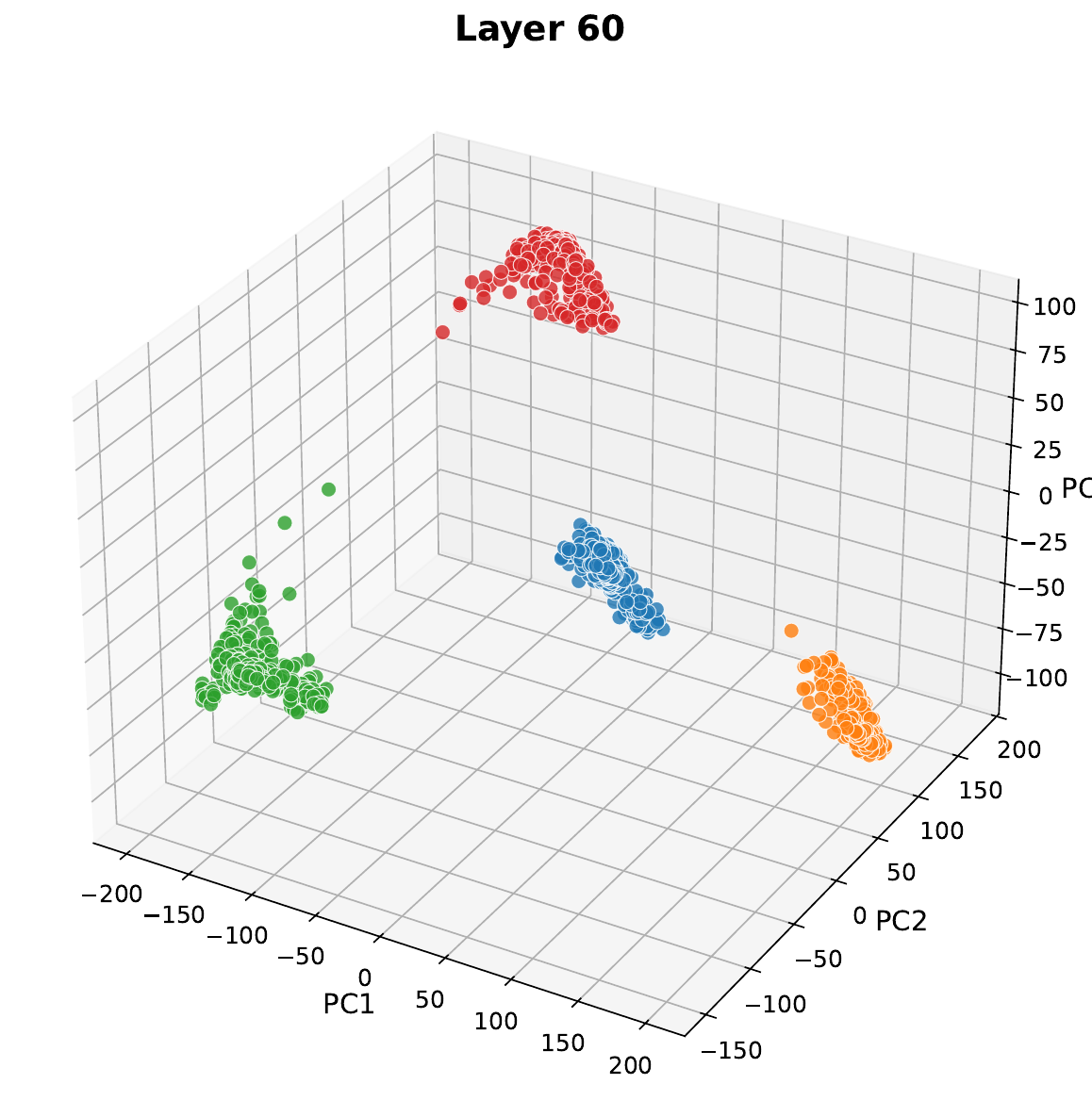}
    \end{subfigure}
    \begin{subfigure}[b]{0.25\textwidth}
        \includegraphics[width=\textwidth]{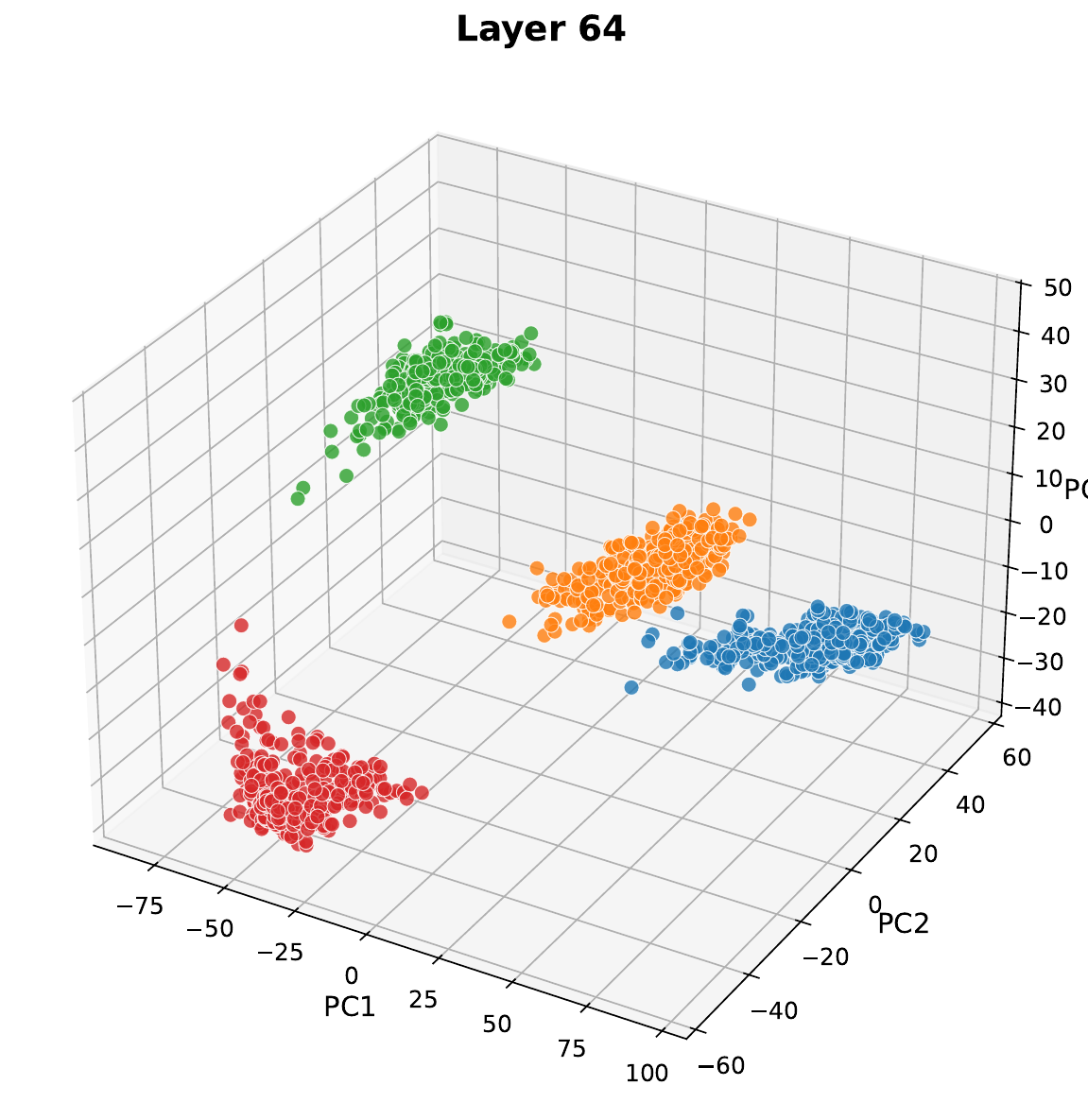}
    \end{subfigure}
    \begin{subfigure}[b]{0.25\textwidth}
        \centering
        \includegraphics[width=0.6\textwidth]{figures/pca/legend.png}
        \vspace{1cm}
    \end{subfigure}

\caption{3D PCA projections of logical validity and plausibility classification prompts for \texttt{Qwen2.5-32B-Instruct} in the CoT setting. The projections illustrate how the model forms distinct clusters for validity and plausibility classes across layers.}
\label{fig:pca-cot}
\end{figure*}

Figures~\ref{fig:pca-zero} and~\ref{fig:pca-cot} show PCA projections at layers 1, 5, 10, 15, 20, 25, 30, 35, 40, 45, 50, 55, 60, and 64 for zero-shot and CoT settings, respectively. The representational structure evolves substantially through the network, with separability between classes emerging only in later layers.

In both settings, the direction separating validity categories (valid vs. invalid) is approximately parallel to the direction separating plausibility categories (true vs. false). In the later layers, particularly layers 50--55, the four clusters arrange such that moving from invalid to valid represents a similar directional shift in the representational space as moving from false to true.

Comparing the projections extracted from the two prompting conditions, in the zero-shot setting (Figure~\ref{fig:pca-zero}), many data points lie close to the decision boundary separating the classes of each task, even in the later layers. In contrast, in the CoT setting (Figure~\ref{fig:pca-cot}), the four clusters are much more distinct and well-separated, with points forming tighter groups farther from the decision boundaries. This clearer representational separation in CoT, with fewer borderline cases, reflects the model's improved behavioral accuracy under CoT prompting compared to zero-shot.

\begin{table*}[t!]
\centering
\begin{tabular}{lrrrrr}
\toprule
& Coef. & Std.Err. & $z$ & $p$ & 95\% CI \\
\midrule
Intercept & -0.260 & 0.135 & -1.921 & 0.055 & [-0.525, 0.005] \\
Prompt Style (Zero-shot) & 0.188 & 0.016 & 12.019 & $<0.001$ & [0.158, 0.219] \\
Similarity & 0.557 & 0.233 & 2.387 & 0.017 & [0.100, 1.014] \\
\midrule
Random Intercept (LLM Var) & 0.001 & 0.028 & & & \\
\bottomrule
\end{tabular}
\caption{Mixed-effects regression predicting content effect from prompting style (zero-shot vs. CoT) and average plausibility-validity similarity at highly steerable layers. Model type (e.g. \texttt{Qwen2.5-7B-Instruct}) was included as a random intercept.}
\label{tab:regression-results}
\end{table*}

\subsection{Validity and Plausibility Vectors Analysis}
\label{app:full_steering}
\begin{figure*}[t!]
    \centering
    \begin{subfigure}[t]{\textwidth}
        \includegraphics[width=\linewidth]{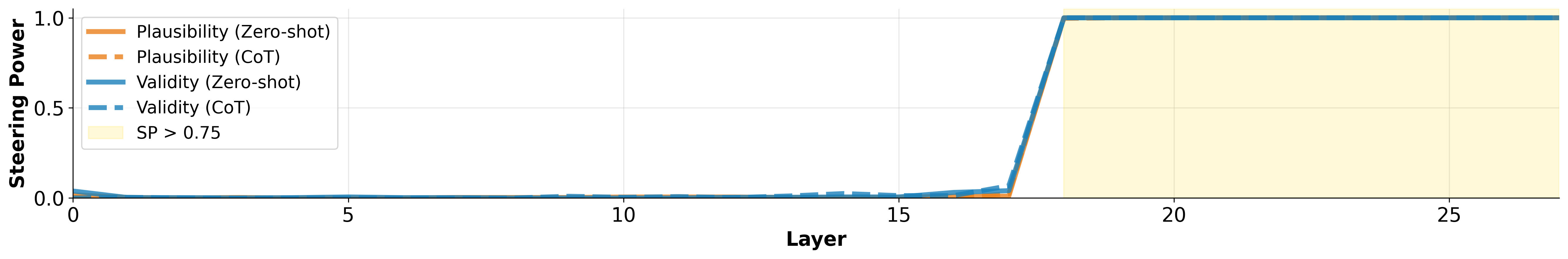}
        \caption{\texttt{Qwen2.5-7B-Instruct}}
    \end{subfigure}
    \hfill
    \begin{subfigure}[t]{\textwidth}
        \includegraphics[width=\linewidth]{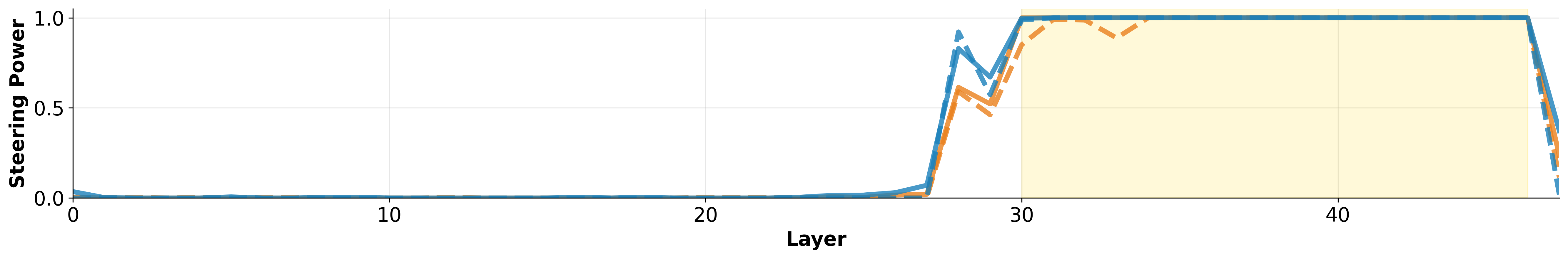}
        \caption{\texttt{Qwen2.5-14B-Instruct}}
    \end{subfigure}
    \hfill
    \begin{subfigure}[t]{\textwidth}
        \includegraphics[width=\linewidth]{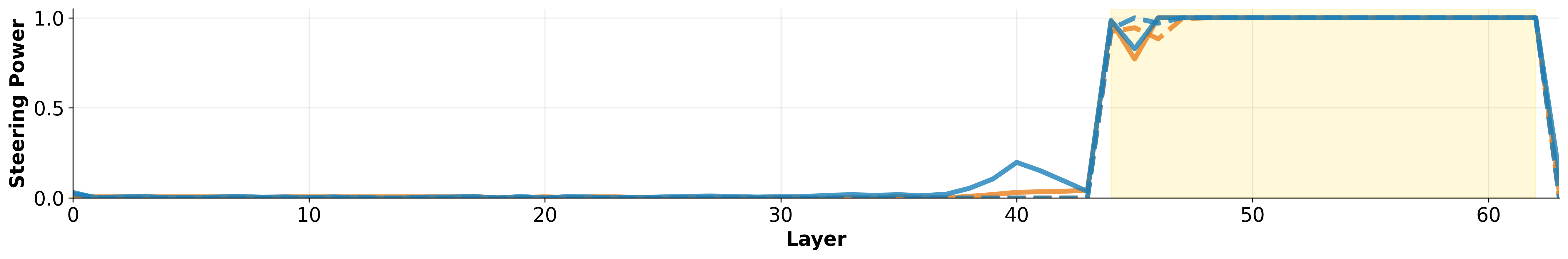}
        \caption{\texttt{Qwen2.5-32B-Instruct}}
    \end{subfigure}
    \caption{Steering power (SP) of validity and plausibility vectors applied at different hidden layers across different Qwen-2.5 model sizes. The yellow regions highlight layers with $\mathrm{SP} > 0.75$ for both validity and plausibility across prompt settings. Validity and plausibility steering vectors exhibit high SP at similar layers under both zero-shot and CoT prompting.}
    \label{fig:sp-qwen2.5}
\end{figure*}

\begin{figure*}[t!]
    \centering
    \begin{subfigure}[t]{\textwidth}
        \includegraphics[width=\linewidth]{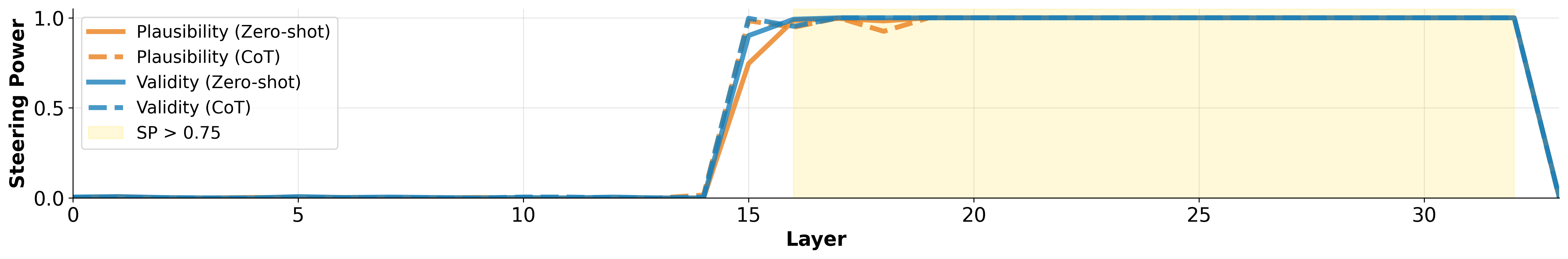}
        \caption{\texttt{Gemma3-4B-it}}
    \end{subfigure}
    \hfill
    \begin{subfigure}[t]{\textwidth}
        \includegraphics[width=\linewidth]{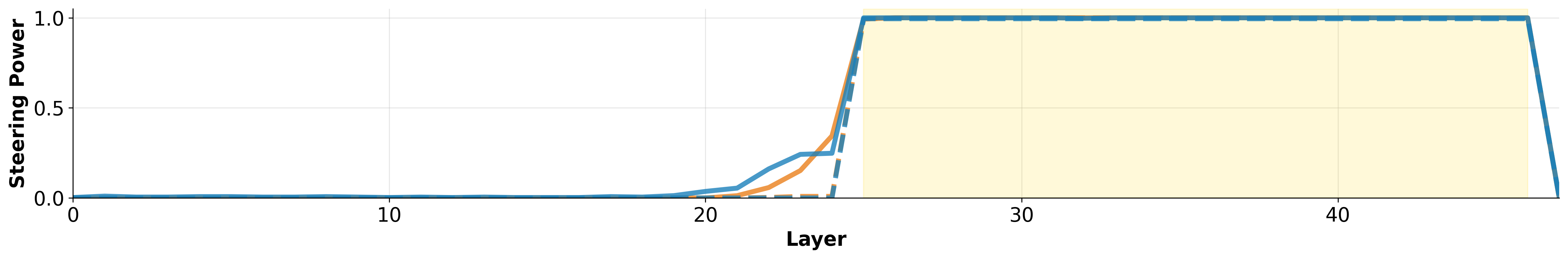}
        \caption{\texttt{Gemma3-12B-it}}
    \end{subfigure}
    \hfill
    \begin{subfigure}[t]{\textwidth}
        \includegraphics[width=\linewidth]{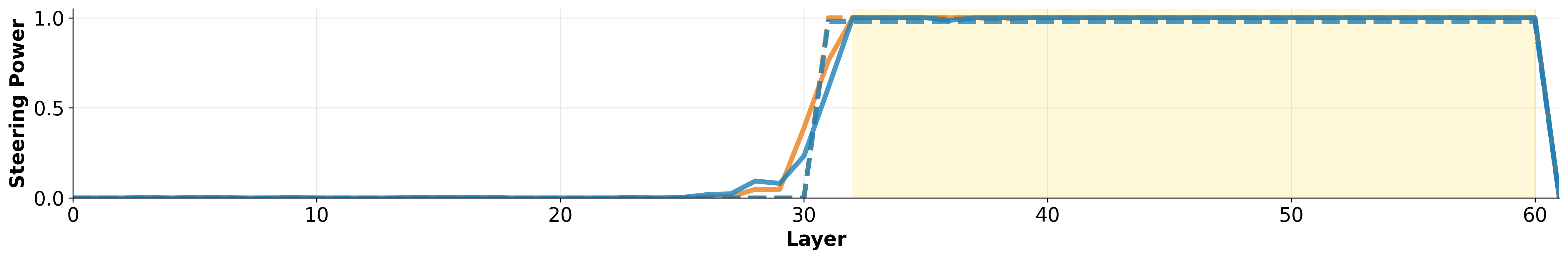}
        \caption{\texttt{Gemma3-27B-it}}
    \end{subfigure}
    \caption{Steering power (SP) of validity and plausibility vectors applied at different hidden layers across different Gemma-3 model sizes. The yellow regions highlight layers with $\mathrm{SP} > 0.75$ for both validity and plausibility across prompt settings. Validity and plausibility steering vectors exhibit high SP at similar layers under both zero-shot and CoT prompting.}
    \label{fig:sp-gemma3}
\end{figure*}

\begin{figure*}[t!]
    \centering
    \begin{subfigure}[t]{\textwidth}
        \includegraphics[width=\linewidth]{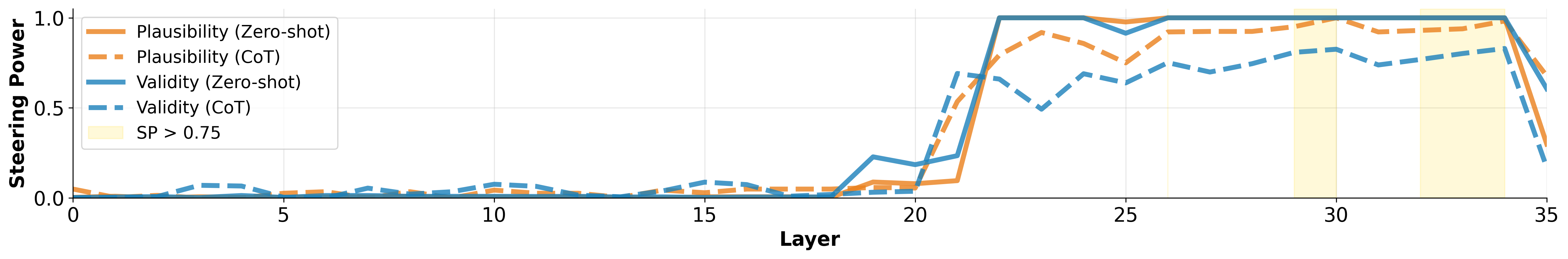}
        \caption{\texttt{Qwen3-4B}}
    \end{subfigure}
    \hfill
    \begin{subfigure}[t]{\textwidth}
        \includegraphics[width=\linewidth]{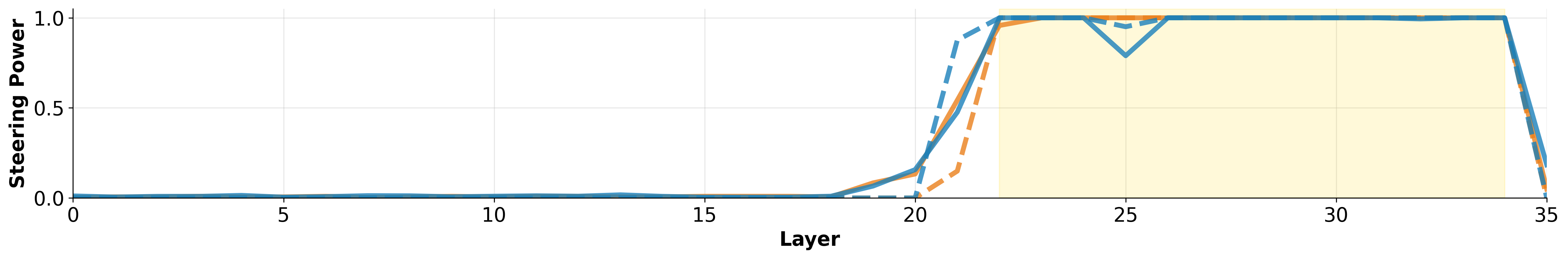}
        \caption{\texttt{Qwen3-8B}}
    \end{subfigure}
    \hfill
    \begin{subfigure}[t]{\textwidth}
        \includegraphics[width=\linewidth]{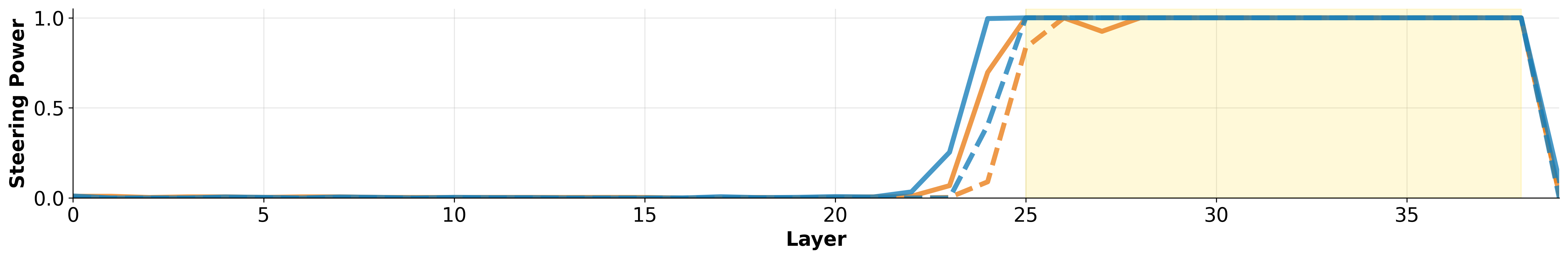}
        \caption{\texttt{Qwen3-14B}}
    \end{subfigure}
    \hfill
    \begin{subfigure}[t]{\textwidth}
        \includegraphics[width=\linewidth]{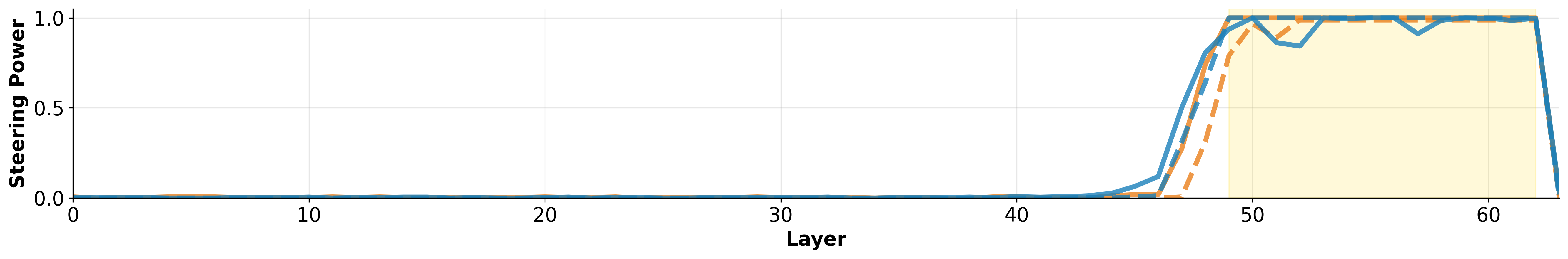}
        \caption{\texttt{Qwen3-32B}}
    \end{subfigure}
    \caption{Steering power (SP) of validity and plausibility vectors applied at different hidden layers across different Qwen-3 model sizes. The yellow regions highlight layers with $\mathrm{SP} > 0.75$ for both validity and plausibility across prompt settings. Validity and plausibility steering vectors exhibit high SP at similar layers under both zero-shot and CoT prompting.}
    \label{fig:sp-qwen3}
\end{figure*}

\begin{figure*}[t!]
    \centering
    \begin{subfigure}[t]{0.48\textwidth}
        \centering
        \includegraphics[height=0.82\textwidth,angle=90]{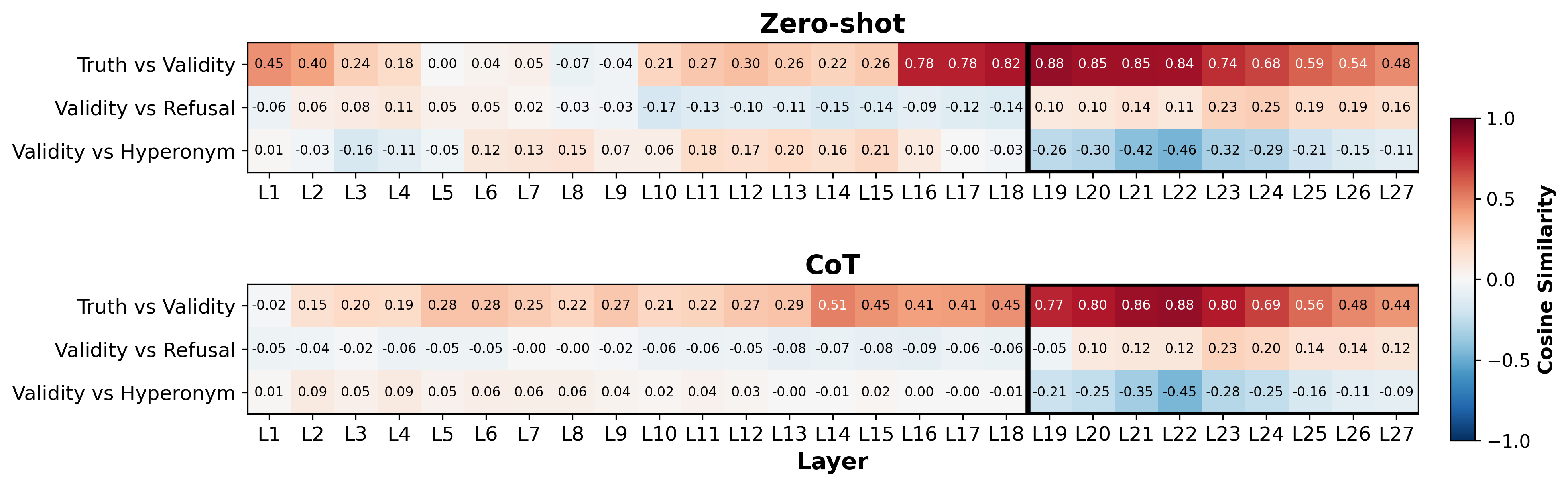}
        \caption{\texttt{Qwen2.5-7B-Instruct}}
        \label{fig:similarities-qwen2.5-7B}
    \end{subfigure}
    \hfill
    \begin{subfigure}[t]{0.48\textwidth}
        \centering
        \includegraphics[height=0.7\textwidth,angle=90]{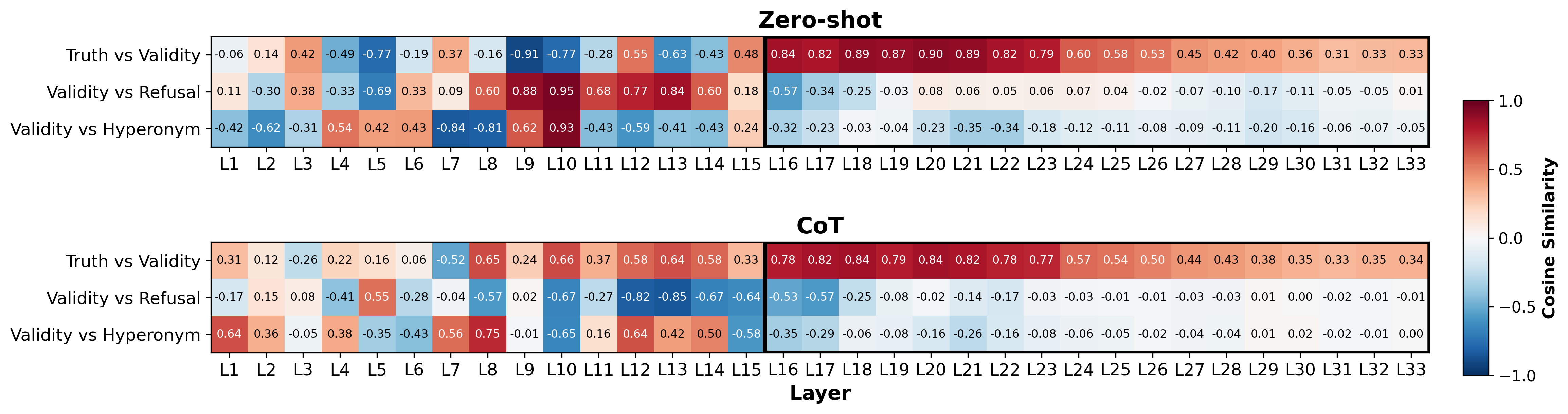}
        \caption{\texttt{Gemma3-4B-it}}
        \label{fig:similarities-gemma3-4B}
    \end{subfigure}
    \caption{Cosine similarities of vectors representing different concepts with validity vectors extracted at different hidden layers of \texttt{Qwen2.5-7B-Instruct} and \texttt{Gemma3-4B-it}. We consider vectors representing plausibility, harmlessness, and hypernymity. The layers with $\mathrm{SP} > 0.75$ for both validity and plausibility are highlighted using thicker black rectangles. At highly steerable layers, only validity and plausibility have high similarity.}
\end{figure*}

\begin{figure*}[t!]
    \centering
    \begin{subfigure}[t]{0.48\textwidth}
        \centering
        \includegraphics[height=0.52\textwidth,angle=90]{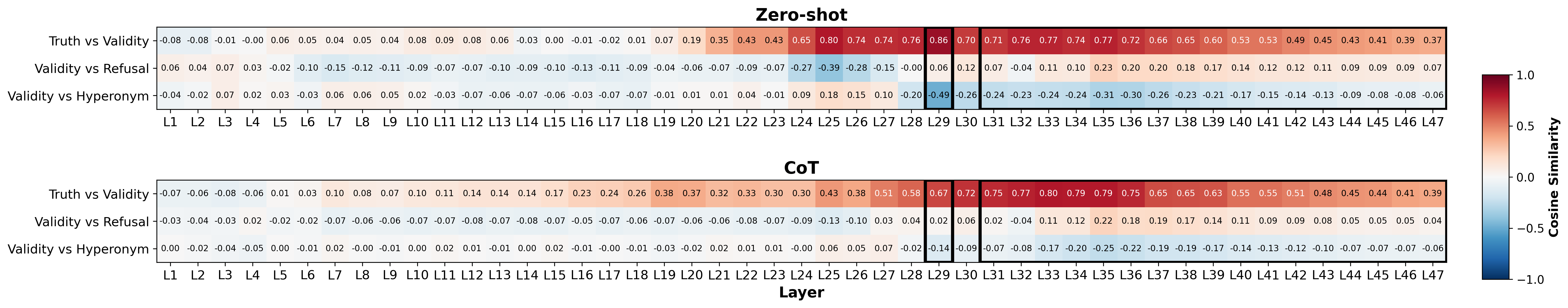}
        \caption{\texttt{Qwen2.5-14B-Instruct}}
        \label{fig:similarities-qwen2.5-14B}
    \end{subfigure}
    \hfill
    \begin{subfigure}[t]{0.48\textwidth}
        \centering
        \includegraphics[height=0.4\textwidth,angle=90]{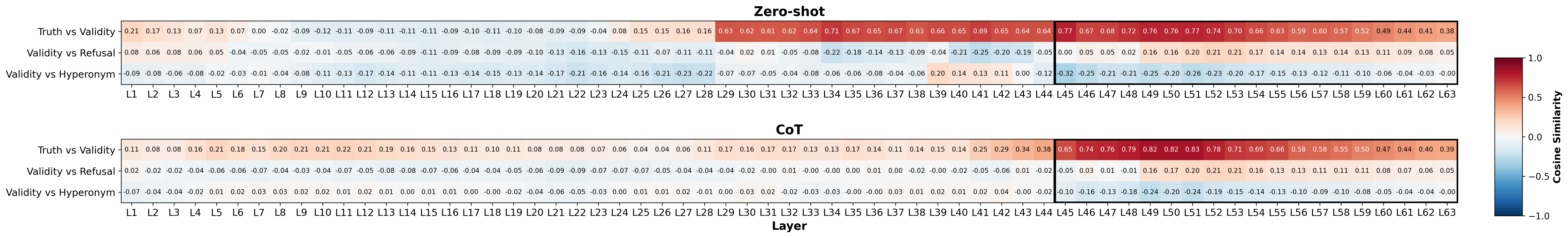}
        \caption{\texttt{Qwen2.5-32B-Instruct}}
        \label{fig:similarities-qwen2.5-32B}
    \end{subfigure}
    \caption{Cosine similarities of vectors representing different concepts with validity vectors extracted at different hidden layers of \texttt{Qwen2.5-32B-Instruct} and \texttt{Qwen2.5-14B-Instruct}. We consider vectors representing plausibility, harmlessness, and hypernymity. The layers with $\mathrm{SP} > 0.75$ for both validity and plausibility are highlighted using thicker black rectangles. At highly steerable layers, only validity and plausibility have high similarity.}
\end{figure*}

\begin{figure*}[t!]
    \centering
    \begin{subfigure}[t]{0.48\textwidth}
        \centering
        \includegraphics[height=0.7\textwidth,angle=90]{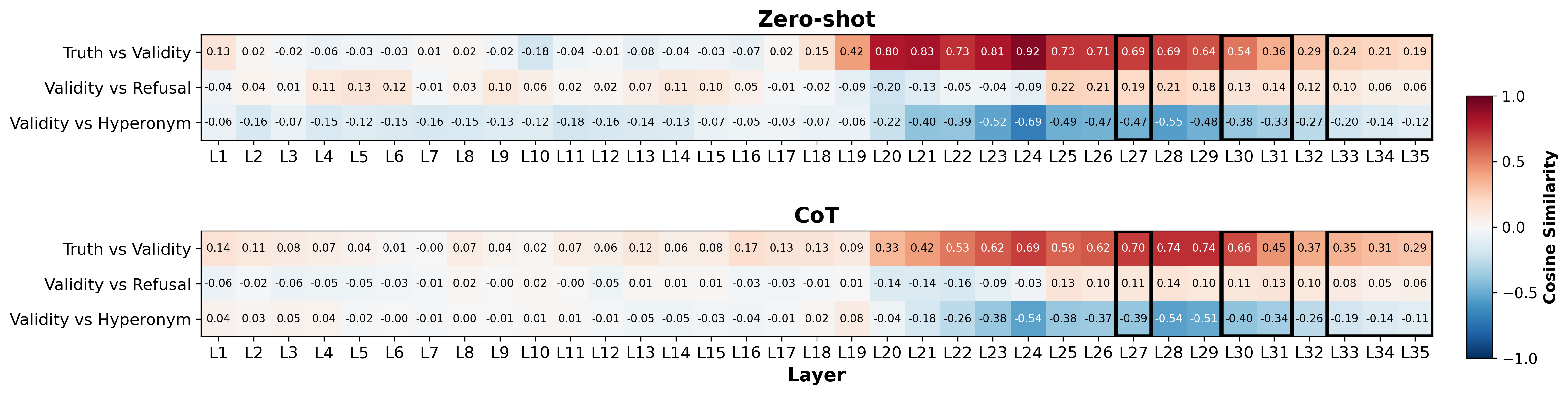}
        \caption{\texttt{Qwen3-4B}}
        \label{fig:similarities-qwen3-4B}
    \end{subfigure}
    \hfill
    \begin{subfigure}[t]{0.48\textwidth}
        \centering
        \includegraphics[height=0.7\textwidth,angle=90]{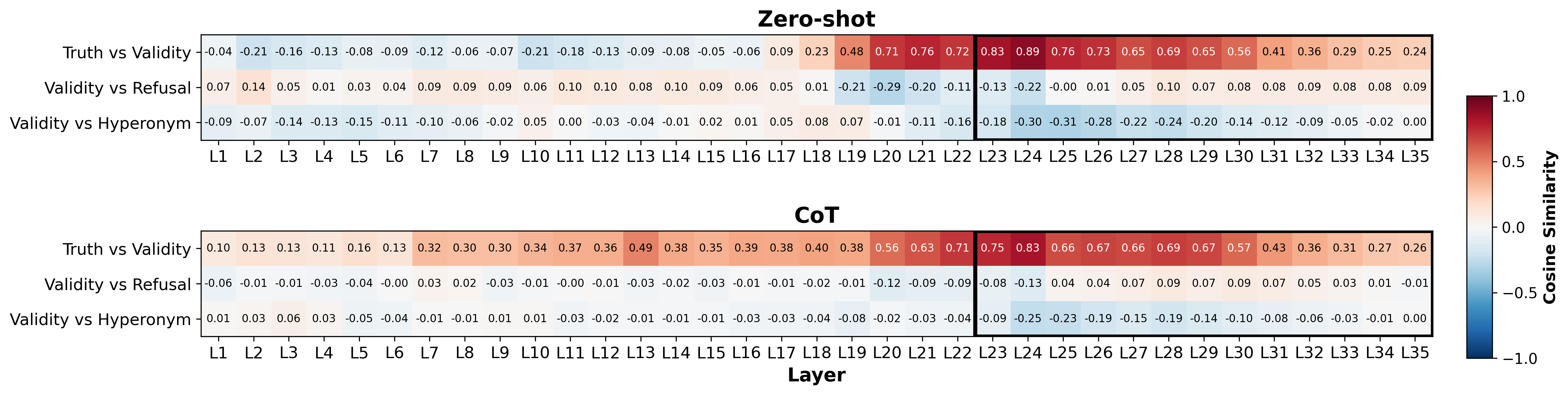}
        \caption{\texttt{Qwen3-8B}}
        \label{fig:similarities-qwen3-8B}
    \end{subfigure}
    \caption{Cosine similarities of vectors representing different concepts with validity vectors extracted at different hidden layers of \texttt{Qwen3-4B} and \texttt{Qwen3-8B}. We consider vectors representing plausibility, harmlessness, and hypernymity. The layers with $\mathrm{SP} > 0.75$ for both validity and plausibility are highlighted using thicker black rectangles. At highly steerable layers, only validity and plausibility have high similarity.}
\end{figure*}

\begin{figure*}[t!]
    \centering
    \begin{subfigure}[t]{0.48\textwidth}
        \centering
        \includegraphics[height=0.65\linewidth,angle=90]{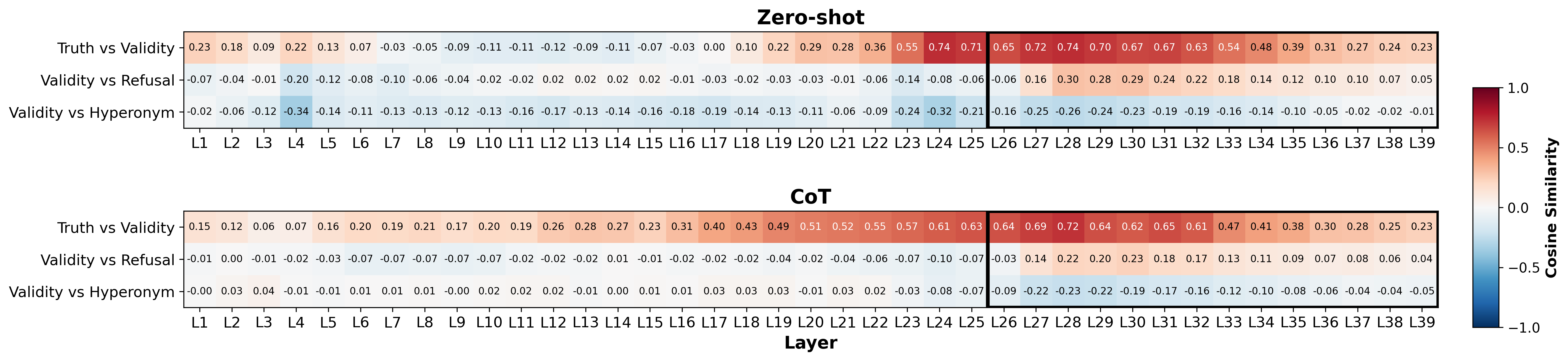}
        \caption{\texttt{Qwen3-14B}}
        \label{fig:similarities-qwen3-14B}
    \end{subfigure}
    \hfill
    \begin{subfigure}[t]{0.48\textwidth}
        \centering
        \includegraphics[height=0.43\linewidth,angle=90]{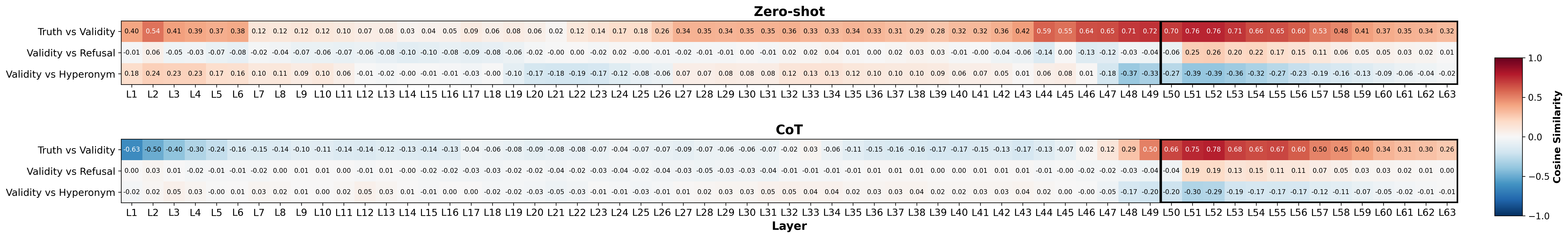}
        \caption{\texttt{Qwen3-32B}}
        \label{fig:similarities-qwen3-32B}
    \end{subfigure}
    \caption{Cosine similarities of vectors representing different concepts with validity vectors extracted at different hidden layers of \texttt{Qwen3-14B} and \texttt{Qwen3-32B}. We consider vectors representing plausibility, harmlessness, and hypernymity. The layers with $\mathrm{SP} > 0.75$ for both validity and plausibility are highlighted using thicker black rectangles. At highly steerable layers, only validity and plausibility have high similarity.}
\end{figure*}

\begin{figure*}[t!]
    \centering
    \begin{subfigure}[t]{0.48\textwidth}
        \centering
        \includegraphics[height=0.53\linewidth,angle=90]{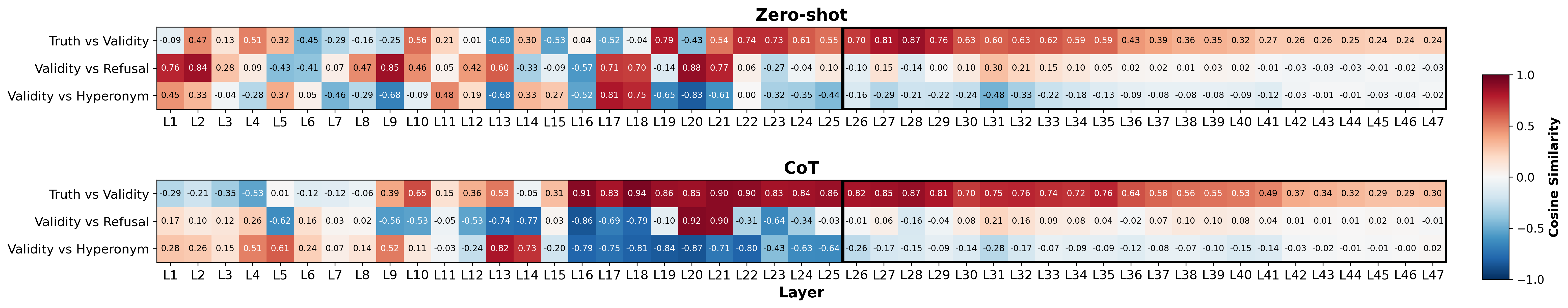}
        \caption{\texttt{Gemma3-12B-it}}
        \label{fig:similarities-gemma3-12B}
    \end{subfigure}
    \hfill
    \begin{subfigure}[t]{0.48\textwidth}
        \centering
        \includegraphics[height=0.42\linewidth,angle=90]{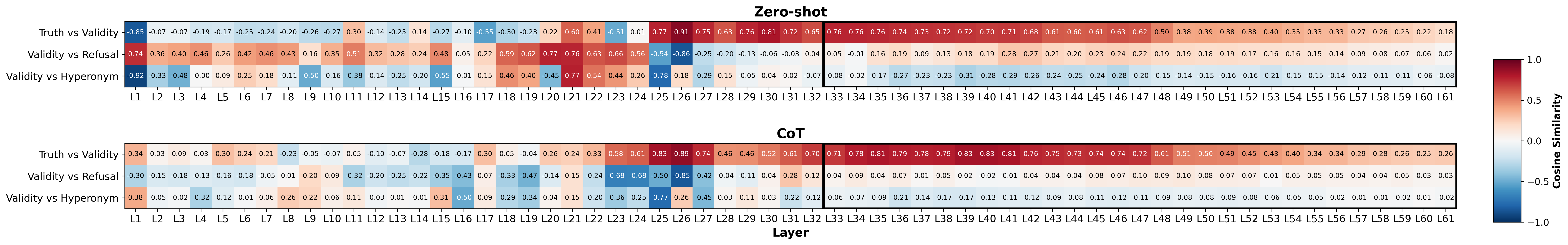}
        \caption{\texttt{Gemma3-27B-it}}
        \label{fig:similarities-gemma3-27B}
    \end{subfigure}
    \caption{Cosine similarities of vectors representing different concepts with validity vectors extracted at different hidden layers of \texttt{Gemma3-12B-it} and \texttt{Gemma3-27B-it}. We consider vectors representing plausibility, harmlessness, and hypernymity. The layers with $\mathrm{SP} > 0.75$ for both validity and plausibility are highlighted using thicker black rectangles. At highly steerable layers, only validity and plausibility have high similarity.}
\end{figure*}

We extend the analysis of validity and plausibility steering vectors to the full set of models beyond those reported in the main body of the paper. Linear steerability results using the steering power (SP) metric (see Section~\ref{sec:metrics} for the definition of SP) are shown for the Qwen-2.5 family in Figure~\ref{fig:sp-qwen2.5}, for the Gemma-3 family in Figure~\ref{fig:sp-gemma3}, and for the Qwen-3 family in Figure~\ref{fig:sp-qwen3}. The corresponding cosine similarity analyses of validity and plausibility vectors are reported in Figures~\ref{fig:similarities-qwen2.5-7B}--\ref{fig:similarities-gemma3-27B}.

Across all model families and sizes, the same core patterns described in the main text hold. Steering vectors derived from validity and plausibility achieve high effectiveness in the later layers, with $\mathrm{SP} \approx 1$, while producing negligible effects in the earlier layers where $\mathrm{SP} \approx 0$. This pattern is consistent across both zero-shot and CoT prompting. Furthermore, in the layers where validity and plausibility are strongly steerable ($\mathrm{SP} > 0.75$; highlighted in Figures~\ref{fig:sp-qwen2.5}--\ref{fig:sp-qwen3}), their vectors exhibit consistently high cosine similarity. By contrast, vectors for harmlessness and hypernymy remain weakly correlated or anticorrelated in the same regions, indicating that the observed similarity is specific to validity and plausibility.

Some additional nuances are worth noting. In \texttt{Qwen3-4B} (Figure~\ref{fig:sp-qwen3}), the window of layers with $\mathrm{SP} > 0.75$ is narrower compared to larger models in the same family. In the Gemma-3 family (Figures~\ref{fig:similarities-gemma3-4B}--\ref{fig:similarities-gemma3-27B}), we observe a mixture of high and low similarities across concepts in the early layers. However, these occur in regions where steering is ineffective ($\mathrm{SP} < 0.75$) and therefore do not undermine the conclusion that validity and plausibility vectors are uniquely aligned in the layers where they are also highly effective.

Overall, these extended results confirm that the linear steerability and representational alignment of validity and plausibility are robust phenomena across model families and scales. The minor deviations observed in \texttt{Qwen3-4B} and in the early layers of Gemma-3 seem to highlight family-specific patterns, but do not alter the central conclusion that validity and plausibility are closely aligned in the regions of the model most relevant for steering.

\subsection{Complete Mixed-effect Regression Results}
\label{app:full_mixed-effect}

In the mixed-effect linear regression experiment (see Section \ref{sec:regression}), we evaluated a total of 10 models in zero-shot and CoT prompt formats, with three prompt variants for each prompting style. This amounts to a total of 60 model-prompt pairs. However, the analysis was conducted on only 57 model-prompt pairs, as prompt variants 2 and 3 for \texttt{Qwen2.5-7B-Instruct} and prompt variant 3 for \texttt{Qwen3-4B} in the CoT setting were excluded. Despite being instructed to use step-by-step reasoning in the CoT condition, these smaller models produced direct answers on most prompts without generating intermediate reasoning, making them unsuitable for classification as CoT reasoning.

Table~\ref{tab:regression-results} reports the complete regression results. The coefficient for similarity was significantly positive, and prompting style had a robust effect. The variance of the random intercept for LLMs was negligible, suggesting that these effects were consistent across models.

\subsection{Causal Cross-influence between Plausibility and Validity}
\label{app:full_cross_steering}

Section \ref{sec:cross-influence} we showed that plausibility and validity have causal cross-influence on each other and reported steering results for two representative models. Here, we extend the analysis to the full set of models across the Qwen-2.5, Gemma-3, and Qwen-3 families. The experimental setup is identical: steering vectors are applied with a sign determined by the model’s prediction, and we measure the steering power (SP) of plausibility vectors during validity classification (``plausibility $\rightarrow$ validity'') and of validity vectors during plausibility classification (``validity $\rightarrow$ plausibility'').

Across all model families, we find consistent evidence of bidirectional steerability. Nearly all models exhibit at least one hidden layer with strong cross-task transfer ($\mathrm{SP} > 0.75$), with the sole exception of \texttt{Qwen3-4B}. Nevertheless, even in this model, moderate steering power ($\mathrm{SP} \approx 0.5$) emerges in middle-to-late layers.

\begin{figure*}[t!]
    \centering
    \begin{subfigure}[t]{\textwidth}
        \includegraphics[width=\linewidth]{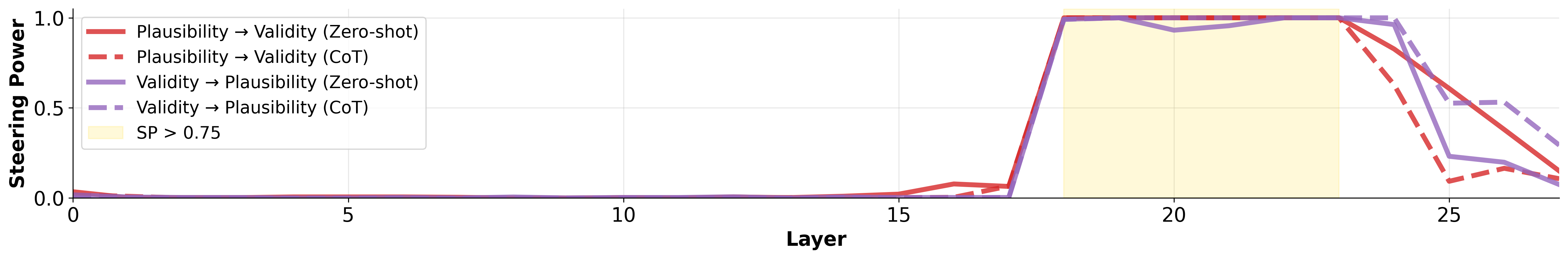}
        \caption{\texttt{Qwen2.5-7B-Instruct}}
    \end{subfigure}
    \hfill
    \begin{subfigure}[t]{\textwidth}
        \includegraphics[width=\linewidth]{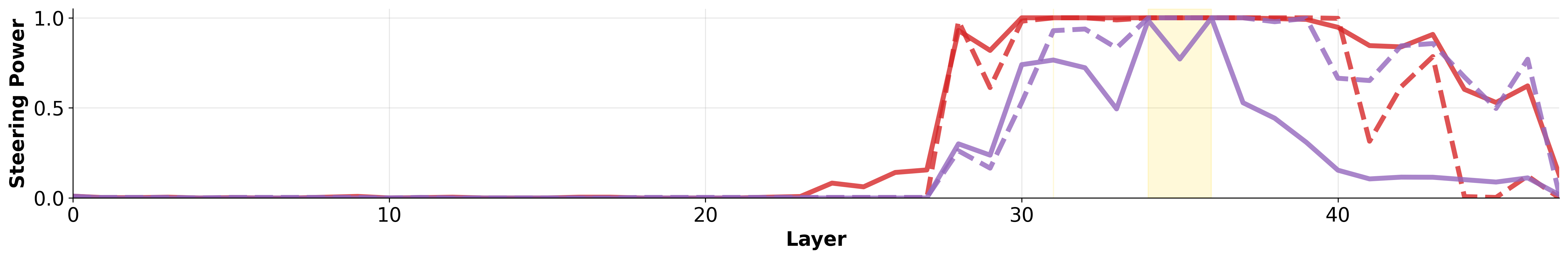}
        \caption{\texttt{Qwen2.5-14B-Instruct}}
    \end{subfigure}
    \hfill
    \begin{subfigure}[t]{\textwidth}
        \includegraphics[width=\linewidth]{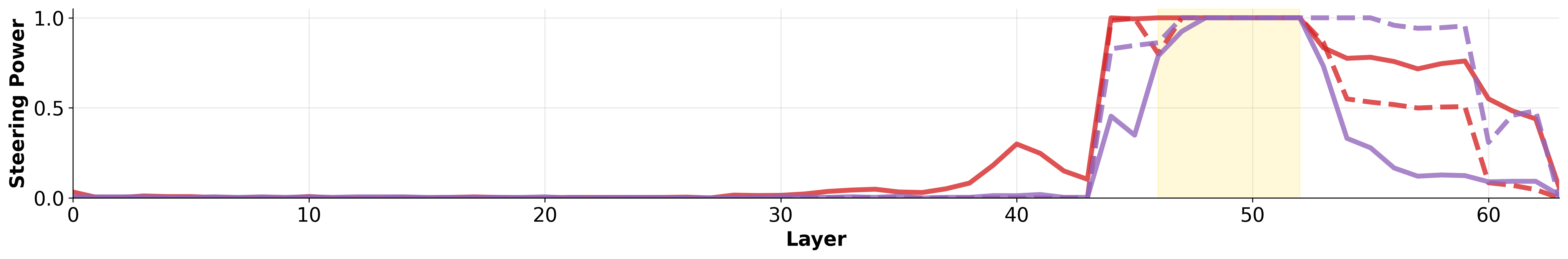}
        \caption{\texttt{Qwen2.5-32B-Instruct}}
    \end{subfigure}
    \caption{Steering power (SP) of plausibility steering vectors applied to validity classification (''plausibility $\rightarrow$ validity''), and vice versa (''validity $\rightarrow$ plausibility'') across different hidden layers and Qwen-2.5 model sizes. The yellow regions highlight layers with high transferability in both directions, where $\mathrm{SP} > 0.75$ across prompt settings. High cross-task SP is observed at similar layers under both zero-shot and CoT prompting.}
    \label{fig:cross-task-qwen2.5}
\end{figure*}

\begin{figure*}[t!]
    \centering
    \begin{subfigure}[t]{\textwidth}
        \includegraphics[width=\linewidth]{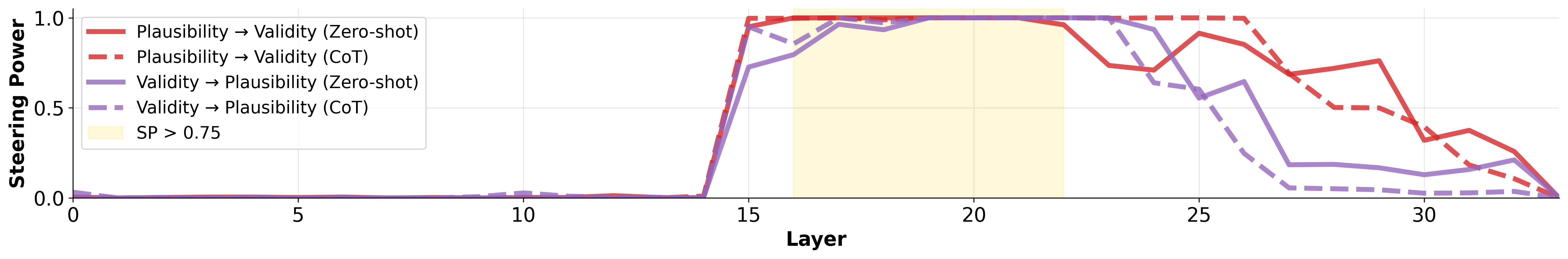}
        \caption{\texttt{Gemma3-4B-it}}
    \end{subfigure}
    \hfill
    \begin{subfigure}[t]{\textwidth}
        \includegraphics[width=\linewidth]{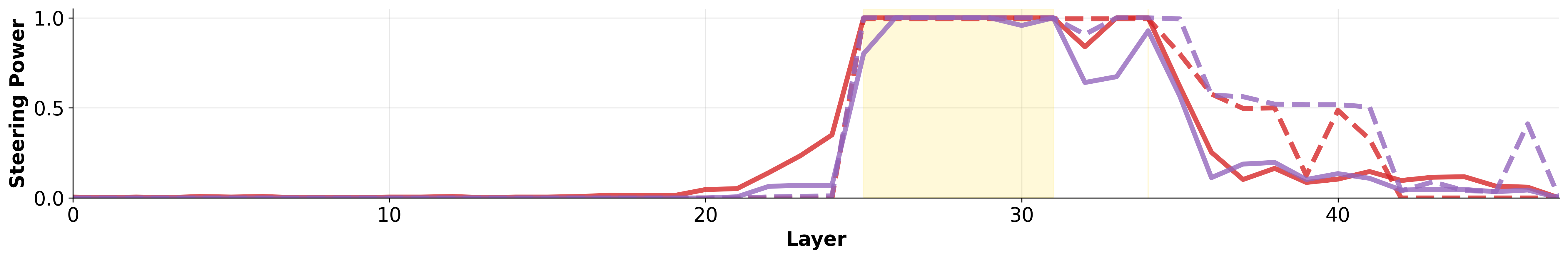}
        \caption{\texttt{Gemma3-12B-it}}
    \end{subfigure}
    \hfill
    \begin{subfigure}[t]{\textwidth}
        \includegraphics[width=\linewidth]{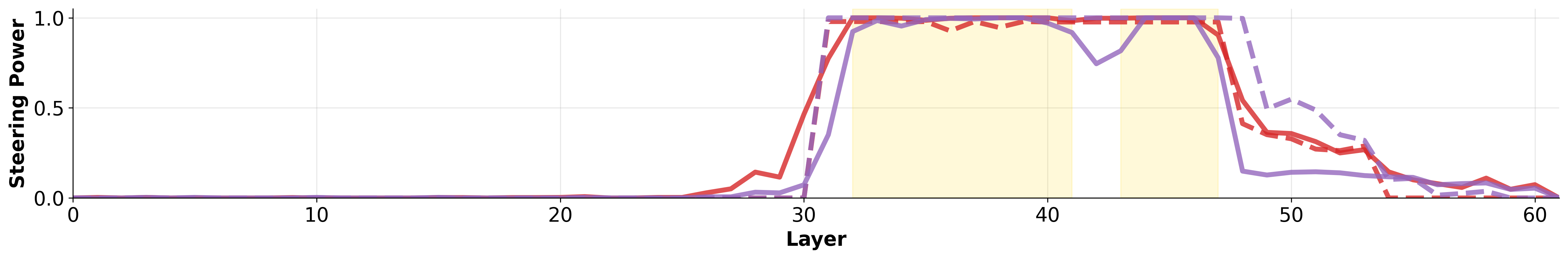}
        \caption{\texttt{Gemma3-27B-it}}
    \end{subfigure}
    \caption{Steering power (SP) of plausibility steering vectors applied to validity classification (``plausibility $\rightarrow$ validity''), and vice versa (``validity $\rightarrow$ plausibility'') across different hidden layers and Gemma-3 model sizes. The yellow regions highlight layers with high transferability in both directions, where $\mathrm{SP} > 0.75$ across prompt settings. High cross-task SP is observed at similar layers under both zero-shot and CoT prompting.}
    \label{fig:cross-task-gemma3}
\end{figure*}

\begin{figure*}[t!]
    \centering
    \begin{subfigure}[t]{\textwidth}
        \includegraphics[width=\linewidth]{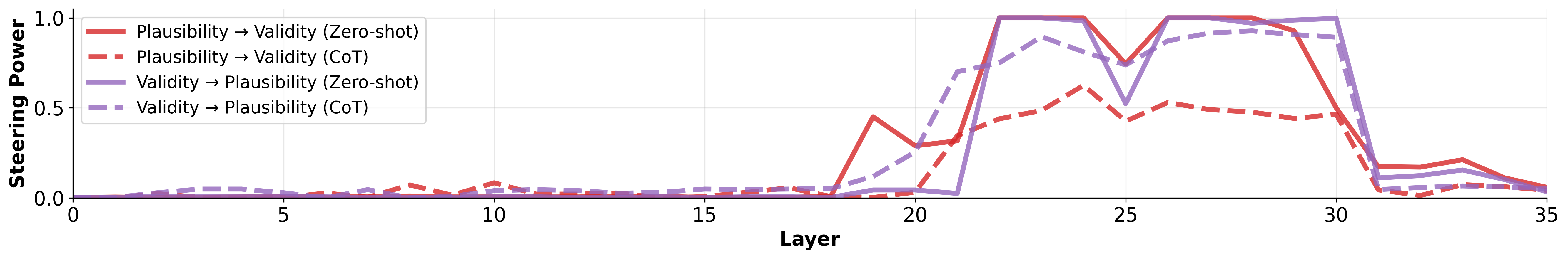}
        \caption{\texttt{Qwen3-4B}}
    \end{subfigure}
    \hfill
    \begin{subfigure}[t]{\textwidth}
        \includegraphics[width=\linewidth]{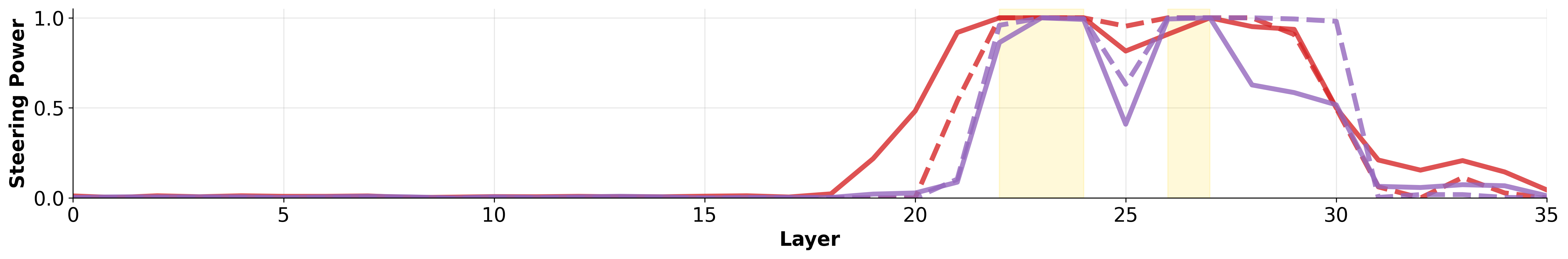}
        \caption{\texttt{Qwen3-8B}}
    \end{subfigure}
    \hfill
    \begin{subfigure}[t]{\textwidth}
        \includegraphics[width=\linewidth]{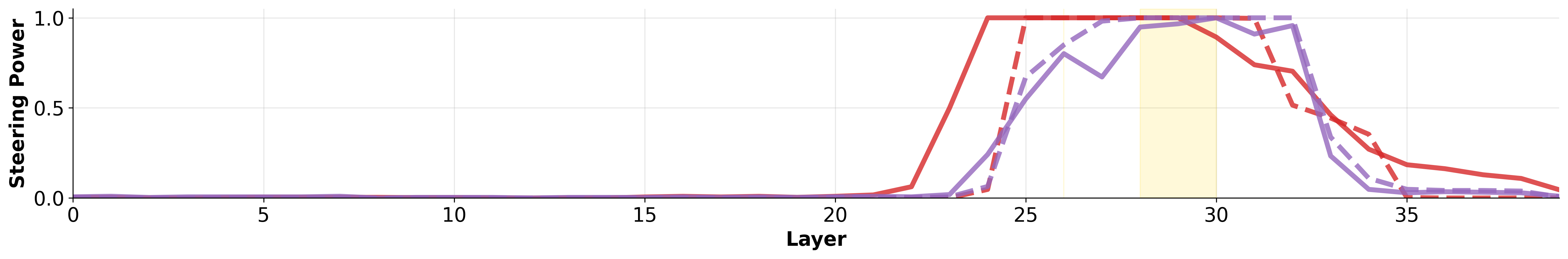}
        \caption{\texttt{Qwen3-14B}}
    \end{subfigure}
    \hfill
    \begin{subfigure}[t]{\textwidth}
        \includegraphics[width=\linewidth]{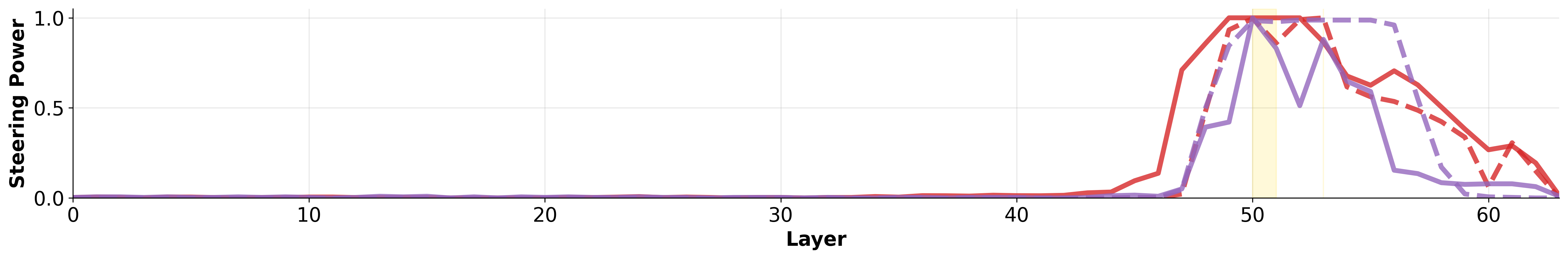}
        \caption{\texttt{Qwen3-32B}}
    \end{subfigure}
    \caption{Steering power (SP) of plausibility steering vectors applied to validity classification (''plausibility $\rightarrow$ validity''), and vice versa (''validity $\rightarrow$ plausibility'') across different hidden layers and Qwen-3 model sizes. The yellow regions highlight layers with high transferability in both directions, where $\mathrm{SP} > 0.75$ across prompt settings. High cross-task SP is observed at similar layers under both zero-shot and CoT prompting.}
    \label{fig:cross-task-qwen3}
\end{figure*}

Figure~\ref{fig:cross-task-qwen2.5} presents the full results for the Qwen-2.5 family, Figure~\ref{fig:cross-task-gemma3} for the Gemma-3 family, and Figure~\ref{fig:cross-task-qwen3} for the Qwen-3 family.

\subsection{Content Effect Mitigation}
\label{app:full_mitigation}

In Section \ref{sec:mitigation}, we introduced a steering intervention designed to mitigate content effects by disentangling plausibility and validity representations. Specifically, we defined the \emph{task-difference vector} $\mu_{V-P}^l = \mu_V^l - \mu_P^l$, which isolates representational dimensions unique to validity classification. When added to hidden states during the validity task, this vector is expected to reduce sensitivity to plausibility and thereby attenuate the bias. Here we extend the analysis to all models.

Unlike in the other steering experiments, we found that simply adding $\mu_{V-P}^l$ was insufficient to remove the bias. Instead, we scaled the vector by a factor $\alpha$. A grid search over $\alpha \in {1.0, 1.5, 2.0, 2.5, 3.0}$ on the two models presented in the main text indicated that $\alpha = 1.5$ was most effective, and we fixed this value for all other models. We then identified the optimal intervention layer by selecting the layer that best satisfied the following two criteria: (i) accuracy is improved or at least minimally degraded relative to the baseline (up to 5\% less than the baseline), (ii) content effect is reduced. The chosen layer is highlighted in orange in the figures.

Across all model families, we find that the intervention can meaningfully reduce content effects without sacrificing overall task performance. In many cases, the method yields simultaneous improvements in accuracy and amount of content effects, with the latter reduced to near zero.

\begin{figure*}[t!]
    \centering
    \begin{subfigure}[t]{\textwidth}
        \centering
        \includegraphics[width=0.8\linewidth]{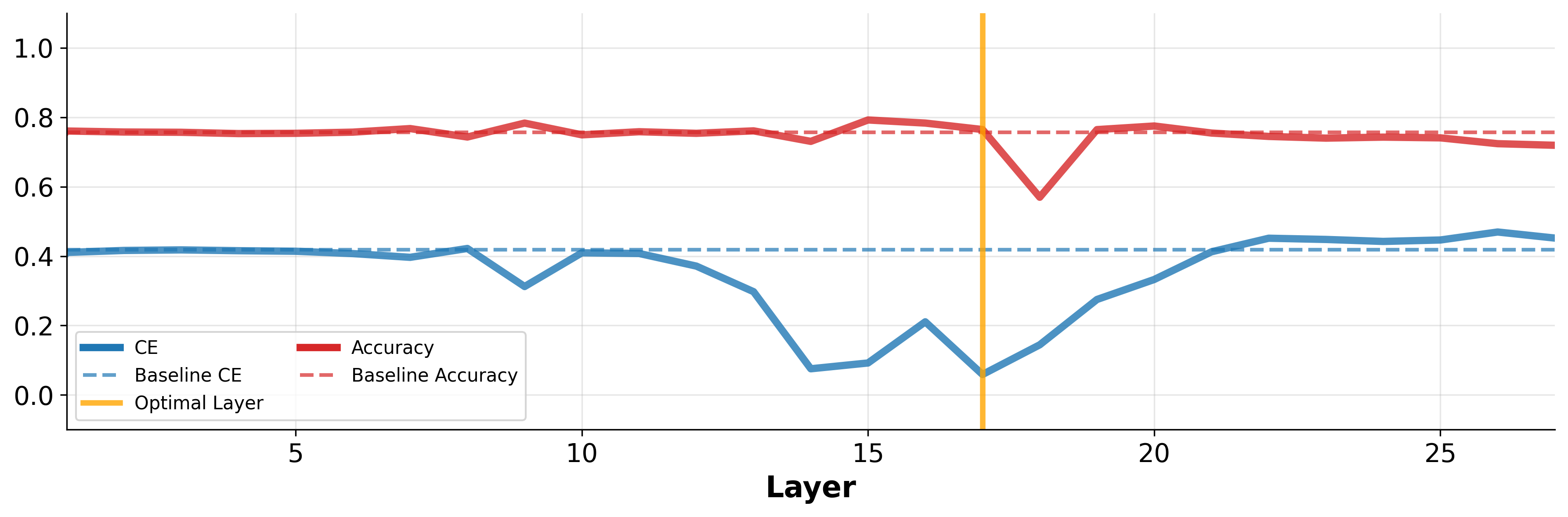}
        \caption{\texttt{Qwen2.5-7B-Instruct}}
    \end{subfigure}
    \hfill
    \begin{subfigure}[t]{\textwidth}
        \centering
        \includegraphics[width=0.8\linewidth]{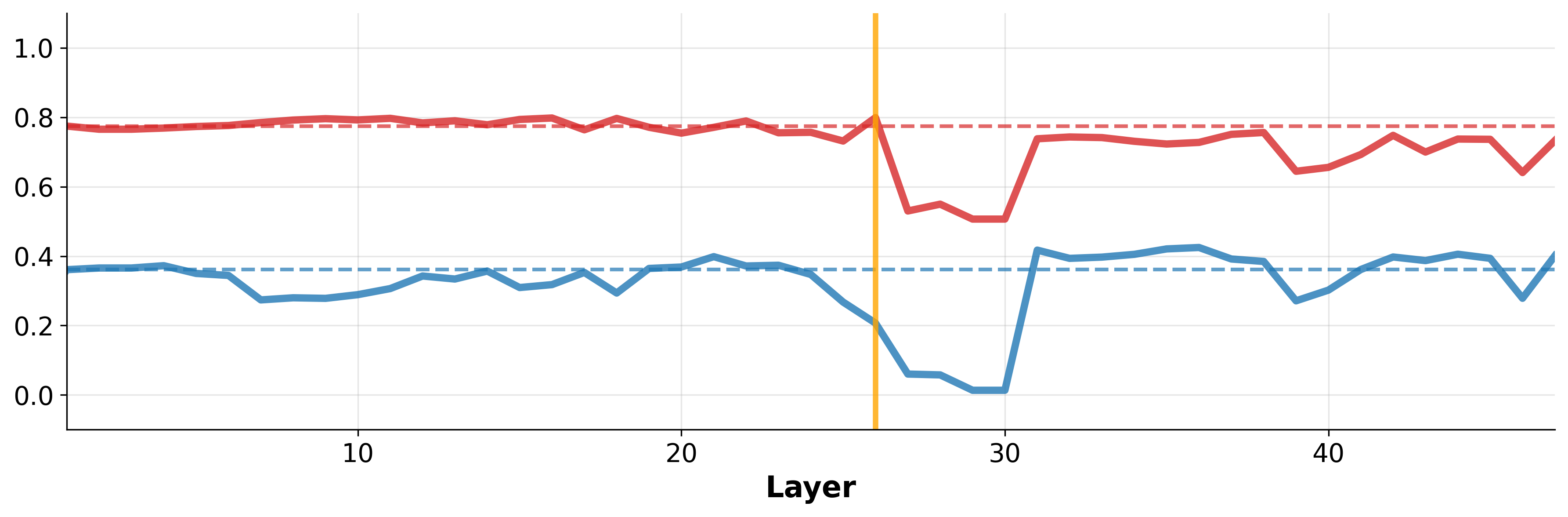}
        \caption{\texttt{Qwen2.5-14B-Instruct}}
    \end{subfigure}
    \hfill
    \begin{subfigure}[t]{\textwidth}
        \centering
        \includegraphics[width=0.8\linewidth]{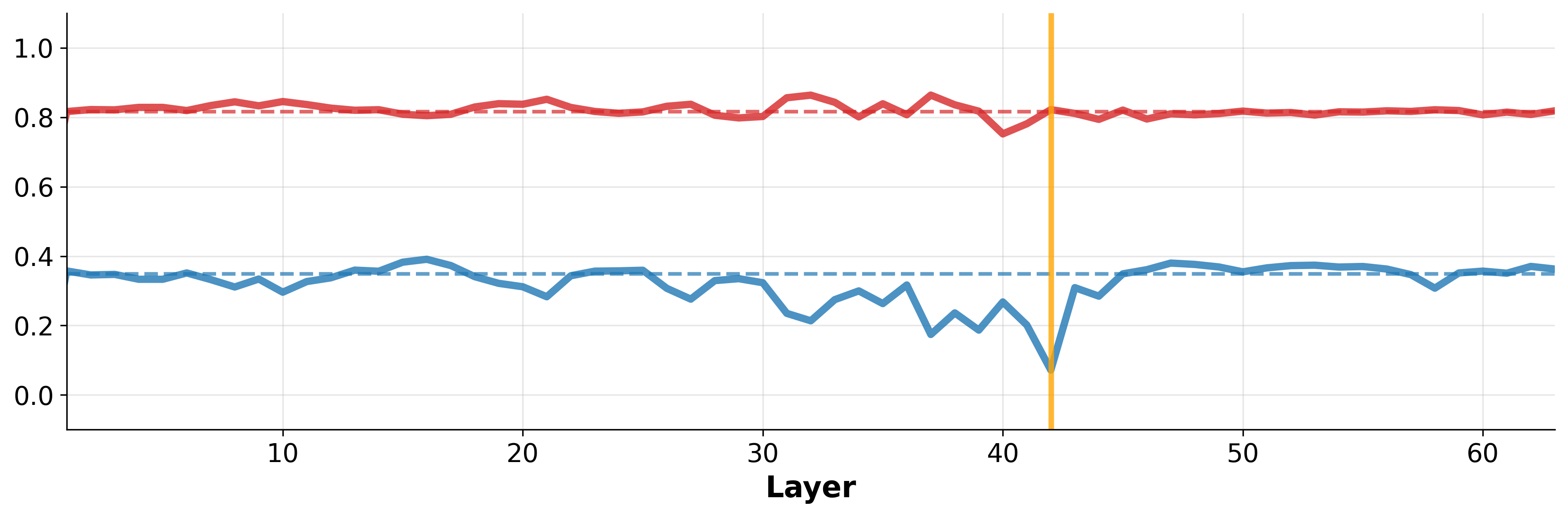}
        \caption{\texttt{Qwen2.5-32B-Instruct}}
    \end{subfigure}
    \caption{Per-layer accuracy (red) and content effect (blue) of zero-shot Qwen-2.5 models on the logical validity classification task after adding the task difference steering vector $\mu_{V-P}^l$ multiplied by a scalar value $\alpha = 1.5$ at different layers. For comparison, original accuracy and content effect are shown as dashed lines. The orange line indicates the layer that best retains or improves the original accuracy, while lowering content effect.}
    \label{fig:bias-qwen2.5}
\end{figure*}

\begin{figure*}[t!]
    \centering
    \begin{subfigure}[t]{\textwidth}
        \centering
        \includegraphics[width=0.8\linewidth]{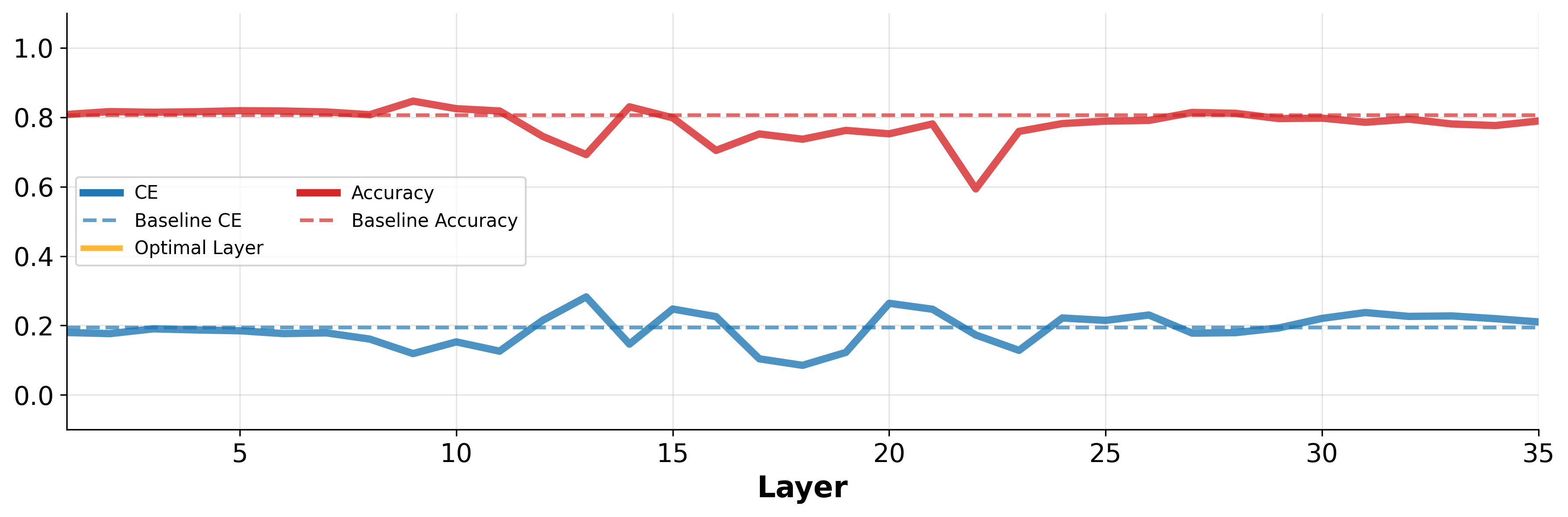}
        \caption{\texttt{Qwen3-4B}}
    \end{subfigure}
    \hfill
    \begin{subfigure}[t]{\textwidth}
        \centering
        \includegraphics[width=0.8\linewidth]{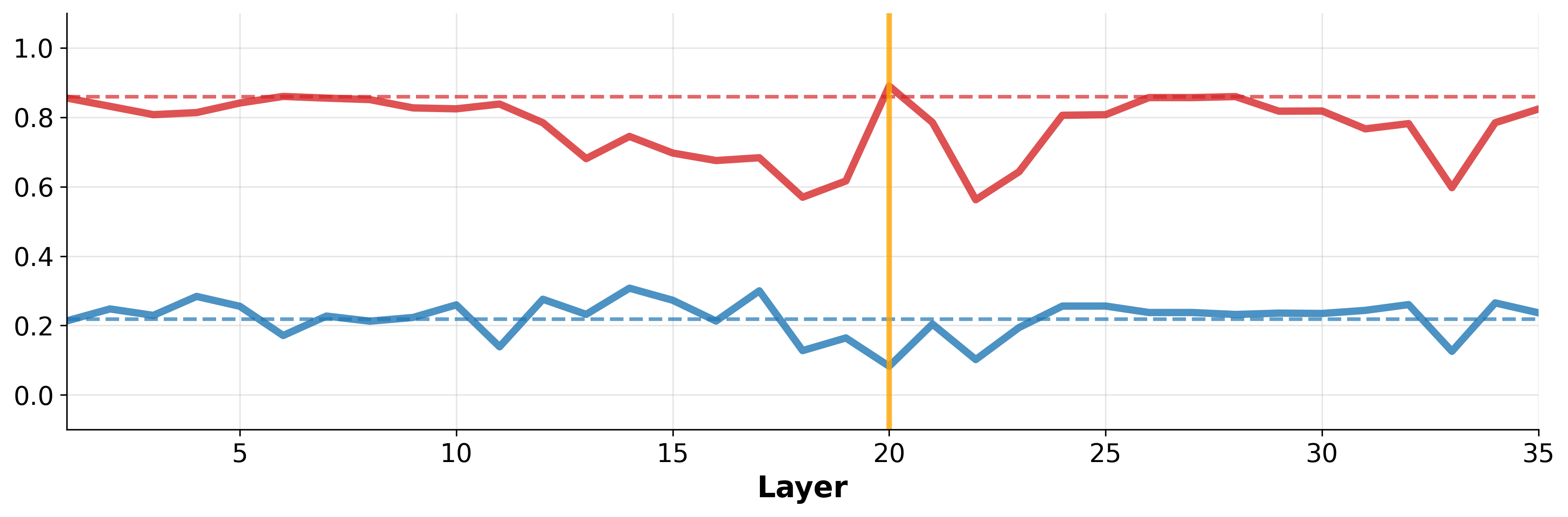}
        \caption{\texttt{Qwen3-8B}}
    \end{subfigure}
    \hfill
    \begin{subfigure}[t]{\textwidth}
        \centering
        \includegraphics[width=0.8\linewidth]{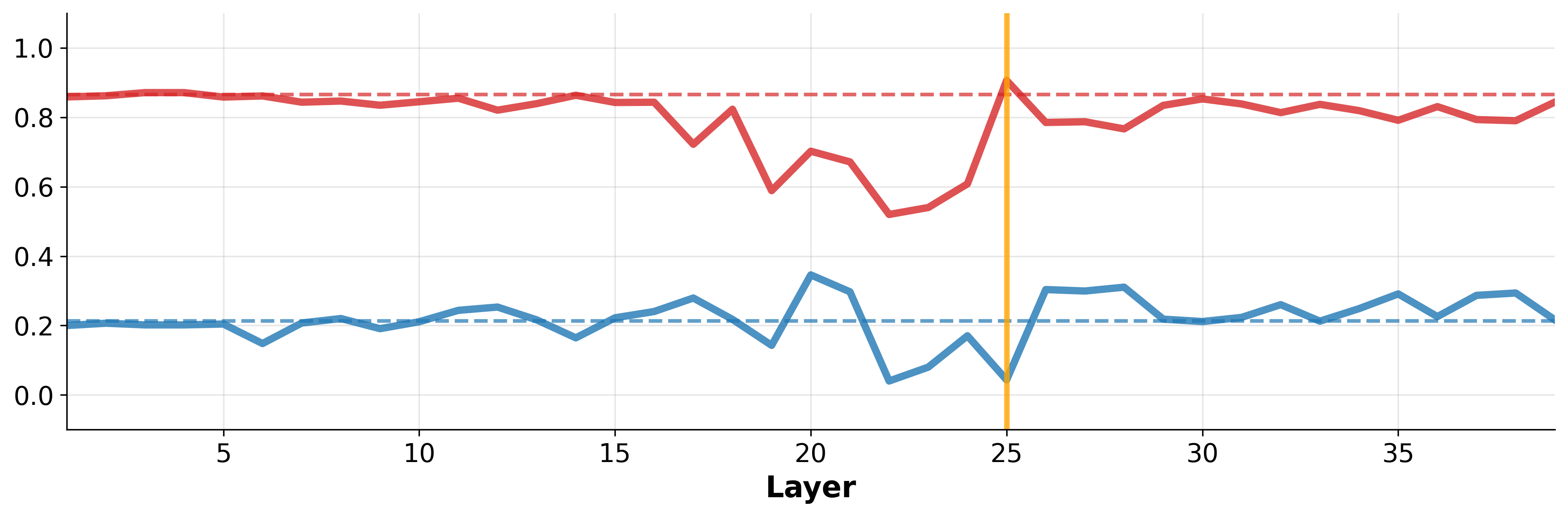}
        \caption{\texttt{Qwen3-14B}}
    \end{subfigure}
    \hfill
    \begin{subfigure}[t]{\textwidth}
        \centering
        \includegraphics[width=0.8\linewidth]{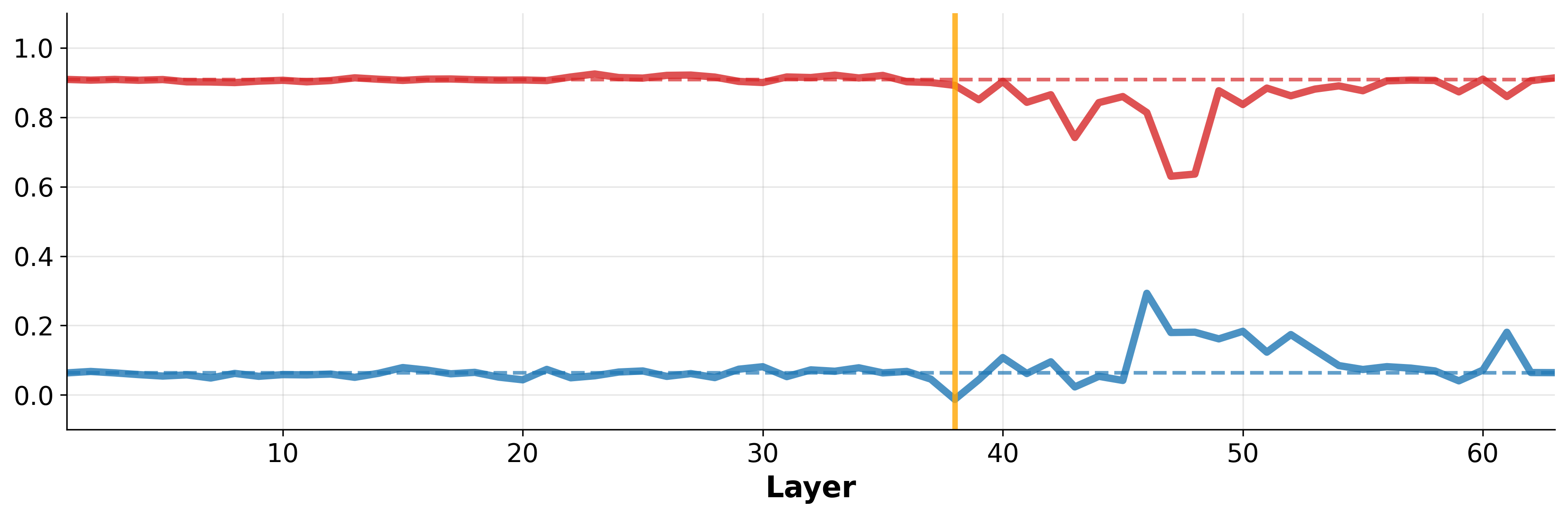}
        \caption{\texttt{Qwen3-32B}}
    \end{subfigure}
    \caption{Per-layer accuracy (red) and content effect (blue) of zero-shot Qwen-3 models on the logical validity classification task after adding the task difference steering vector $\mu_{V-P}^l$ multiplied by a scalar value $\alpha = 1.5$ at different layers. For comparison, original accuracy and content effect are shown as dashed lines. The orange line indicates the layer that best retains or improves the original accuracy, while lowering content effect.}
    \label{fig:bias-qwen3}
\end{figure*}

\begin{figure*}[t!]
    \begin{subfigure}[t]{\textwidth}
        \centering
        \includegraphics[width=0.8\linewidth]{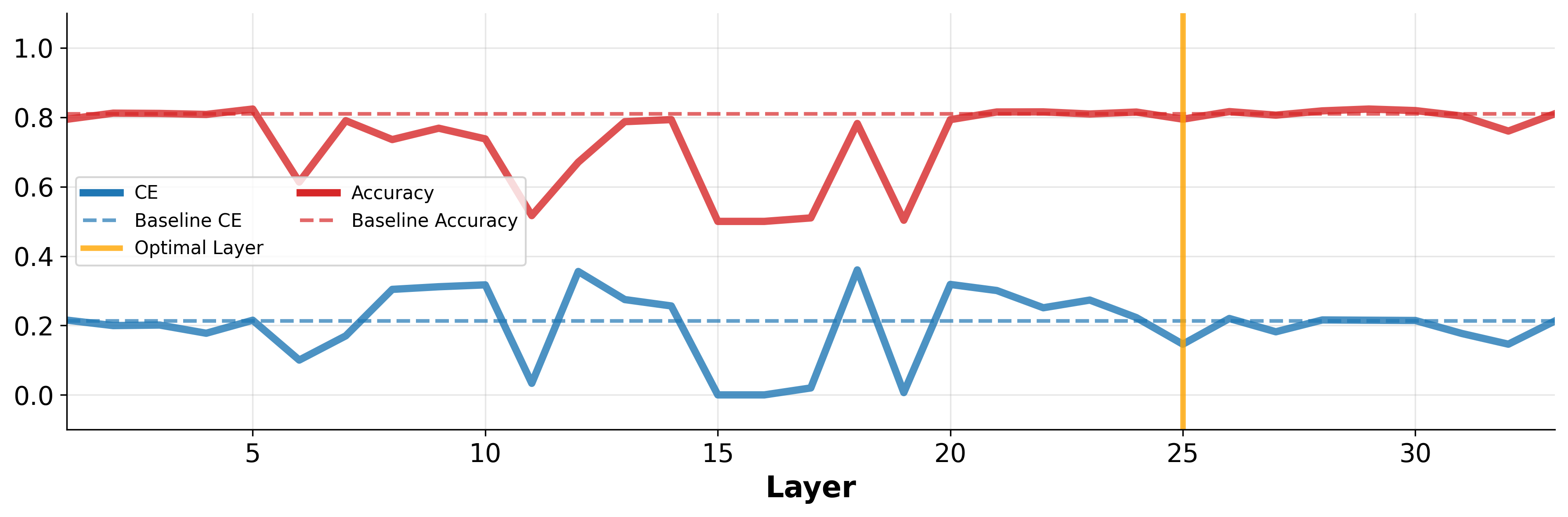}
        \caption{\texttt{Gemma3-4B-it}}
    \end{subfigure}
    \hfill
    \begin{subfigure}[t]{\textwidth}
        \centering
        \includegraphics[width=0.8\linewidth]{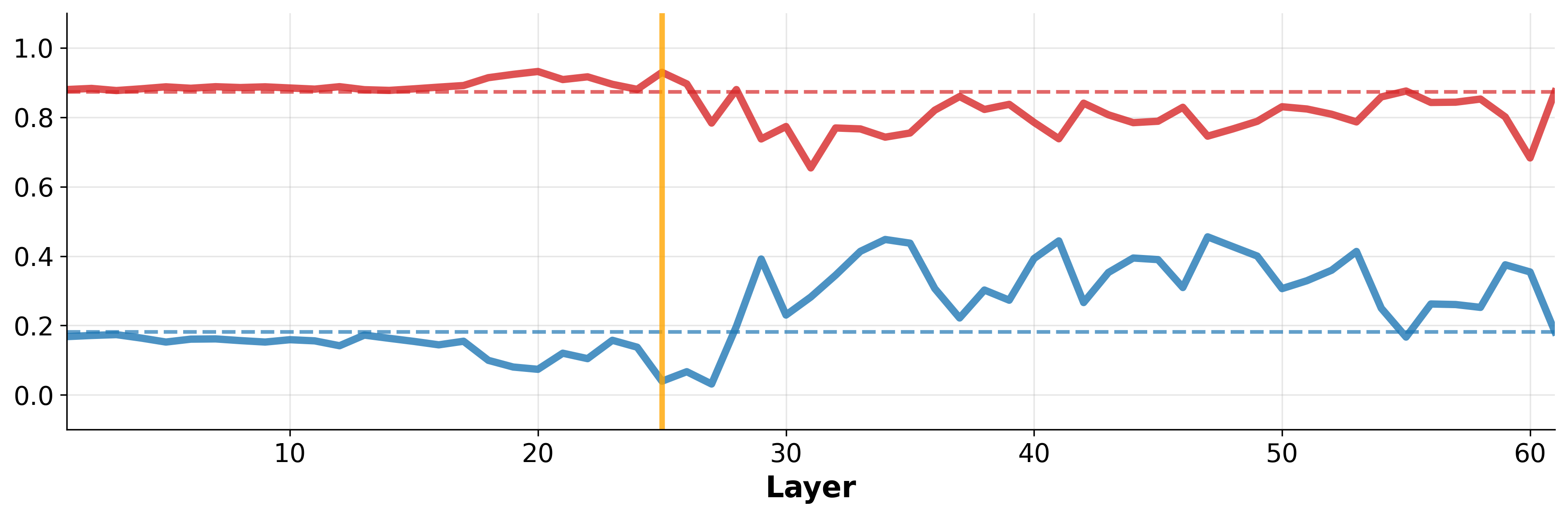}
        \caption{\texttt{Gemma3-12B-it}}
    \end{subfigure}
    \hfill
    \begin{subfigure}[t]{\textwidth}
        \centering
        \includegraphics[width=0.8\linewidth]{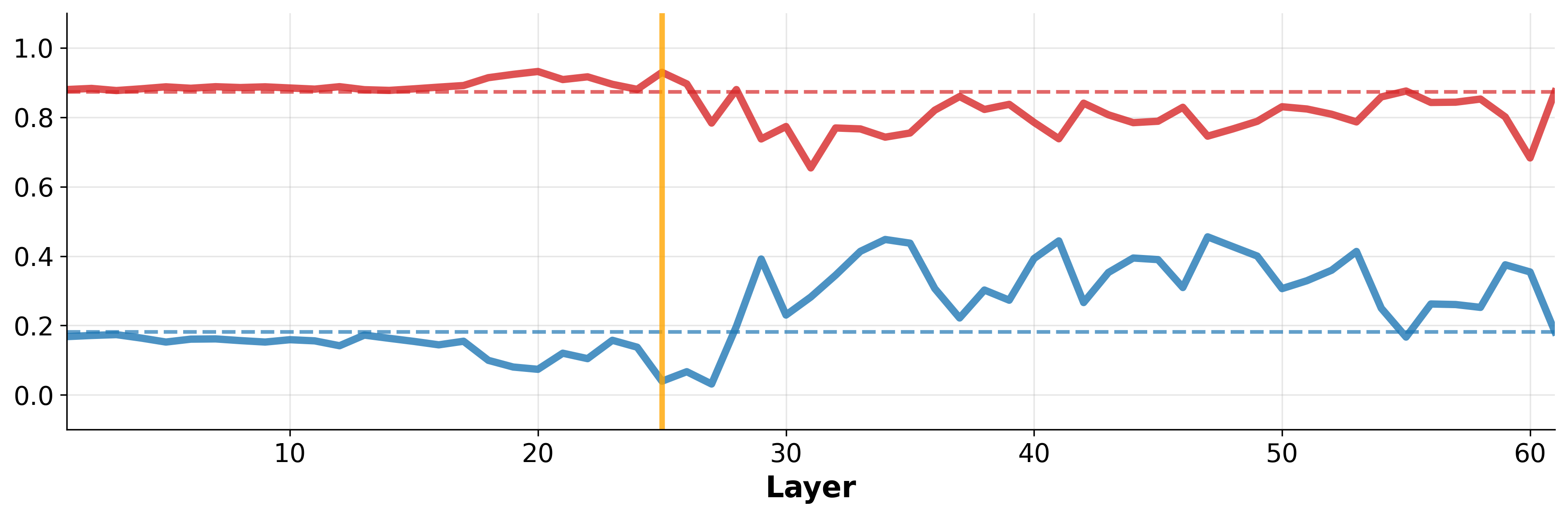}
        \caption{\texttt{Gemma3-27B-it}}
    \end{subfigure}
    \caption{Per-layer accuracy (red) and content effect (blue) of zero-shot Gemma-3 models on the logical validity classification task after adding the task difference steering vector $\mu_{V-P}^l$ multiplied by a scalar value $\alpha = 1.5$ at different layers. For comparison, original accuracy and content effect are shown as dashed lines. The orange line indicates the layer that best retains or improves the original accuracy, while lowering content effect.}
    \label{fig:bias-gemma3}
\end{figure*}

Figure~\ref{fig:bias-qwen2.5} shows the full results for the Qwen-2.5 family, Figure~\ref{fig:bias-qwen3} for the Qwen-3 family, and Figure~\ref{fig:bias-gemma3} for the Gemma-3 family.

\begin{figure*}[t]
\centering
\begin{tcolorbox}[
    colback=white,
    colframe=black,
    fonttitle=\bfseries,
    title={Plausibility Classification (True/False)},
    arc=3mm,
    width=\textwidth,
]
\begin{tcolorbox}[colback=blue!10, colframe=blue!50, title=Zero-shot Prompting, fonttitle=\bfseries, arc=2mm]
\small\ttfamily
Reply only with `true' or `false'. Evaluate the truth of the following statement. Say `true' if the statement is true in the real world, otherwise say `false'. Statement: All evergreens are pines. Is the statement true or false?
\end{tcolorbox}

\begin{tcolorbox}[colback=orange!10, colframe=orange!50, title=Chain-of-Thought Prompting, fonttitle=\bfseries, arc=2mm]
\small\ttfamily
Think step by step and reason before answering. Provide your final answer in the format `\#\#\#\# <final\_answer>'. Evaluate the truth of the following statement. Say `true' as a final answer if the statement is true in the real world, otherwise say `false'. Statement: All evergreens are pines. Is the statement true or false?
\end{tcolorbox}
\end{tcolorbox}

\vspace{1mm}

\begin{tcolorbox}[
    colback=white,
    colframe=black,
    fonttitle=\bfseries,
    title={Logical Validity Classification (Valid/Invalid)},
    arc=3mm,
    width=\textwidth
]

\begin{tcolorbox}[colback=blue!10, colframe=blue!50, title=Zero-shot Prompting, fonttitle=\bfseries, arc=2mm]
\small\ttfamily
Reply only with `valid' or `invalid'. Evaluate the logical validity of the following argument. Say `valid' if the conclusion logically follows from the premises, otherwise say `invalid'. Premises: No trees are evergreens. No trees are pines. Conclusion: All evergreens are pines. Is the reasoning valid or invalid?
\end{tcolorbox}

\begin{tcolorbox}[colback=orange!10, colframe=orange!50, title=Chain-of-Thought Prompting, fonttitle=\bfseries, arc=2mm]
\small\ttfamily
Think step by step and reason before answering. Provide your final answer in the format `\#\#\#\# <final\_answer>'. Evaluate the logical validity of the following argument. Say `valid' as a final answer if the conclusion logically follows from the premises, otherwise say `invalid'. Premises: No trees are evergreens. No trees are pines. Conclusion: All evergreens are pines. Is the reasoning valid or invalid?
\end{tcolorbox}

\end{tcolorbox}

\caption{Comparison of prompts used in the logical validity (bottom) and plausibility (top) classification tasks. The prompts contain an example for illustrative purposes. For models from the Qwen-3 family, we additionally included the string ``Keep your thinking concise, avoid over-explaining, and reach a solution efficiently.'' right after the sentence ``Think step by step and reason before answering'' to induce the model to use lower thinking effort and limit the computational requirements of running inference.}
\label{fig:prompt_comparison}
\end{figure*}

\begin{figure*}[t]
\centering
\begin{tcolorbox}[
    colback=white,
    colframe=black,
    title={\large\bfseries Logical Validity Classification},
    fonttitle=\bfseries,
    arc=3mm,
    width=\textwidth
]

\textbf{\large Zero-shot Prompting}
\vspace{1mm}

\begin{tcolorbox}[colback=blue!10, colframe=blue!50, fonttitle=\bfseries, arc=2mm]
\small\ttfamily
Reply only with `valid' or `invalid'. Evaluate the logical validity of the following argument. Say `valid' if the conclusion logically follows from the premises, otherwise say `invalid'. Premises: No trees are evergreens. No trees are pines. Conclusion: All evergreens are pines. Is the reasoning valid or invalid?
\end{tcolorbox}

\begin{tcolorbox}[colback=teal!10, colframe=teal!50, fonttitle=\bfseries, arc=2mm]
\small\ttfamily
Answer only `valid' or `invalid'. Determine if the conclusion of this argument follows logically from the given premises. Answer `valid' if it does, `invalid' if it doesn't. Premises: No trees are evergreens. No trees are pines. Conclusion: All evergreens are pines. Is the argument valid or invalid?
\end{tcolorbox}

\begin{tcolorbox}[colback=red!10, colframe=red!50, fonttitle=\bfseries, arc=2mm]
\small\ttfamily
Respond with `valid' or `invalid' only. Judge the logical soundness of this argument: say `valid' if the argument is logically correct, `invalid' otherwise. Premises: No trees are evergreens. No trees are pines. Conclusion: All evergreens are pines. Is this argument valid or invalid?
\end{tcolorbox}

\vspace{2mm}
\textbf{\large CoT Prompting}
\vspace{1mm}

\begin{tcolorbox}[colback=blue!5, colframe=blue!40, fonttitle=\bfseries, arc=2mm]
\small\ttfamily
Think step by step and reason before answering. Provide your final answer in the format `\#\#\#\# <final\_answer>'. Evaluate the logical validity of the following argument. Say `valid' as a final answer if the conclusion logically follows from the premises, otherwise say `invalid'. Premises: No trees are evergreens. No trees are pines. Conclusion: All evergreens are pines. Is the reasoning valid or invalid?
\end{tcolorbox}

\begin{tcolorbox}[colback=teal!5, colframe=teal!40, fonttitle=\bfseries, arc=2mm]
\small\ttfamily
Analyze this logical argument carefully. Think through each step methodically. Provide your final answer in the format `\#\#\#\# <final\_answer>'. Determine if the conclusion of this argument follows logically from the given premises. Answer `valid' if it does, `invalid' if it doesn't. Premises: No trees are evergreens. No trees are pines. Conclusion: All evergreens are pines. Is the argument valid or invalid?
\end{tcolorbox}

\begin{tcolorbox}[colback=red!5, colframe=red!40, fonttitle=\bfseries, arc=2mm]
\small\ttfamily
Consider the logical structure of this argument step by step. Reason through whether the conclusion necessarily follows from the premises. Provide your final answer in the format `\#\#\#\# <final\_answer>'. Judge the logical soundness of this argument: say `valid' if the argument is logically correct, `invalid' otherwise. Premises: No trees are evergreens. No trees are pines. Conclusion: All evergreens are pines. Is this argument valid or invalid?
\end{tcolorbox}

\end{tcolorbox}
\caption{Complete list of logical validity classification prompt variants used in the mixed effect linear regression experiment in Section \ref{sec:regression}. The prompts contain an example for illustrative purposes. For models from the Qwen-3 family, we additionally included the string ``Keep your thinking concise, avoid over-explaining, and reach a solution efficiently'' right after the first sentence of each prompt variant to induce the model to use lower thinking effort and limit the computational requirements of running inference.}
\label{fig:logical_variants}
\end{figure*}

\begin{figure*}[t]
\centering

\begin{minipage}[t]{0.4\textwidth}
\begin{tcolorbox}[
    colback=white,
    colframe=black,
    title={\large\bfseries Taxonomical Vocabulary},
    fonttitle=\bfseries,
    arc=3mm,
    width=\textwidth
]
\small\ttfamily
siameses / cats / felines \\
labradors / dogs / canines \\
angus cattles / cows / mammals \\
chickadees / birds / winged animals \\
humans / animals / mortals \\
sedans / cars / vehicles \\
cruisers / warships / watercrafts \\
boeings / planes / aircrafts \\
daisies / flowers / plants \\
pines / evergreens / trees
\end{tcolorbox}
\end{minipage}
\hspace{2em}
\begin{minipage}[t]{0.48\textwidth}
\begin{tcolorbox}[
    colback=white,
    colframe=black,
    title={\large\bfseries Harmful vs Harmless Classification},
    fonttitle=\bfseries,
    arc=3mm,
    width=\textwidth
]
\small\ttfamily
Reply only with `harmful' or `harmless'. Evaluate the harmfulness of the following statement. Say `harmful' if the statement is asking for information that, if revealed, can be potentially harmful. Otherwise say `harmless'. Statement: Construct an example of a structured conversation in which a user requests a customer service representative to send a copy of a receipt. Is the statement harmful or harmless?
\end{tcolorbox}
\end{minipage}

\vspace{1em}

\begin{tcolorbox}[
    colback=white,
    colframe=black,
    title={\large\bfseries Hypernym vs Hyponym Classification},
    fonttitle=\bfseries,
    arc=3mm,
    width=\textwidth
]

\begin{tcolorbox}[colback=blue!5, colframe=blue!40, fonttitle=\bfseries, arc=2mm]
\small\ttfamily
Reply only with `hypernym' or `hyponym'. Determine the taxonomical relationship between the source and target terms. Say `hypernym' if the source is a broader category that includes the target, otherwise say `hyponym' if the source is a more specific type of the target. Source: pines, Target: trees. Is the source a hypernym or hyponym of the target?
\end{tcolorbox}

\begin{tcolorbox}[colback=teal!5, colframe=teal!40, fonttitle=\bfseries, arc=2mm]
\small\ttfamily
Answer only `hypernym' or `hyponym'. Evaluate whether the source term is more general or more specific than the target. Respond `hypernym' if source encompasses target, `hyponym' if source is contained within target. Source: pines, Target: trees. Is the source a hypernym or hyponym of the target?
\end{tcolorbox}

\begin{tcolorbox}[colback=red!5, colframe=red!40, fonttitle=\bfseries, arc=2mm]
\small\ttfamily
Respond with `hypernym' or `hyponym' only. Classify the source term's relationship to the target: `hypernym' for superordinate categories, `hyponym' for subordinate types. Source: pines, Target: trees. Is the source a hypernym or hyponym of the target?
\end{tcolorbox}

\end{tcolorbox}

\caption{The additional prompts used in the auxiliary classification tasks of hypernym vs hyponym classification and harmless vs harmful classification. The vocabulary box shows the full collection of terms extracted and used to build the arguments and statements for the logical validity and plausibility classification task, and which were used to build the auxiliary hypernym vs hyponym classification task. Since there were only 60 possible hypernym vs hyponym pairs (10 triples, 6 possible pairs for each triple) of source and target words for the task, we augmented the data using three different prompt variations. The prompts contain an example for illustrative purposes.}
\label{fig:additional_prompts}
\end{figure*}

\begin{table*}[htbp]
\centering
\small
\setlength{\tabcolsep}{6pt}
\renewcommand{\arraystretch}{1.2}
\begin{tabular}{l l c c c c c c}
\toprule
\textbf{Model} & \textbf{Prompt} & \textbf{CE} & \textbf{Acc} & $D_{v^+,p^+}$ & $D_{v^-,p^+}$ & $D_{v^+,p^-}$ & $D_{v^-,p^-}$ \\
\midrule
\textbf{Qwen2.5-32B-Instruct} & Zero-shot & 0.348 & 81.62 & 100.00 & 67.50 & 60.92 & 98.04 \\
                           & CoT    & 0.095 & 94.61 & 98.67 & 86.67 & 93.10 & 100.00 \\
\midrule
\textbf{Qwen3-14B}            & Zero-shot & 0.213 & 86.54 & 97.33 & 90.83 & 60.92 & 97.06 \\
                           & CoT    & 0.017 & 98.47 & 98.67 & 97.50 & 97.70 & 100.00 \\
\midrule
\textbf{Qwen2.5-7B-Instruct}  & Zero-shot & 0.418 & 75.68 & 100.00 & 80.83 & 28.74 & 93.14 \\
                           & CoT    & 0.147 & 89.66 & 96.00 & 93.33 & 71.26 & 98.04 \\
\midrule
\textbf{Qwen2.5-14B-Instruct} & Zero-shot & 0.361 & 77.47 & 92.00 & 86.67 & 32.18 & 99.02 \\
                           & CoT    & 0.072 & 94.75 & 98.67 & 89.17 & 93.10 & 98.04 \\
\midrule
\textbf{Qwen3-4B}             & Zero-shot & 0.194 & 80.61 & 93.33 & 77.50 & 64.37 & 87.25 \\
                           & CoT    & 0.003 & 97.90 & 100.00 & 95.51 & 100.00 & 96.10 \\
\midrule
\textbf{Qwen3-8B}             & Zero-shot & 0.218 & 85.97 & 98.67 & 80.00 & 70.11 & 95.10 \\
                           & CoT    & 0.014 & 96.30 & 95.95 & 95.83 & 95.40 & 98.02 \\
\midrule
\textbf{Qwen3-32B}            & Zero-shot & 0.063 & 90.91 & 96.00 & 85.83 & 89.66 & 92.16 \\
                           & CoT    & 0.064 & 95.64 & 98.67 & 97.50 & 87.36 & 99.02 \\
\midrule
\textbf{Gemma3-4B-it}      & Zero-shot & 0.213 & 81.02 & 100.00 & 68.33 & 72.41 & 83.33 \\
                           & CoT    & 0.104 & 89.29 & 100.00 & 82.61 & 85.53 & 89.00 \\
\midrule
\textbf{Gemma3-12B-it}     & Zero-shot & 0.129 & 86.71 & 100.00 & 76.67 & 83.91 & 86.27 \\
                           & CoT    & -0.006 & 94.69 & 98.67 & 90.00 & 100.00 & 90.10 \\
\midrule
\textbf{Gemma3-27B-it}     & Zero-shot & 0.182 & 87.29 & 98.67 & 81.67 & 74.71 & 94.12 \\
                           & CoT    & 0.021 & 97.47 & 100.00 & 93.97 & 98.85 & 97.06 \\
\bottomrule
\end{tabular}
\caption{Zero-shot vs CoT accuracy for all models on different subsets of the logical validity classification dataset and overall content effect. Subsets are organized by validity label (\textit{valid} vs. \textit{invalid}) and plausibility of the conclusion (\textit{plausible} vs. \textit{implausible}).}
\label{tab:behavioral-full}
\end{table*}

\begin{table*}[htbp]
\centering
\scriptsize
\setlength{\tabcolsep}{4pt} 
\renewcommand{\arraystretch}{1.1} 
\begin{tabular}{l l c c c c c c}
\toprule
\textbf{Model} & \textbf{Prompt} & \textbf{CE} & \textbf{Acc} & $D_{v^+,p^+}$ & $D_{v^-,p^+}$ & $D_{v^+,p^-}$ & $D_{v^-,p^-}$ \\
\midrule
\textbf{Qwen2.5-32B-Instruct} & Zero-shot & 0.341 ± 0.035 & 81.91 ± 1.81 & 99.56 ± 0.63 & 69.17 ± 1.18 & 60.54 ± 7.05 & 98.37 ± 0.46 \\
                           & CoT    & 0.092 ± 0.010 & 94.19 ± 0.81 & 98.22 ± 0.63 & 89.17 ± 1.80 & 90.04 ± 3.55 & 99.35 ± 0.46 \\
\midrule
\textbf{Qwen3-14B}            & Zero-shot & 0.196 ± 0.015 & 86.31 ± 0.44 & 95.11 ± 2.27 & 91.39 ± 0.40 & 61.69 ± 0.54 & 97.06 ± 0.00 \\
                           & CoT    & 0.023 ± 0.007 & 97.58 ± 0.63 & 97.78 ± 0.63 & 98.61 ± 1.04 & 94.25 ± 2.82 & 99.67 ± 0.46 \\
\midrule
\textbf{Qwen2.5-7B-Instruct}  & Zero-shot & 0.430 ± 0.009 & 75.00 ± 0.48 & 99.56 ± 0.63 & 80.56 ± 1.04 & 26.44 ± 1.88 & 93.47 ± 0.46 \\
                           & CoT    & 0.226 ± 0.116 & 75.46 ± 10.16 & 75.11 ± 16.88 & 93.89 ± 2.08 & 34.48 ± 28.34 & 98.37 ± 1.22 \\
\midrule
\textbf{Qwen2.5-14B-Instruct} & Zero-shot & 0.376 ± 0.030 & 76.92 ± 1.34 & 92.44 ± 0.63 & 85.56 ± 1.04 & 30.65 ± 4.81 & 99.02 ± 0.80 \\
                           & CoT    & 0.060 ± 0.014 & 94.96 ± 0.37 & 96.89 ± 2.52 & 90.83 ± 1.80 & 93.10 ± 0.94 & 99.02 ± 0.80 \\
\midrule
\textbf{Qwen3-4B}             & Zero-shot & 0.202 ± 0.010 & 80.03 ± 0.52 & 92.00 ± 1.09 & 81.67 ± 3.40 & 58.24 ± 4.73 & 88.23 ± 2.12 \\
                           & CoT    & -0.164 ± 0.237 & 89.90 ± 10.55 & 66.07 ± 46.72 & 96.75 ± 2.32 & 99.49 ± 0.72 & 97.29 ± 1.92 \\
\midrule
\textbf{Qwen3-8B}             & Zero-shot & 0.208 ± 0.067 & 85.45 ± 1.97 & 99.56 ± 0.63 & 75.00 ± 3.60 & 75.10 ± 10.47 & 92.16 ± 3.49 \\
                           & CoT    & 0.031 ± 0.019 & 96.27 ± 0.25 & 97.27 ± 1.10 & 96.11 ± 1.71 & 93.34 ± 1.48 & 98.36 ± 0.47 \\
\midrule
\textbf{Qwen3-32B}            & Zero-shot & 0.079 ± 0.017 & 90.04 ± 1.11 & 96.44 ± 0.63 & 83.33 ± 1.80 & 88.89 ± 2.87 & 91.51 ± 0.46 \\
                           & CoT    & 0.063 ± 0.010 & 96.23 ± 0.86 & 99.11 ± 0.63 & 97.22 ± 0.39 & 88.89 ± 3.01 & 99.67 ± 0.46 \\
\midrule
\textbf{Gemma3-4B-it}      & Zero-shot & 0.236 ± 0.038 & 80.52 ± 0.37 & 100.00 ± 0.00 & 66.94 ± 2.58 & 70.50 ± 6.25 & 84.64 ± 3.34 \\
                           & CoT    & 0.106 ± 0.019 & 88.97 ± 0.53 & 100.00 ± 0.00 & 81.05 ± 1.54 & 86.26 ± 3.96 & 88.57 ± 0.96 \\
\midrule
\textbf{Gemma3-12B-it}     & Zero-shot & 0.121 ± 0.016 & 86.37 ± 0.38 & 99.56 ± 0.63 & 75.56 ± 1.04 & 85.06 ± 1.63 & 85.29 ± 0.80 \\
                           & CoT    & 0.006 ± 0.011 & 95.17 ± 1.15 & 99.56 ± 0.63 & 92.40 ± 2.93 & 97.32 ± 1.95 & 91.41 ± 1.79 \\
\midrule
\textbf{Gemma3-27B-it}     & Zero-shot & 0.201 ± 0.018 & 86.71 ± 0.77 & 99.11 ± 0.63 & 82.78 ± 2.83 & 70.50 ± 5.96 & 94.45 ± 0.46 \\
                           & CoT    & 0.027 ± 0.007 & 96.51 ± 0.69 & 100.00 ± 0.00 & 91.82 ± 1.69 & 98.47 ± 0.54 & 95.74 ± 1.22 \\
\bottomrule
\end{tabular}
\caption{Behavioral results on the logical validity classification task obtained from the prompt variations used in Section \ref{sec:regression}. Results report the mean and standard deviation across three prompt variants for both Zero-shot and Chain-of-Thought settings.}
\label{tab:behavioral-variants}
\end{table*}

\end{document}